\documentclass{article}

\usepackage{arxiv}

\usepackage[utf8]{inputenc} 
\usepackage[T1]{fontenc}    

\usepackage{url}            
\usepackage{booktabs}       
\usepackage{amsfonts}       
\usepackage{nicefrac}       
\usepackage{microtype}      
\usepackage{lipsum}
\usepackage{neuralnetwork}
\usepackage{lscape}

\usepackage{amsmath}
\usepackage{algorithm}
\usepackage[]{algpseudocode}

\usepackage{tikz}
\usetikzlibrary{calc,fadings}

\usetikzlibrary{matrix,chains,positioning,decorations.pathreplacing,arrows}
\usetikzlibrary{positioning}
\usetikzlibrary{patterns,decorations.pathmorphing,decorations.markings}

\tikzset{%
  every neuron/.style={
    circle,
    draw,
    fill = gray!25,
    thick,
    minimum size=1.1cm
  },
  neuron missing/.style={
    draw=none, 
    scale=4,
    fill = none,
    text height=0.333cm,
    execute at begin node=\color{black}$\vdots$
  },
  input neuron/.style={
    circle,
    draw,
    thick,
    fill = yellow!40,
    minimum size=1cm
  },
  output neuron/.style={
    circle,
    draw,
    thick,
    fill = blue!40,
    minimum size=1cm
  },
}

\usetikzlibrary{3d,decorations.text,shapes.arrows,positioning,fit,backgrounds}
\tikzset{pics/fake box/.style args={
#1 with dimensions #2 and #3 and #4}{
code={
\draw[gray,ultra thin,fill=#1]  (0,0,0) coordinate(-front-bottom-left) to
++ (0,#3,0) coordinate(-front-top-right) --++
(#2,0,0) coordinate(-front-top-right) --++ (0,-#3,0) 
coordinate(-front-bottom-right) -- cycle;
\draw[gray,ultra thin,fill=#1] (0,#3,0)  --++ 
 (0,0,#4) coordinate(-back-top-left) --++ (#2,0,0) 
 coordinate(-back-top-right) --++ (0,0,-#4)  -- cycle;
\draw[gray,ultra thin,fill=#1!80!black] (#2,0,0) --++ (0,0,#4) coordinate(-back-bottom-right)
--++ (0,#3,0) --++ (0,0,-#4) -- cycle;
\path[gray,decorate,decoration={text effects along path,text={}}] (#2/2,{2+(#3-2)/2},0) -- (#2/2,0,0);
}
}}
\tikzset{circle dotted/.style={dash pattern=on .05mm off 2mm,
                                         line cap=round}}

\usepackage{tikz-3dplot}
\tdplotsetmaincoords{60}{115}

\usepackage[]{algpseudocode}
\usepackage{pgfplots}
\usepackage{subcaption}
\usepackage{hyperref}       
\usepackage[]{threeparttable}
\usepackage{multirow,bm}

\newcommand{\D}{\mathcal{D}}

\newcommand{\Dval}{\mathcal{D}_\mathrm{val}}
\newcommand{\Dh}{\mathcal{D}_h}
\newcommand{\Dl}{\mathcal{D}_l}

\newcommand{\eem}{\mathbf{e}}

\newcommand{\mm}{\mathbf{m}}
\newcommand{\Ntr}{{N}}

\newcommand{\gm}{\mathbf{g}}

\newcommand{\um}{\mathbf{u}}
\newcommand{\vm}{\mathbf{v}}

\newcommand{\Wm}{\mathbf{W}}
\newcommand{\xm}{\mathbf{x}}

\newcommand{\ym}{\mathbf{y}}

\newcommand{\thetaa}{\boldsymbol{\theta}}
\newcommand{\betaa}{\boldsymbol{\beta}}
\newcommand{\pinn}{\widehat{\mathcal{M}}}
 
\newcommand{\sigmaa}{\boldsymbol\sigma} 
\newcommand{\Psii}{\boldsymbol\Psi}

\newcommand\Tstrut{\rule{0pt}{2.9ex}}       
\newcommand\Bstrut{\rule[-1.2ex]{0pt}{0pt}} 

\title{Neural Network Training Using $\ell_1$-Regularization and Bi-fidelity Data}

\author{
  Subhayan De\\ 
  Aerospace Engineering Sciences\\
  University of Colorado\\
  Boulder, CO 80303 \\
  \texttt{Subhayan.De@colorado.edu} \\
   \And
 Alireza Doostan \\
  Aerospace Engineering Sciences\\
  University of Colorado\\
  Boulder, CO 80303 \\
  \texttt{Alireza.Doostan@colorado.edu} \\
}

\begin{document}
\maketitle

\begin{abstract} 
With the capability of accurately representing a functional relationship between the inputs of a physical system's model and output quantities of interest, neural networks have become popular for surrogate modeling in scientific applications. However, as these networks are over-parameterized, their training often requires a large amount of data. To prevent overfitting and improve generalization error, regularization based on, e.g., $\ell_1$- and $\ell_2$-norms of the parameters is applied. Similarly, multiple connections of the network may be pruned to increase sparsity in the network parameters. In this paper, we explore the effects of sparsity promoting $\ell_1$-regularization on training neural networks when only a small training dataset from a high-fidelity model is available. As opposed to standard $\ell_1$-regularization that is known to be inadequate, we consider two variants of $\ell_1$-regularization informed by the parameters of an identical network trained using data from lower-fidelity models of the problem at hand. These bi-fidelity strategies are generalizations of transfer learning of neural networks that uses the parameters learned from a large low-fidelity dataset to efficiently train networks for a small high-fidelity dataset. We also compare the bi-fidelity strategies with two $\ell_1$-regularization methods that only use the high-fidelity dataset. Three numerical examples for propagating uncertainty through physical systems are used to show that the proposed bi-fidelity $\ell_1$-regularization strategies produce errors that are one order of magnitude smaller than those of networks trained only using datasets from the high-fidelity models. 

\end{abstract}


\section{Introduction}

The behavior of a physical system is often described by models expressed using partial differential equations. There may be many sources of uncertainty in these models, \textit{e.g.}, in material properties, geometry, or loading conditions. Uncertainty quantification (UQ) aims to estimate the effects of these uncertainties on the response of the system predicted by the models. Approaches such as Monte Carlo for UQ evaluate the models for multiple realizations of the uncertainty. However, for complex models this exercise becomes computationally expensive. To ameliorate the large cost of UQ using accurate and complex models, surrogate models can be developed using polynomial chaos expansion \cite{ghanem2003stochastic,xiu2002wiener}, stochastic collocation \cite{babuvska2007stochastic,nobile2008sparse}, and Gaussian process regression \cite{williams2006gaussian,forrester2007multi}. However, the cost of developing these surrogate models increases with the dimension of the uncertain parameters~\cite{Doostan09b,doostan2011non}. 
In scientific machine learning, neural networks are used as surrogate models to represent the behavior of complex physical systems \cite{baker2019workshop}. Recently, Raissi et al. \cite{raissi2018hidden} trained neural networks for modeling physical systems that are described using differential equations. Bhatnagar et al. \cite{bhatnagar2019prediction} and Lee and You \cite{lee2019data} used neural networks for predicting aerodynamic flows. Adcock et al. \cite{adcock2020deep} used neural networks to learn Hilbert-valued functions using limited data. Adcock and Dexter \cite{adcock2020gap} further compared the performance of neural networks for approximating quantities of interest with  function approximations based on compressed sensing \cite{donoho2006compressed} to show that much progress is needed in neural network training to outperform the current state-of-the-art methods for general functions.  

As deep neural networks with multiple hidden layers have a large number of parameters, they are prone to overfitting when trained using small datasets. To mitigate this issue, connections in large neural networks are \textit{pruned} to obtain sparsity in the network and its parameters \cite{cheng2017survey,liu2018rethinking,hoefler2021sparsity}. 
In one of the early works, Reed \cite{reed1993pruning} classified the pruning algorithms for neural networks into two categories. In the first category, a network connection is deleted based on a hard threshold on the sensitivity of the training error with respect to that connection. For example,  LeCun et al. \cite{lecun1990optimal}, and Hassibi and Stork \cite{hassibi1993second} used the second derivative of the loss function with respect to the weights to prune a network. However, this requires additional computation and the pruned network needs retraining for the new sparse configuration \cite{han2015learning}.  Han et al. \cite{han2015learning,han2016dsd} pruned neural networks by dropping connections with smaller weights and then retraining the pruned network. In convolutional neural networks (CNN), Li et al. \cite{li2016pruning} entirely removed filters with small contribution, instead of magnitude-based pruning of weights. 
Mallya and Lazebnik \cite{mallya2018packnet} used a single network for multiple tasks but for each of these tasks the connections were pruned following Han et al. \cite{han2015learning,han2016dsd}. 
In the \textit{dropout} strategy to prevent overfitting, some of the outputs from fully connected layers  are dropped randomly but with a prescribed probability \cite{hinton2012improving,srivastava2014dropout}. 
Wan et al. \cite{wan2013regularization} generalized the dropout by introducing random sparsity in the connections rather than the layer output.   
Recently, Frankle and Carbin \cite{frankle2018lottery} proposed a \textit{lottery ticket hypothesis} for pruned neural networks, which states that a pruned subnetwork known as the \textit{winning ticket} can be obtained from a randomly initialized dense network that matches the performance of the dense network. However, this subnetwork needs to be trained from the same initialization as the dense network. Also, the computational cost of finding such a subnetwork can be prohibitive. 

The second category augments the training objective by including regularization terms that suppress some of the connections. For example, an $\ell_2$-norm of the weights were added to the objective to train classifiers in Krogh and Hertz \cite{krogh1992simple}. Zhou et al. \cite{zhou2016less}, and Lebedev and Lempitsky \cite{lebedev2016fast} used a summation of $\ell_2$-norms of different groups of parameters of a CNN, known as $\ell_{2,1}$-norm, as the regularization term. Motivated by the developments in sparse approximation via compressed sensing, $\ell_0$- and $\ell_1$-regularizations have been widely used in training neural networks, see, e.g, Collins and Kohli \cite{collins2014memory} and Liu et al. \cite{liu2017learning}. 
Sparsity was sought in groups of network parameters in Scardapane \cite{scardapane2017group} and Wen et al. \cite{wen2016learning} using the $\ell_1$- and $\ell_{2,1}$-norms. A combination of $\ell_1$- and $\ell_2$-regularizations was also experimented on in Alvarez and Salzmann \cite{alvarez2016learning} and Yang et al. \cite{yang2019deephoyer}. As $\ell_0$-norm is not differentiable, the treatment of $\ell_0$-regularization requires special optimization solvers. Additionally, the standard $\ell_1$-norm is often inadequate in promoting sparsity in the network parameters due to its fast shrinkage property and lack of scale invariance, as discussed in Yang et al.~\cite{yang2019deephoyer} and Hoefler et al.~\cite{hoefler2021sparsity}.

In this paper, we consider three novel $\ell_1$-regularization strategies to induce sparsity in the neural network parameters. The first strategy, inspired by the work of Candes et al. \cite{candes2008enhancing}, implements a weighted $\ell_1$-norm of parameters where the weights are iteratively set based on the values of the parameters obtained from the previous optimization epoch. The second and third approaches leverage data from lower-fidelity models of the problem at hand to construct $\ell_1$-regularization. These {\it bi-fidelity} strategies are motivated by the fact that in scientific applications often multiple models are used to predict output quantities of interest (QoIs). 
Among them, \textit{high-fidelity} models provide higher levels of accuracy, but in general are computationally expensive. The use of these models to generate a large and accurate training dataset may become computationally infeasible as a result. On the other hand, models with lower accuracy levels, known as \textit{low-fidelity} models, are often computationally less expensive. Examples of low-fidelity models include a coarse finite element discretization of partial differential equations, reduced order models \cite{benner2015survey}, or regression models \cite{rasmussen2004gaussian,forrester2009recent}. In particular, the first bi-fidelity strategy uses $\ell_1$-norm of the difference between parameters of the network trained from the high-fidelity dataset and the corresponding parameters from the identical low-fidelity trained network. In doing so, the deviation from low-fidelity network parameters are sparsely distributed over the entire network parameters, determined by the high-fidelity data. In essence, this bi-fidelity strategy is a generalization of the transfer learning techniques of~\cite{motamed2019multi,de2020transfer,meng2020composite}, where deviations from the low-fidelity network are allowed only for parameters of a single -- often the last hidden -- layer. The second bi-fidelity strategy uses weighted $\ell_1$-norm of the parameters where the weights are inversely proportional to the magnitude of the corresponding parameters of an identical network trained using low-fidelity data. Unlike in standard pruning based on hard thresholding, this weighted $\ell_1$-regularization allows for small parameters of the low-fidelity network to take large values, if indeed they happen to considerably influence the approximation of the high-fidelity QoI. 

Three numerical examples for building surrogate models are used to illustrate and compare the three approaches to $\ell_1$-regularization. Results based on the standard $\ell_1$-regularization, where $\ell_1$-norm of the parameters of the networks is penalized, will also be presented. In the first example, we consider bending of a composite beam, whereas the second example studies heat driven flow through a cavity. In both examples the surrogate model is a feed-forward neural network. The third example trains a feed-forward autoencoder networks for the steady-state solution of an isentropic flow in a dual-throat nozzle under uncertain initial conditions. These three examples show that the bi-fidelity $\ell_1$-regularization strategies to train the neural networks produce more accurate surrogates, when only a handful of data points are available from the high-fidelity model along with a large training dataset generated from the low-fidelity model. 

The rest of the paper is organized as follows. A background section first discusses uncertainty propagation through a physical system, neural networks for developing surrogate models, and the use of $\ell_1$-regularization during optimization. In the subsequent section, we describe the three $\ell_1$-regularization strategies of this work. Three numerical examples are used thereafter to illustrate these strategies. We finally summarize and conclude the paper with a discussion on future directions for the use of $\ell_1$-regularization in scientific machine learning applications. 

\section{Background} 

In this section, we briefly describe the construction of neural network surrogates for uncertainty propagation through a physical system. The use of neural networks as autoencoder and their training using gradient descent methods are discussed next. A brief background on standard $\ell_1$-regularization is also reviewed.  

\subsection{Uncertainty Propagation}

In engineering applications, models in the form of partial differential equations are used to represent the behavior of a physical system. They map the inputs $\xm\in\mathbb{R}^{d_i}$ to a QoI $\ym\in\mathbb{R}^{d_o}$ given by
\begin{equation}\label{eq:model}
\ym=\mathcal{M}(\xm), 
\end{equation}
where $\mathcal{M}(\cdot)$ represents the model. 
In UQ applications, the input contains uncertainty in loading, boundary conditions, and/or material properties often modeled by random variables following some known probability distribution functions. As a result, $\ym$ is also a random vector, and the goal of UQ is to estimate its statistics, such as, mean, covariance, probability distribution function, etc. A surrogate $\widehat{\mathcal{M}}$ (e.g., a neural network) to the model $\mathcal{M}$ can be used for uncertainty propagation to ameliorate the computational cost of evaluating complex and large-scale models many times. However, the construction of $\widehat{\mathcal{M}}$ requires datasets from the model $\mathcal{M}$. In this paper, we investigate developing neural networks as the surrogate model $\widehat{\mathcal{M}}$ with datasets from the accurate model $\mathcal{M}$ as well as from a less accurate and cheaper model of the same system. We focus our efforts on standard feed-forward neural networks (FNN) as well as autoencoders, while noting that the proposed strategies may be readily extended to other architectures.

\subsection{Standard Feed-Forward Neural Networks}

Neural networks consist of a sequence of connected layers, which contains an input layer, one or more hidden layers, and an output layer. The hidden layers have multiple neurons, where an affine transformation is applied to its input followed by a nonlinear function known as activation. 
In this paper, we use FNN \cite{goodfellow2016deep} among many available network configurations. The FNN maps the inputs $\xm$ to output $\ym$ as follows
\begin{equation}\label{eq:nn}
\begin{split}
\ym &\approx \pinn\left(\xm;\{\Psii_i\}_{i=0}^{H}, \{\betaa_i\}_{i=0}^{H} \right)\\
&:= \Psii_0\sigmaa_{H}(\dots\sigmaa_2(\Psii_2(\sigmaa_1(\Psii_1\xm+\betaa_1)+\betaa_2)\dots)+\betaa_0,\\
\end{split}
\end{equation}
where $H$ is the number of hidden layers; $\{\Psii_i\}_{i=1}^{H}$ and $\{\betaa_i\}_{i=1}^{H}$ are weight matrices and bias vectors, respectively; $\Psii_0$ and $\betaa_0$ are the weight matrix and bias vector for the output layer, respectively; and $\sigmaa_i(\cdot)$ are (vector-valued) activation functions for the $i$th hidden layer. Note that an activation function can also be applied to the output layer. 

Among many choices for activation function, rectified linear unit (ReLU) is most commonly used, which is given by
\begin{equation}
\sigma_{\!_\mathrm{ReLU}}(z) = \max(0,z),
\end{equation}
for an input $z$. However, the gradient of the output becomes zero for negative inputs. This creates a problem, known as the dying ReLU problem, during the training using gradient descent methods. An exponential linear unit (ELU) avoids this problem by introducing an exponential relation for negative inputs as follows \cite{clevert2015fast}
\begin{equation} \label{eq:elu}
\sigma_{\!_\mathrm{ELU}}(z) = \begin{cases}
z, \qquad \qquad \quad~~~\! \text{for } z>0,\\
\alpha (e^z-1), \qquad \text{for } z\leq0,\\
\end{cases}
\end{equation}
where $\alpha$ is a positive parameter and assumed as one herein. 
In one of our numerical examples, we also use a hyperbolic tangent as activation function, i.e., $\sigma_{\!_\mathrm{Tanh}}(z)=\mathrm{tanh}(z)$. 
In this paper, we choose activation functions for the numerical examples as the one that produces the smallest prediction errors.

\subsection{Autoencoders}
\begin{figure}[!htb]
	\centering
	\includegraphics[scale=1.0]{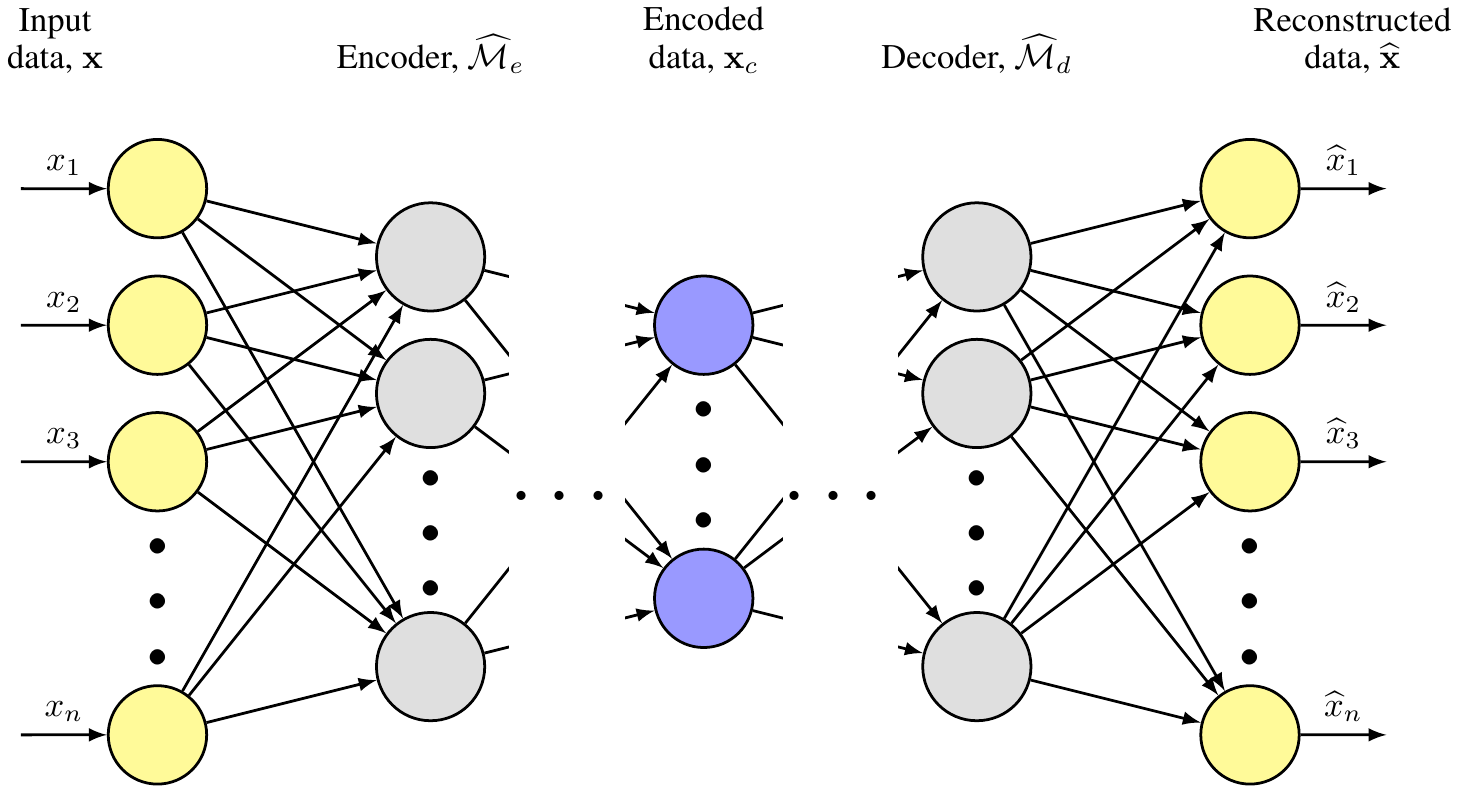}
	\caption{A schematic of an autoencoder, where an FNN encoder $\widehat{\mathcal{M}}_e$ is used to encode the data $\xm$ in a latent space as $\xm_{c}$ and an FNN decoder $\widehat{\mathcal{M}}_d$ is used to reconstruct the data as $\widehat\xm$.}\label{fig:ae}
\end{figure}

Autoencoders are neural networks that are trained to reproduce the input, i.e., $\ym\approx \xm$ in (\ref{eq:model}). 
These neural networks have an encoder part, where the input is mapped onto a latent space and a decoder part to reconstruct the input data from the latent space information. Hence, if FNNs are used as encoder and decoder as shown in Figure \ref{fig:ae}, we can write 
\begin{equation}
\begin{split}
\xm_{c} &= \widehat{\mathcal{M}}_{e}\left(\xm;\{\Psii_{e,i}\}_{i=0}^{H_e}, \{\betaa_{e,i}\}_{i=0}^{H_e} \right);\\
\widehat{\xm} &= \widehat{\mathcal{M}}_{d}\left(\xm_{c};\{\Psii_{d,i}\}_{i=0}^{H_d}, \{\betaa_{d,i}\}_{i=0}^{H_d} \right),\\
\end{split}
\end{equation}
where $\xm_{c}\in\mathbb{R}^{d_l}$ is the encoded data vector in the latent space; $\widehat{\mathcal{M}}_{e}(\cdot;\cdot,\cdot)$ and $\widehat{\mathcal{M}}_{d}(\cdot;\cdot,\cdot)$ denote the encoding and decoding networks, respectively; $\{\Psii_{e,i}\}_{i=0}^{H_e}, \{\betaa_{e,i}\}_{i=0}^{H_e}$ are weights and biases of the encoding network with $H_e$ hidden layers, respectively; and $\{\Psii_{d,i}\}_{i=0}^{H_d}, \{\betaa_{d,i}\}_{i=0}^{H_d}$ are weights and biases of the decoding network with $H_d$ hidden layers, respectively. 
For undercomplete autoencoders, the dimension of the latent space for encoded data is smaller than the dimension of the input, i.e., $d_l<d_i$, which helps in identifying the important characteristics of the input data and can be used for dimensionality reduction \cite{goodfellow2016deep}. In this paper, we use an example of autoencoder trained using strategies discussed in Section \ref{sec:l1_strategies} to illustrate their efficacy. 

\subsection{Training of a Neural Network}
\label{sec:training}

During supervised training of neural networks, we estimate the parameters $\thetaa:= \{\Psii_i, \betaa_i\}_{i=0}^{H}$ by minimizing a loss function $J(\thetaa)$ using a dataset $\mathcal{D}:=\{(\xm_i,\ym_i)\}_{i=1}^{\Ntr}$, where $\ym_i := \mathcal{M}(\xm_i;\thetaa)$. For a loss function that consists of the mean squared error (MSE), the minimization problem $\mathcal{P}_2$ can be defined as
\begin{equation} \label{eq:P2}
\mathcal{P}_2\equiv \left\{ \arg\min_{\thetaa} J(\thetaa):=\frac{1}{\Ntr}\sum_{i=1}^{\Ntr} J_i(\thetaa):= \frac{1}{\Ntr}\sum_{i=1}^{\Ntr} \Big\lVert\ym_i - \pinn(\xm_i;\thetaa) \Big\rVert^2_2 \right\},
\end{equation}
where $\lVert\cdot\rVert_2$ denotes the $\ell_2$-norm of its vector argument. Note that for the training of an autoencoder the loss function is defined using the reconstructed data $\widehat{\xm}$ instead of $\ym$. A stochastic gradient descent (SGD) method can be used to solve the optimization problem $\mathcal{P}_2$, which updates the parameters at $k$th iteration as follows
\begin{equation}
\begin{split}
\thetaa_{k+1} &\leftarrow \thetaa_{k} - \eta_k\frac{\partial J_{i_k}(\thetaa)}{ \partial \thetaa_{k}}.\\
\end{split}
\end{equation}
Here, $\eta_k$ is the step size, also known as the learning rate, and $i_k$ is randomly selected index from $\{1,\dots,N\}$, which is used to estimate the gradient. There are many modifications available for the SGD method \cite{ruder2016overview,de2019topology,de2019bifidelity}. In this paper, we use adaptive moment or Adam algorithm \cite{kingma2014adam} for training. This modification retards the movement in the direction of large historical gradients. A brief description of Adam is provided in Appendix \ref{sec:adam}.

\subsection{$\ell_1$-Regularization}\label{sec:l1_background}
\begin{figure}[!htb]
	\centering
	\begin{subfigure}[t]{0.3\textwidth}
		\centering
		\includegraphics[scale=1.0]{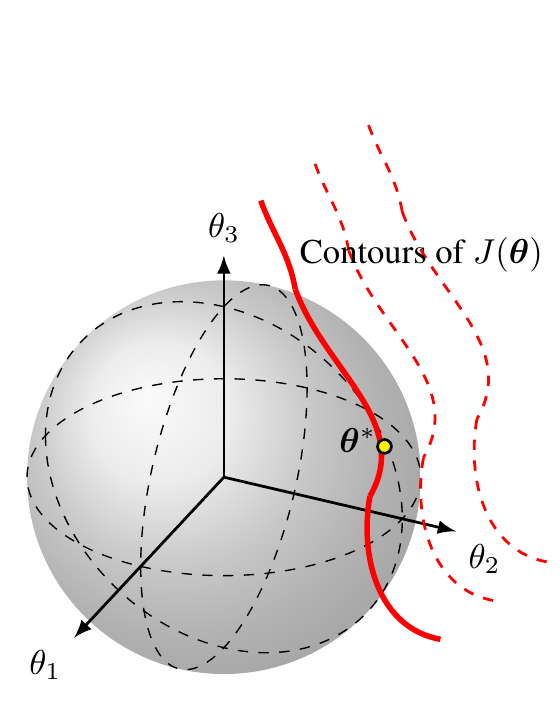}
		\caption{$\ell_2$ regularization}\label{fig:l2}
	\end{subfigure}\hfill
	\begin{subfigure}[t]{0.3\textwidth}
		\centering
		\includegraphics[scale=1.0]{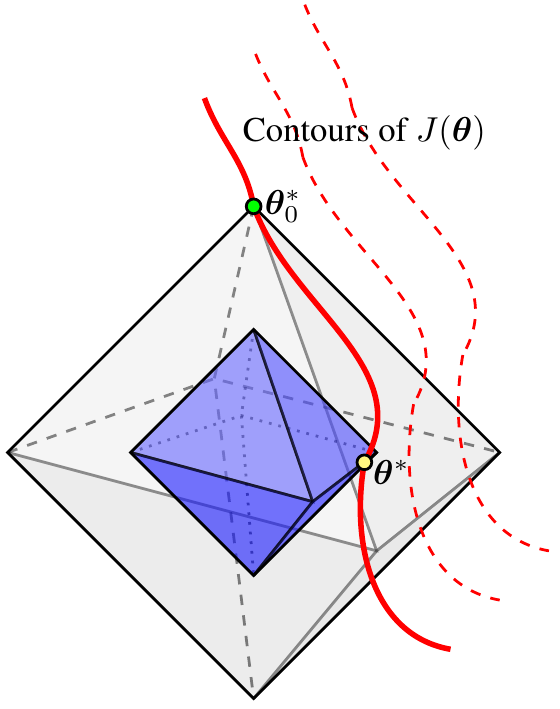}
		\caption{Standard $\ell_1$-regularization}\label{fig:std_l1}
	\end{subfigure}\hfill
	\begin{subfigure}[t]{0.3\textwidth}
		\centering
		
		
		
		
		
		
		
		\includegraphics[scale=1.0]{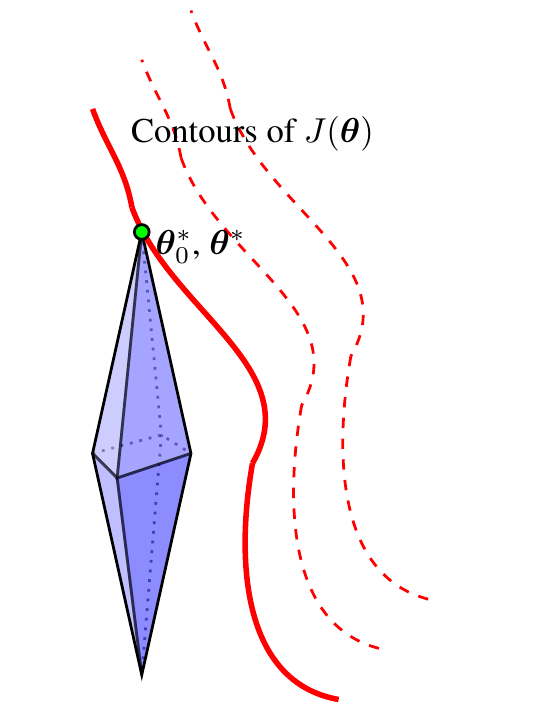}
		\caption{Weighted $\ell_1$-regularization}\label{fig:wl_1}
	\end{subfigure}\hfill
	\caption{Comparison of $\ell_2$, standard $\ell_1$, and weighted $\ell_1$-regularization. The sparse solution is denoted as $\thetaa^*_0$, whereas $\thetaa^*$ denotes a solution of the regularized problem. The contours of the loss function $J(\thetaa)$ are drawn on a plane perpendicular to the regularization surface.} 
	\label{fig:l1_l2_comp}
\end{figure}

A regularization term is often added to the objective of the optimization in \eqref{eq:P2} to prevent overfitting in ill-posed problems. The most common is to use a Tikhonov regularization \cite{tikhonov2013numerical}, which can be implemented in a minimization problem $\mathcal{P}_{2,\lambda}$ as follows 
\begin{equation}\label{eq:std_l2}
\mathcal{P}_{2,\lambda} \equiv \Big\{ \arg\min_{\thetaa} R(\thetaa) := J(\thetaa) + \lambda\lVert \thetaa \rVert_2 \Big \},
\end{equation}
where $R(\thetaa)$ is the objective of the minimization problem and $\lambda$ denotes the strength of the regularization. Figure \ref{fig:l2} shows that the regularization term in $\mathcal{P}_{2,\lambda}$ defines the surface of a hypersphere. Hence, for a loss function $J(\thetaa)$ shown with red contours, the solution $\thetaa^*$ of the problem $\mathcal{P}_{2,\lambda}$ will not be sparse. 
To induce sparseness in the solution for the parameters $\thetaa$, the following optimization problem can be used instead 
\begin{equation}\label{eq:std_l1}
\mathcal{P}_{1,\lambda} \equiv \Big\{ \arg\min_{\thetaa} R(\thetaa) := J(\thetaa) + \lambda\lVert \thetaa \rVert_1 \Big \},
\end{equation}
where $\lVert \cdot \rVert_1$ is the $\ell_1$-norm of its argument. 
Figure \ref{fig:std_l1} shows that a sparse solution $\thetaa^*_0$ can be obtained here, where the regularization forces the solution to be on the surface of the lightly (green) shaded hyper-pyramid. 
A Bayesian interpretation of $\mathcal{P}_{1,\lambda}$ can be given using a maximum \textit{a posteriori} (MAP) estimate of the parameters as 
\begin{equation}
\begin{split}
{\thetaa}_\mathrm{MAP} &= \arg\max_{\thetaa} \mathbb{P}(\thetaa|\D),\\
& = \arg\max_{\thetaa} \Big( \ln \mathbb{P}(\D|\thetaa) + \ln \mathbb{P}(\thetaa) \Big), 
\end{split}
\end{equation}
where $\mathbb{P}(\cdot)$ denotes the probability of its argument event and Bayes' theorem is applied to express $\mathbb{P}(\thetaa|\D)$. Next, assuming a Gaussian distributed dataset with standard deviation $\sigma_D$ and independent zero-mean Laplace distributions with scale parameter $\sigma_\theta$ for the prior of each of the parameters in $\thetaa$, the MAP estimate can be written as the solution of $\mathcal{P}_{1,\lambda}$ as follows 
\begin{equation}
{\thetaa}_\mathrm{MAP} = \arg\min_{\thetaa} \Bigg( \frac{1}{\Ntr} \sum_{i=1}^{\Ntr}\Big\lVert\ym_i-\pinn(\xm_i;\thetaa)\Big\rVert_2^2 + \lambda \lVert\thetaa\rVert_1 \Bigg),
\end{equation}
where $\lambda=\frac{{2}\sigma_D^2}{\Ntr \sigma_\theta}$.




In standard $\ell_1$-regularization, a different non-sparse solution $\thetaa^*$ of $\mathcal{P}_{1,\lambda}$ may exist with $\lVert \thetaa^* \rVert >\lVert \thetaa^*_0 \rVert$ as shown in Figure \ref{fig:std_l1}. To avoid ending up with the solution $\thetaa^*$ instead of the sparse solution $\thetaa^*_0$, a weighted $\ell_1$-regularization can be used, which is given by 
\begin{equation}\label{eq:wt_l1}
\mathcal{P}_{1,\lambda}^{(\Wm)} \equiv \Big\{ \arg\min_{\thetaa} R(\thetaa) := J(\thetaa) + \lambda\lVert \Wm\thetaa \rVert_1 \Big \},
\end{equation}
where $\Wm\in\mathbb{R}^{n_{\thetaa}\times n_{\thetaa}}$ is a diagonal weight matrix with elements $w_{ii}>0, ~~ i=1,\dots,n_{\thetaa}$, and $\lVert \Wm \thetaa \rVert_1 = \sum_{i=1}^{n_{\thetaa}}w_{ii}|\theta_i|$. As Figure \ref{fig:wl_1} shows, for some chosen $\Wm$ there can only be one sparse solution of $\mathcal{P}_{1,\lambda}^{(\Wm)}$. 

In this paper, we use different strategies to estimate the neural network parameters $\thetaa$ by solving the optimization problems  $\mathcal{P}_{1,\lambda}$ and $\mathcal{P}_{1,\lambda}^{(\Wm)}$.
%
%
In $\ell_1$-regularization, the objective becomes non-differentiable at zero parameter values. As a result, we use subgradients \footnote{A subgradient for a convex function $f(x):\mathcal{X}\rightarrow\mathcal{Y}$ at $x_0\in\mathcal{X}$ is defined as any $g(x)$ which satisfies $f(x)\geq f(x_0) + \langle g(x), x-x_0 \rangle$ $\forall x\in\mathcal{X}$, where $\langle \cdot,\cdot \rangle$ denotes an inner product. The set of all subgradients are termed as subdifferential $\partial f(x_0)$.} and subgradient methods. A standard stochastic subgradient update at $k$th iteration is given by 
\begin{equation}\label{eq:subgrad}
\thetaa_{k+1} = \thetaa_k - \eta_k {\gm}(\thetaa_k),
\end{equation}
where ${\gm}(\thetaa_k)$ is an unbiased stochastic subgradient\footnote{Here,  $\mathbb{E}[{\gm}(\thetaa_k)|\thetaa_k]\in \partial R(\thetaa_k)$ is a subgradient of the objective $R(\thetaa)$, where $\mathbb{E}[\cdot]$ denotes the mathematical expectation operator and $\partial R(\thetaa_k)$ is the subdifferential of the objective at $k$th iteration.} of $R(\thetaa)$. In this paper, we use the right-hand derivatives\footnote{For a real function $f(x):\mathbb{R}\rightarrow\mathbb{R}$, the right-hand derivative is defined as $\lim\limits_{~~h \to 0^+} \frac { f ({x + h}) - f (x)}{h}$.} as the subgradients when parameter values are zero. The convergence of stochastic subgradient method can be given by the following theorem \cite[Theorem 46]{shor2013nondifferentiable}, \cite[Section 3]{boyd18convex}, which is the stochastic counterpart of Theorem 3.2.2 in Nesterov \cite{nesterov2013introductory}. Note that the objective $R(\thetaa)$ is non-convex in the training of neural networks. However, the convergence result for a convex objective in the following theorem provides insights in the regularization strategies implemented herein.

\newtheorem{thm}{Theorem}

\begin{thm} \label{thm1} (Schor \cite[Theorem 46]{shor2013nondifferentiable} and Boyd and Mutapcic \cite[Section 3]{boyd18convex})
	Let us assume the initial and optimal parameters are given by $\thetaa_0$ and $\thetaa^*$ with $\mathbb{E}\left[\lVert \thetaa_0 - \thetaa^* \rVert^2\right]\leq \rho^2$; the subgradients are bounded by $\mathbb{E}\left[\lVert{\gm}(\thetaa_k)\rVert_2^2\right]\leq \gamma^2~\forall k$; and the learning rates satisfy $\eta_k>0$, $\sum_k \eta_k \rightarrow \infty$, and $\sum_k \eta_k^2 < \infty$. Then, the stochastic subgradient update \eqref{eq:subgrad} for a convex objective function $R(\thetaa)$ satisfies
	\begin{equation}
	\min_{i=0,\dots,k} \mathbb{E}\left[R(\thetaa_i) - R(\thetaa^*)\right] \leq \frac{\rho^2 + \gamma^2 \sum_{i=0}^k \eta_i^2}{2\sum_{i=0}^k \eta_i}. 
	\end{equation}
\end{thm}

\section{Iteratively Reweighted and Bi-fidelity $\ell_1$-regularization Strategies} \label{sec:l1_strategies}

In this paper, we consider three novel $\ell_1$-regularization strategies to train neural network surrogates using only high-fidelity and both low- and high-fidelity data. The performance of these methods will be compared against that of the standard $\ell_1$-regularization (Strategy I) using high-fidelity data denoted as $\Dh$. One of these strategies (Strategy II) uses only high-fidelity data, and the other two strategies (Strategy III and Strategy IV) use low-fidelity data, referred to as $\Dl$, to inform the training using $\Dh$. Further, two of these strategies implement weighted $\ell_1$-regularization to promote sparsity in the network parameters as standard $\ell_1$-norm is often inadequate due to its fast shrinkage property and lack of scale invariance \cite{yang2019deephoyer,hoefler2021sparsity}. Herein, we make the practical assumptions that (i) the size of $\Dl$ is {\it large enough} to accurately train the low-fidelity neural network at hand and (ii) in contrast, since generating high-fidelity data is expensive, the size of $\Dh$ is small so that training only using $\Dh$ leads to poor generalization errors. In all strategies, we estimate the regularization strength $\lambda$ using a grid-based approach as explained in Section \ref{sec:ex}.  

\subsection{Strategy I: Standard $\ell_1$-Regularization Using High-fidelity Data}

The first strategy that we employ for the training of neural networks is the solution of $\mathcal{P}_{1,\lambda}$ in \eqref{eq:std_l1} in the presence of a high-fidelity training dataset $\Dh$. This approach is considered as the baseline to compare against the other three approaches. 

\subsection{Strategy II: Iteratively Reweighted $\ell_1$-Regularization Using High-fidelity Data}

In our second strategy, we use the weighted $\ell_1$-regularization problem $\mathcal{P}_{1,\lambda}^{(\Wm)}$. A heuristic approach to set the weights is to assign small values to those associated with parameters that are expected to be large and vice versa. 
For example, the diagonal elements of the weight matrix $\Wm$ can be set as 
\begin{equation}
w_{ii} = \begin{cases}
|\theta_i|^{-1}, \quad \theta_i \neq 0;\\
\infty, \quad \quad ~~ \theta_i=0;\quad i=1,\dots,n_{\theta}.\\
\end{cases}
\end{equation}
However, this requires knowing the parameter values. Instead, motivated by techniques from compressed sensing and variable selection~\cite{candes2008enhancing,zou2006adaptive}, an approximation of $\thetaa$ from the previous iterations is used in this strategy\footnote{In the first numerical example in Section \ref{sec:ex}, we did not see any improvements in the presence of a small $\Dh$ when we train the network using Strategy II after initially training the network using Strategy I.}. For stability, we use a damped version of the weights at $k$th iteration as follows
\begin{equation}
w_{ii,k} = \left( \Big|\theta_{k-1,i}\Big| + \epsilon_w \right)^{-1};\quad i=1,\dots,n_{\theta},
\end{equation}
where $\theta_{k-1,i}$ is the parameter from $(k-1)$th iteration and $\epsilon_w$ is a small number\footnote{To simplify the notation, we use both parameter or layer index and iteration number in the subscript, which can be differentiated from the context. }. We use $\epsilon_w=10^{-5}$ for all numerical examples of this study.


\subsection{Strategy III: Standard $\ell_1$-Regularization Using Bi-fidelity Data}

In the bi-fidelity approaches discussed in this work, we use two identical networks --- a low-fidelity network $\pinn_{l}$ trained with $\Dl$ and a high-fidelity network $\pinn_{h}$ trained with $\Dh$. 
Training the network $\pinn_{l}$ using the low-fidelity dataset $\Dl$ and a standard $\ell_1$-regularization similar to Strategy I results in the network parameters $\thetaa_\mathrm{LF}$. In our first bi-fidelity training strategy, we assume the parameters $\thetaa$ of the high-fidelity network $\pinn_{h}$ follows Laplace distribution with mean $\thetaa_\mathrm{LF}$. The modified optimization problem for training of the high-fidelity network $\pinn_{h}$ is given by
\begin{equation} \label{eq:P_lambda_BF}
\mathcal{P}_{1,\lambda}^{\mathrm{(BF)}} \equiv \Big\{ \arg\min_{\thetaa} R(\thetaa):= J(\thetaa) + \lambda\lVert \thetaa - \thetaa_\mathrm{LF} \rVert_1 \Big \}.
\end{equation}
Hence, this strategy assumes that most of the parameters in the high-fidelity network $\pinn_h$ are similar to the low-fidelity network parameters $\thetaa_\mathrm{LF}$. This is a variation of the bi-fidelity transfer learning scenario used in De et al. \cite{de2020transfer}, where the high-fidelity network $\pinn_{h}$ keeps the parameter values from the low-fidelity network $\pinn_l$ for most of the layers. However, instead of keeping the parameters of the first few layers fixed after training using the low-fidelity data, herein we use regularization to induce small variations in the high-fidelity network parameters. Depending on the choice of $\lambda$, such a variation may occur at a small subset of the entries of $\bm\theta$ distributed over the entire network. 

\subsection{Strategy IV: Weighted $\ell_1$-Regularization Using Bi-fidelity Data}

In the second bi-fidelity approach, we use the weighted $\ell_1$-regularization  as described in Section \ref{sec:l1_background}. 
Here, the weight matrix $\Wm$ is diagonal with entries constructed from the parameters $\thetaa_\mathrm{LF}$ of the low-fidelity network $\pinn_l$ as follows
\begin{equation}
w_{ii} = \left( |\theta_{\mathrm{LF},i}^{}| + \epsilon_w \right)^{-1}; \quad i=1,\dots,n_{\theta},
\end{equation}
where $\epsilon_w$ is a small number which we set to $10^{-5}$ in our numerical experiments. The optimization problem is similar to $\mathcal{P}_{1,\lambda}^{(\Wm)}$ in \eqref{eq:wt_l1}. Again, in this strategy we encourage large parameters from the low-fidelity network to remain large in the high-fidelity network and vice versa. Hence, the two bi-fidelity Strategies III and IV are similar to the standard transfer learning using bi-fidelity datasets discussed in De et al. \cite{de2020transfer}. However, these two strategies allow changes in the parameter values across all layers instead of a few hidden layers as the case with transfer learning and hence Strategies III and IV here can be considered as generalizations of transfer learning. In addition, the Strategy IV may be considered a generalization of a pruning approach wherein the entries of $\bm\theta$ are set to zero, i.e., hard-thresholded, based on their low-fidelity counterparts in $\thetaa_\mathrm{LF}$.


\textbf{Remark:} The networks trained with $\ell_1$-regularization strategies of this work may be further pruned by removing the connections with magnitude of the parameters smaller than a prescribed threshold to arrive at a {\it truly} sparse network. However, this post-processing step is not performed in the numerical examples of this study.

\subsection{Bounds on the Generalization Error} 
The hypothesis space of the neural network $\widehat{\mathcal{M}}$, i.e., the set of all possible neural networks that have the same number of hidden layers and neurons per hidden layer is denoted by $\mathcal{H}:=\mathcal{H}_{\mathrm{inp}}\cup\left(\bigcup_{j=1}^{N_H}\mathcal{H}_j\right)\cup \mathcal{H}_{0}$, where $\mathcal{H}_\mathrm{inp}$, $\{\mathcal{H}_j\}_{j=1}^{N_H}$, and $\mathcal{H}_0$ are the hypothesis spaces of the input layer, $N_H$ hidden layers, and output layer, respectively. 
For the hidden layers and the output layer we have
\begin{equation}\label{eq:nn_hyp}
\begin{split}
\mathcal{H}_j&:=\Big\{ \xm \mapsto \sum_i \psi_{j,i} \sigma(f_i(\xm)) + \beta_j: f_i(\xm)\in\mathcal{H}_{j-1}, \lVert \thetaa_j \rVert_1\leq {L}_j \Big\},\quad j=1,\dots,N_H;\\
\mathcal{H}_0&:=\Big\{ \xm \mapsto \sum_i \psi_{0,i} f_i(\xm) + \beta_0: f_i(\xm)\in\mathcal{H}_{N_H}, \lVert \thetaa_0 \rVert_1\leq {L}_0 \Big\},
\end{split}
\end{equation}
where $L_j$, $j=0,\dots,N_H$ are the upper limits on $\ell_1$-norm of the parameters in the hidden and the output layer, which depend on the regularization strength $\lambda$. In (\ref{eq:nn_hyp}), $\psi_{j,i}$ is the $i$th weight in the $j$th hidden layer. The generalization error of a neural network $\widehat{\mathcal{M}}$ is defined next as
\begin{equation}\label{eq:gen_err}
\mathcal{L}_{\mathcal{S}}({\mathcal{F}},\mathcal{D}) := \sup_{\widehat{\mathcal{M}}\in\mathcal{H}} \left( \mathbb{E}_{\mathcal{D}^\prime\sim\mathcal{S}}\left[ J\left(\widehat{\mathcal{M}};{\mathcal{D}^\prime}\right) \right] -  \frac{1}{N} \sum_{i=1}^{N} J\left(\widehat{\mathcal{M}};\mathcal{D}_i\right) \right),
\end{equation}
where $\mathcal{S}$ is the probability measure of the data; $\mathcal{F}$ is the function composition of the hypothesis space and the loss function, i.e., $\mathcal{F}:=J\circ\mathcal{H}:=\{\mathcal{D}\rightarrow J(h;\mathcal{D}):h\in\mathcal{H}\}$; and the expectation is with respect to $\mathcal{S}$. 

\newtheorem{prop}{Proposition}
\begin{prop}\label{prop} 
	For the four neural network training strategies used in this paper, the expected value of the generalization error defined in \eqref{eq:gen_err} can be bounded by
	\begin{equation} \label{eq:gen_err_bound}
	\mathbb{E}_{\mathcal{D}\sim \mathcal{S}}[\mathcal{L}_{\mathcal{S}}({\mathcal{F}},\mathcal{D})] \leq 2 ~K~ \mathbb{E}_{\mathcal{D}\sim \mathcal{S}}\left[ \mathcal{R}_{\mathcal{D}}(\mathcal{H}_\mathrm{inp}) \right],
	\end{equation}
	where $\mathcal{R}_{\mathcal{D}}(\mathcal{H}_\mathrm{inp})$ is the empirical Rademacher complexity of the input layer \footnote{Empirical Rademacher complexity of a family of real-valued function $\mathcal{H}$ defined on a domain $\mathcal{X}$ for a sample $\xm=\{x_i\}_{i=1}^N$ with $x_i\in\mathcal{X}$ is defined as $\mathcal{R}_{\mathcal{D}}(\mathcal{H}):=\frac{1}{N}\mathbb{E}_{\mathbf{\mathcal{E}}}\left[ \sup_{h\in\mathcal{H}} \sum_{i=1}^N\epsilon_i h(x_i) \right]$, where $\epsilon_i\sim\mathcal{E}$, $i=1,\dots,N$, are i.i.d. Rademacher random variables, i.e., $\mathbb{P}[\epsilon_i=+1]=\mathbb{P}[\epsilon_i=-1]=0.5$. }. 
	The constant $K$ for the four different strategies are given by 
	\begin{equation} \label{eq:K}
	\begin{split}
	K_\mathrm{std}^{\mathrm{HF}} & = \prod_{j=0}^{N_H} 2L_{j};\\
	K_\mathrm{wgt}^{\mathrm{HF}} & = \prod_{j=0}^{N_H} 2L_{j} \leq \prod_{j=0}^{N_H} 2L_{\mathrm{w},j}\left( \Big\lvert \thetaa_{k-1,j}^{\mathrm{max}}\Big\rvert + \epsilon_w \right);\\
	K_\mathrm{std}^{\mathrm{BF}} & = \prod_{j=0}^{N_H} 2L_{j} \leq \prod_{j=0}^{N_H} 2\left( L_{\mathrm{d},j} + L_{\mathrm{LF},j} \right);\\
	K_\mathrm{wgt}^{\mathrm{BF}} & = \prod_{j=0}^{N_H} 2L_{j} \leq \prod_{j=0}^{N_H} 2L_{\mathrm{w},j}\left(\Big\lvert \thetaa_{\mathrm{LF},j}^{\mathrm{max}}\Big\rvert + \epsilon_w \right),\\
	\end{split}
	\end{equation}
	where $K_\mathrm{std}^{\mathrm{HF}}$ is for standard high-fidelity $\ell_1$-regularization (Strategy I); $K_\mathrm{wgt}^{\mathrm{HF}}$ is for iteratively reweighted high-fidelity $\ell_1$-regularization (Strategy II); $K_\mathrm{std}^{\mathrm{BF}}$ is for standard bi-fidelity $\ell_1$-regularization (Strategy III); and $K_\mathrm{wgt}^{\mathrm{BF}}$ is for weighted bi-fidelity $\ell_1$-regularization (Strategy IV); $\lVert \thetaa_j \rVert_1 \leq L_{j}$; $\lVert \Wm_j\thetaa_j \rVert_1 \leq L_{\mathrm{w},j}$; $\lVert \thetaa_j - \thetaa_{\mathrm{LF},j} \rVert_1 \leq L_{\mathrm{d},j}$; $\lVert \thetaa_{\mathrm{LF},j} \rVert_1 \leq L_{\mathrm{LF},j}$; $|\thetaa_{k-1,j}^{\mathrm{max}}|$ is the absolute maximum value of the $j$th layer parameter at $(k-1)$th iteration; and $|\thetaa_{\mathrm{LF},j}^{\mathrm{max}}|$ is the absolute maximum value of the $j$th layer parameter of the low-fidelity network. 
	
	Further, if the subgradient method in \eqref{eq:subgrad} is used for a convex objective $R(\thetaa)$ for $k$ iterations, $L_j$, $L_{\mathrm{w},j}$, and $L_{\mathrm{d},j}$, $j=0,1,\dots,N_H$, in (\ref{eq:K}) are bounded as follows
	\begin{equation} \label{eq:Lj_inequals}
	\begin{split}
	\textbf{Strategy I:\phantom{I and IV}} \qquad &\mathbb{E} \left[ \lVert \thetaa_j \rVert_1 \right] \leq ~~L_{j}~  \leq C + \lVert \thetaa^* \rVert_1 -\frac{\mathcal{E}_\mathrm{tr}}{\lambda}; \\
	\textbf{Strategies II and IV:  } ~\quad &\mathbb{E} \left[ \lVert \Wm_j\thetaa_j \rVert_1 \right] = \sum_{i=1}^{n_{\thetaa_j}} \left(\mathrm{Cov}(w^j_{ii},|\theta_{j,i}|) + \mathbb{E}[w^j_{ii}] \mathbb{E}[|\theta_{j,i}|]\right)\\
	& \qquad \qquad \quad ~~ \leq L_{\mathrm{w},j}  \leq C + \mathbb{E}\left[\lVert \Wm\thetaa^* \rVert_1\right] -\frac{\mathcal{E}_\mathrm{tr}}{\lambda}; \\
	\textbf{Strategy III:\phantom{and IV}} \qquad &\mathbb{E} \left[ \lVert \thetaa_j - \thetaa_{\mathrm{LF},j} \rVert_1 \right]   \leq L_{\mathrm{d},j}  \leq C + \mathbb{E} \left[ \lVert \thetaa_\mathrm{LF} - \thetaa^* \rVert_1 \right] -\frac{\mathcal{E}_\mathrm{tr}}{\lambda}. \\
	\end{split}
	\end{equation}
	Here, $C=\frac{\rho^2 + \gamma^2 \sum_{i=1}^k \eta_i^2}{2\lambda\sum_{i=1}^k \eta_i}$ using Theorem \ref{thm1}; $\mathcal{E}_\mathrm{tr} = \mathbb{E}[J(\thetaa_{k^*}) - J(\thetaa^*)]$ is the best training error as we select the parameters from $k^*$th iteration; $\thetaa_j$ is the selected parameters of $j$th layer after the end of optimization; and $\text{Cov}(\cdot,\cdot)$ denotes the covariance of its argument. The optimal value of the parameters are $\thetaa^*$, and the expectations are taken with respect to $\thetaa_j$ and $\thetaa_{\mathrm{LF},j}$ or $\thetaa^{(k-1)}_j$ depending on the strategy\footnote{The randomness in the network parameters comes from the fact that the subgradients estimated during training are stochastic in nature.}. 
\end{prop}
The proof of this proposition is provided in Appendix \ref{sec:lemmas}. 
From the above proposition the following observations can be made:

(i) From \eqref{eq:K}, for high-fidelity iteratively reweighted Strategy II the bound will be tighter if the absolute maximum value of the parameter from the previous iteration is small. Similarly, for bi-fidelity weighted Strategy IV a tighter bound is obtained if the absolute maximum value of the parameter from the low-fidelity network is small. 

(ii) From \eqref{eq:K}, for  bi-fidelity Strategy III, we obtain a tighter bound if the high-fidelity network is very similar to the low-fidelity one, i.e., smaller $L_{\mathrm{d},j}$, $j=0,\dots,N_H$. 

(iii) From \eqref{eq:Lj_inequals}, for a very small training error $\mathcal{E}_\mathrm{tr}$, the upper bound on $L_j$, $L_{\mathrm{w},j}$, and $L_{\mathrm{d},j}$ can be large. As a result, $K$ in \eqref{eq:gen_err_bound} can be large and therefore (\ref{eq:gen_err_bound}) leads to a loose bound.  This shows the effect of overfitting. The opposite happens for large training error providing a tighter bound. However, we cannot have a large $\mathcal{E}_\mathrm{tr}$ to provide $L_j$, $L_{\mathrm{w},j}$, and $L_{\mathrm{d},j}$ as zero for all the layers in the network since a stochastic counterpart of Theorem 3.2.1 in Nesterov \cite{nesterov2013introductory} will introduce a positive lower bound for these terms. 

(iv) From \eqref{eq:Lj_inequals}, for the weighted strategies II and IV, if the covariance between each weight and the absolute value of the corresponding parameter is negative, we can select a smaller $L_{\mathrm{w},j}$ to achieve a tighter bound in \eqref{eq:gen_err_bound}. In particular, if $\thetaa_{k-1}$ and $\thetaa_\mathrm{LF}$ are correlated with $\thetaa$, we have negative covariance since the weights are inversely proportional to $\vert\thetaa_{k-1}\vert$ and $\vert\thetaa_\mathrm{LF}\vert$. 

(v) Further assume the neural network uses only ReLU activations with at least $\lceil c_1 \left(\ln(1/\epsilon)+1\right) \rceil$ number of layers and same number of neurons per layer and approximates a function from the unit ball $\mathcal{B}_{d,n} = \{f\in\mathcal{W}^{n,\infty}([0,1]^d): ||f||_{\mathcal{W}^{n,\infty}([0,1]^d)}\leq 1\}$ in the Sobolev space $\mathcal{W}^{n,\infty}([0,1]^d)$ up to an accuracy of $\epsilon$ in the $L_\infty$ norm \footnote{In other words, the unit ball $\mathcal{B}_{d,n}$ denotes a family of functions that have $d$-dimensional input in $[0,1]^d$ and up to $n$th order weak derivative with $L_\infty$ norm less than one.}. 
Then, using Theorem 1 in Yarotsky \cite{yarotsky2017error}, $L_j \leq |\thetaa_{j}^{\max}|\epsilon^{-d/n}$ and $L_{\mathrm{w},j}, L_{\mathrm{d},j} \leq c_2 \epsilon^{-d/n}$, where $c_1$ is a constant that depends on $d$ and $n$. The constant $c_2$ is close to one for the weighted strategies II and IV if  $\thetaa_{k-1,j}$ or $\thetaa_{\mathrm{LF},j}$ is close to $\thetaa_j$, and $c_2$ is close to zero for the standard bi-fidelity Strategy III if $\thetaa_{\mathrm{LF},j}$ is close to $\thetaa_j$. This shows that when approximating a function with a large $n$ (i.e., with bounded higher order derivatives) using a ReLU network then the generalization error bound becomes tight and vice versa.

\textbf{Remark:} Note that the bounds in \eqref{eq:Lj_inequals} depend on the optimal values of $\thetaa$, which is unknown in practice. Additionally, we use Adam for the subgradient method in our numerical examples instead of the standard subgradient update in \eqref{eq:subgrad}. However, results in \eqref{eq:K} and \eqref{eq:Lj_inequals} help understanding different aspects of the $\ell_1$-regularization strategies discussed in this paper and show the effectiveness of $\ell_1$-regularized training in terms of generalization error. 
In practice, once a network is trained using any of the four strategies, we can estimate $K$ from its parameters. During training, we can estimate $|\thetaa_{k-1,j}^{\max}|$ from the previous iteration, or $|\thetaa_{\mathrm{LF},j}^{\max}|$ and $L_{\mathrm{LF},j}$ from an already trained low-fidelity network to judge the effectiveness of these strategies. 


\section{Numerical Examples}\label{sec:ex} 

The four strategies for training neural networks using $\ell_1$-regularization are compared in this section with three numerical examples. The first two examples use two FNN networks to predict a QoI, whereas we choose the third example to be the training of an autoencoder. We use two datasets, namely, a training dataset $\D_{\mathrm{tr},h}$ and a validation dataset $\Dval$ from the high-fidelity model. In the bi-fidelity strategies, we also use a training dataset $\D_{\mathrm{tr},l}$ from the low-fidelity model. The datasets $\D_{\mathrm{tr},h}$ and $\D_{\mathrm{tr},l}$ are used to estimate the gradients during training of the neural networks. The accuracy of the network predictions are measured for the high-fidelity validation dataset $\Dval$ using a relative root mean squared error (RMSE), estimated as $\varepsilon_{\mathrm{v}}= \frac{\lVert \ym_{\mathrm{val}} - \ym_{\mathrm{pred}} \rVert_2}{\lVert \ym_{\mathrm{val}} \rVert_2}$, where $\{\ym_{\mathrm{val},j}\}_{j=1}^{\mathrm{val}}$ are the QoIs in the validation dataset and $\{\ym_{\mathrm{pred},j}\}_{j=1}^{\mathrm{val}}$ are the corresponding predictions from the neural networks. We choose configurations of the neural network by observing $\varepsilon_{\mathrm{v}}$ and then selecting the one that corresponds to the smallest error. For example, we increase the number of neurons per hidden layer until we reach a pre-fixed number (200 herein) and then the number of hidden layers by one, as long as we see a reduction in $\varepsilon_{\mathrm{v}}$ during pilot runs. Furthermore, we choose the regularization strength $\lambda$ for the $\ell_1$-regularization using a grid-based approach in the logarithmic scale with base 10, i.e., we measure $\varepsilon_{\mathrm{v}}$ for $\lambda=10^{-3},10^{-2},\dots$ and then choose the value of $\lambda$ that produces the smallest error. Also, we use 10 different initializations for each instance of the training/validation dataset and report the best errors for the numerical examples.

\subsection{Example I: Deflection of a Composite Beam}
\begin{figure}[!htb]
	\centering
	\includegraphics[scale=1.5]{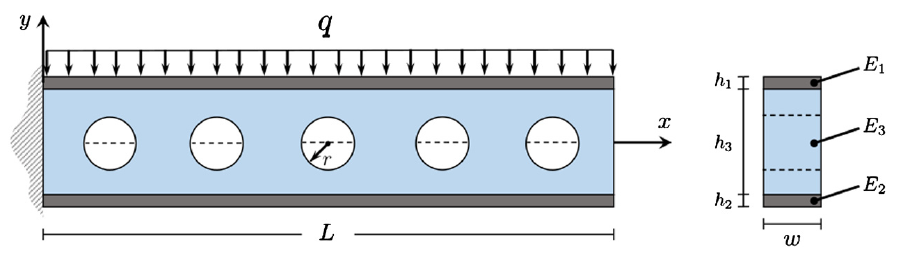}
	\caption{The composite beam used in Example I (adapted from \cite{hampton2018practical}). The dimensions are as follows: $L=50$\,m, $r=1.5$\,m, $h_1=h_2=0.1$\,m, and $h_3=5$\,m. The elastic moduli $E_1$, $E_2$, $E_3$, and the external load $q$ are assumed uncertain.}
	\label{fig:ex1}
\end{figure}

\begin{figure}[!htb]
	\centering
	\includegraphics[scale=1.5]{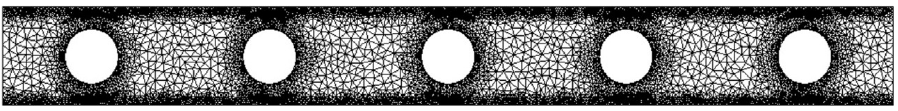}
	\caption{The composite beam is solved in FEniCS with the mesh shown here and used as the high-fidelity model.}
	\label{fig:ex1_hf}
\end{figure}

In our first example, we use a cantilever beam subjected to a uniformly distributed load $q$ with composite cross-section as shown in Figure \ref{fig:ex1} in our first example. The length and width of the beam are $L=50$ m and $w=1$ m, respectively. The composite cross-section has three different parts with heights $h_1=h_2=0.1$ m, and $h_3=5$ m, and elastic moduli $E_1$, $E_2$, and $E_3$, respectively. The web of the beam also has five circular holes with radius $r=1.5$ m. The magnitude of the distributed load $q$ and the elastic moduli $E_1$, $E_2$, and $E_3$ are assumed uncertain with their probability distributions given in Table \ref{tab:beam_unc}. For the neural network, we use $\xm = [q,E_1, E_2, E_3, E_4]^T$ as the input and maximum deflection at the free end as the output $y$. We generate the high-fidelity dataset $\Dh$ using a finite element mesh shown in Figure \ref{fig:ex1_hf} and solved in FEniCS \cite{LoggMardalEtAl2012a}. For the low-fidelity dataset $\Dl$, we use the Euler-Bernoulli beam theory ignoring the circular holes, which is given by
\begin{equation}\label{eq:beam}
\begin{split}
& EI\frac{\mathrm{d}^4u_l(x)}{\mathrm{d}x^4} = -q,\\
& \text{subject to }u_l(0) = 0;~ \frac{du_l(0)}{dx} = 0; ~ \frac{d^3u_l(L)}{dx^3}=0;~\frac{d^4u_l(L)}{dx^4}=0,\\
\end{split}
\end{equation}
where $E$ is the elastic modulus and $I$ is the moment of inertia of an {\it equivalent} cross-section consisting of a homogenized material. The solution of \eqref{eq:beam} is given by 
\begin{equation}\label{eq:beam_lf}
u_l(x) = -\frac{qL^4}{24EI}\left[ \left(\frac{x}{L}\right)^4 - 4\left(\frac{x}{L}\right)^3 + 6\left(\frac{x}{L}\right)^2 \right],
\end{equation}
which shows that the uncertainty enters in the solution \eqref{eq:beam_lf} as a multiplicative factor. However, this is not the case when we solve the high-fidelity model in FEniCS. 

\begin{table}[!htb]
	\caption{The uncertain parameters in Example I and their corresponding probability distributions. Note that $\mathcal{U}[a,b]$ denotes a uniform distribution between $a$ and $b$.}
	\label{tab:beam_unc}
	\centering
	\begin{tabular}{l|c|c}
		\hline \rule{0pt}{2ex}Parameter  & Unit & Distribution \\
		\hline
		\rule{0pt}{2ex}Distributed load, $q$ & kN/m & $\mathcal{U}[9,11]$\\
		Elastic modulus, $E_1$ & MPa & $\mathcal{U}[0.9, 1.1]$\\
		Elastic modulus, $E_2$ & MPa & $\mathcal{U}[0.9, 1.1]$\\
		Elastic modulus, $E_3$ & kPa & $\mathcal{U}[9, 11]$ \\\hline
	\end{tabular}
\end{table}

\subsubsection{Results} 

The high-fidelity training dataset  $\D_{\mathrm{tr},h}$ and validation dataset $\Dval$ consist of three and 50 realizations of the uncertain parameters, respectively, from their respective distributions in Table \ref{tab:beam_unc} and corresponding deflections at the free end. 
For the bi-fidelity strategies, we use a low-fidelity dataset $\D_{\mathrm{tr},l}$ that consists of deflections at the free end generated using \eqref{eq:beam_lf} for 250 realizations of the uncertain parameters.
An FNN with two hidden layers with 20 neurons each and ELU activation function is used in this example. 
We use Adam (see Appendix \ref{sec:adam}) with a learning rate of $10^{-4}$ to train the neural network. At the end of 30,000 iterations, we select the network configuration that produces the smallest $\varepsilon_{\mathrm{v}}$. Note that this is equivalent to the commonly used early stopping criterion \cite{bengio2012practical} with infinite \textit{patience} subject to a maximum of 30,000 iterations. The low-fidelity network $\pinn_l$ is trained using a regularization strength of $\lambda=0.01$ chosen using the grid-based search. 

\begin{figure}[!htb]
	\centering
	\begin{subfigure}[!htb]{0.48\textwidth}
		\centering
		\includegraphics[scale=0.25]{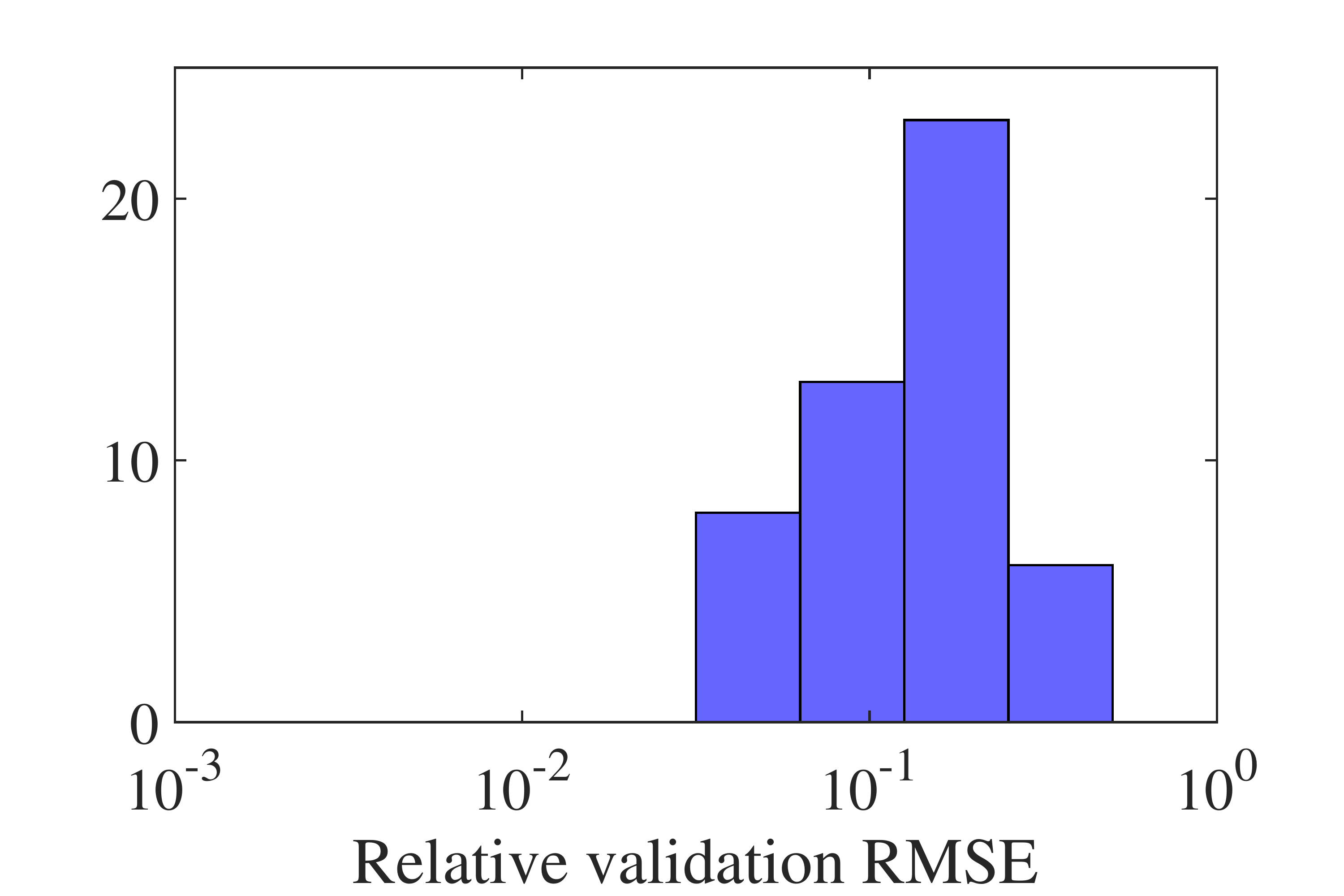}
		\caption{High-fidelity with no regularization}
	\end{subfigure}
	\hfill
	\begin{subfigure}[!htb]{0.48\textwidth}
		\centering
		\includegraphics[scale=0.25]{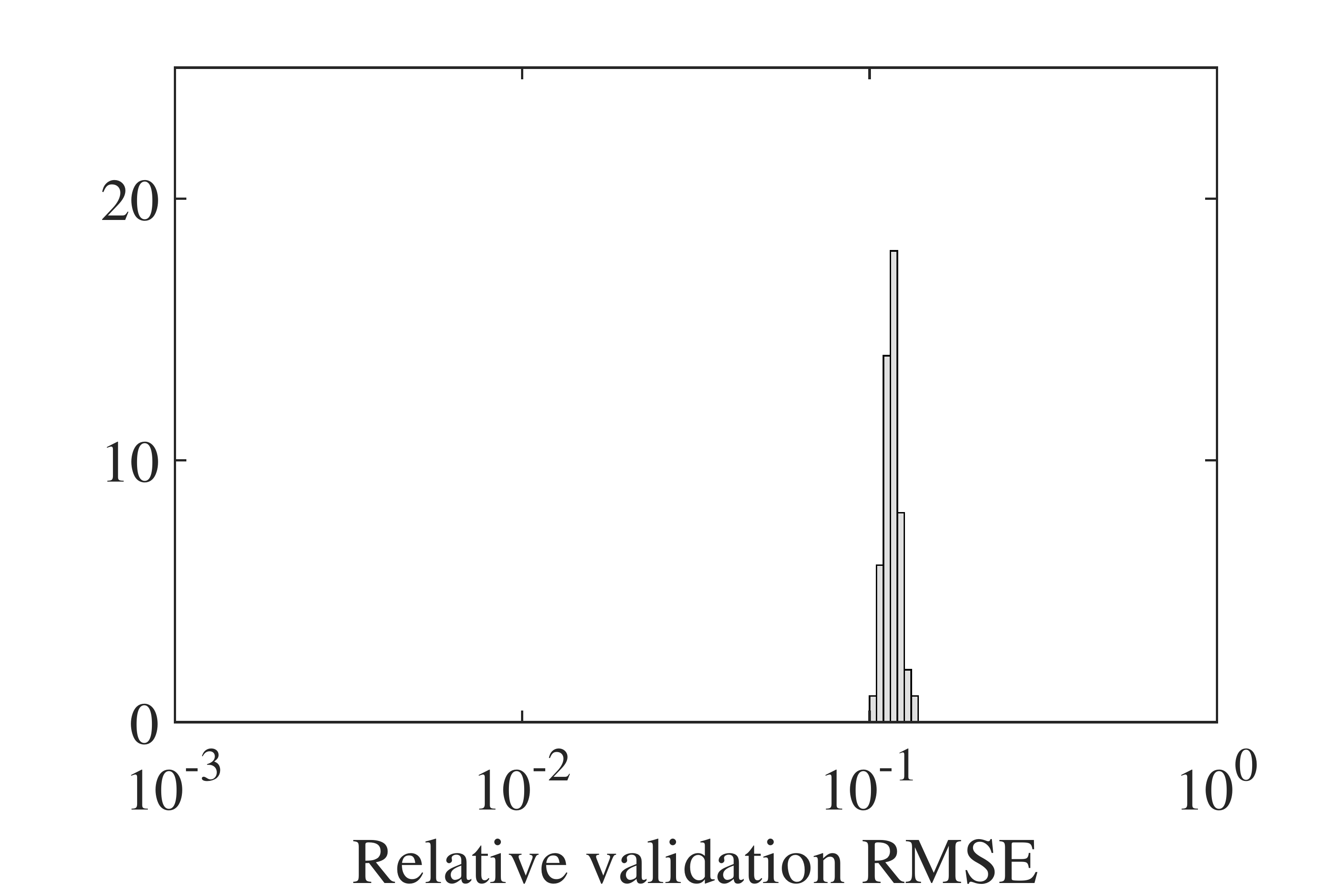}
		\caption{High-fidelity with dropout ($p=0.6$)}
	\end{subfigure}\\
	\begin{subfigure}[!htb]{0.48\textwidth}
		\centering
		\includegraphics[scale=0.25]{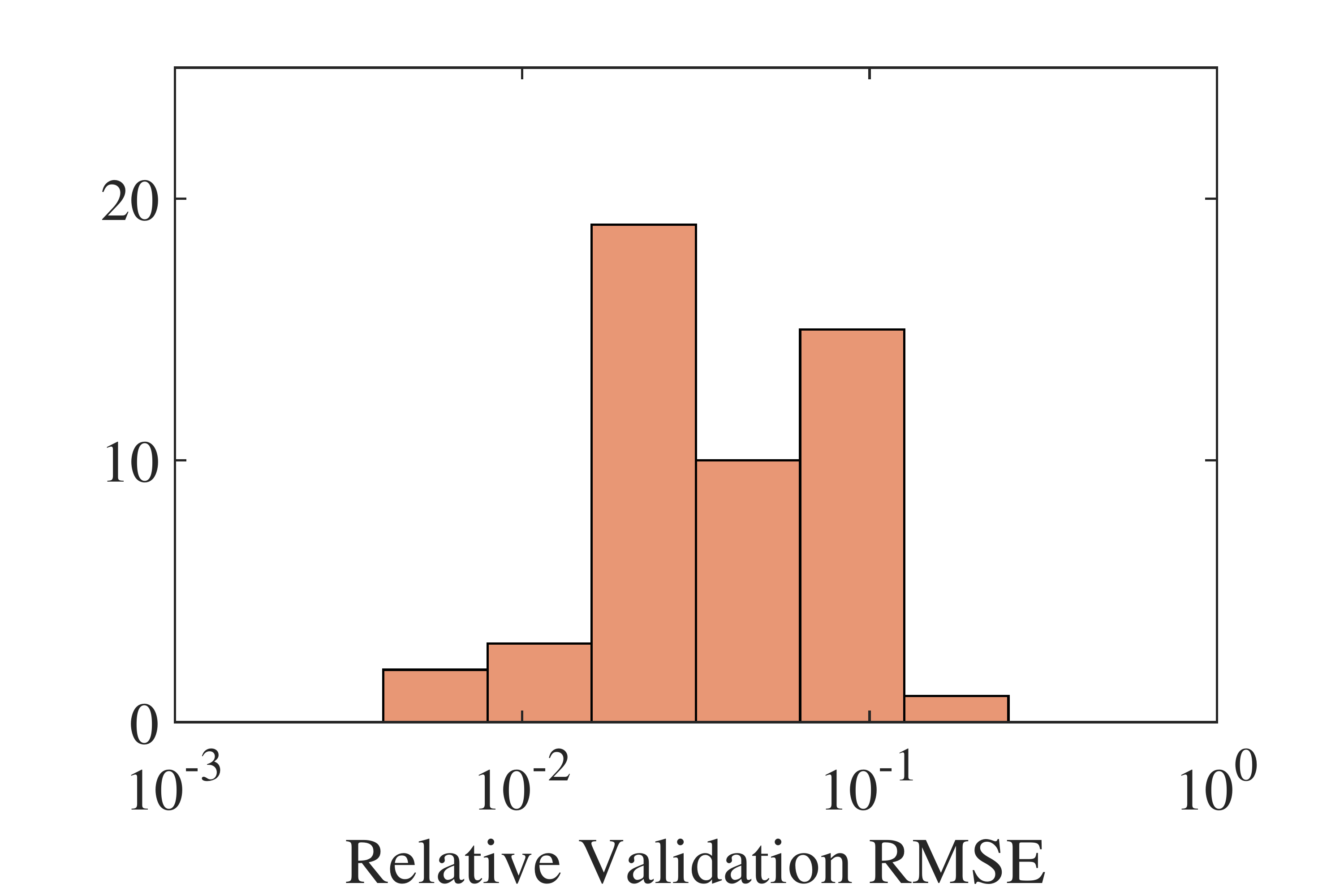}
		\caption{Strategy I: High-fidelity standard $\ell_1$-regularization}
	\end{subfigure}
	\hfill
	\begin{subfigure}[!htb]{0.48\textwidth}
		\centering
		\includegraphics[scale=0.25]{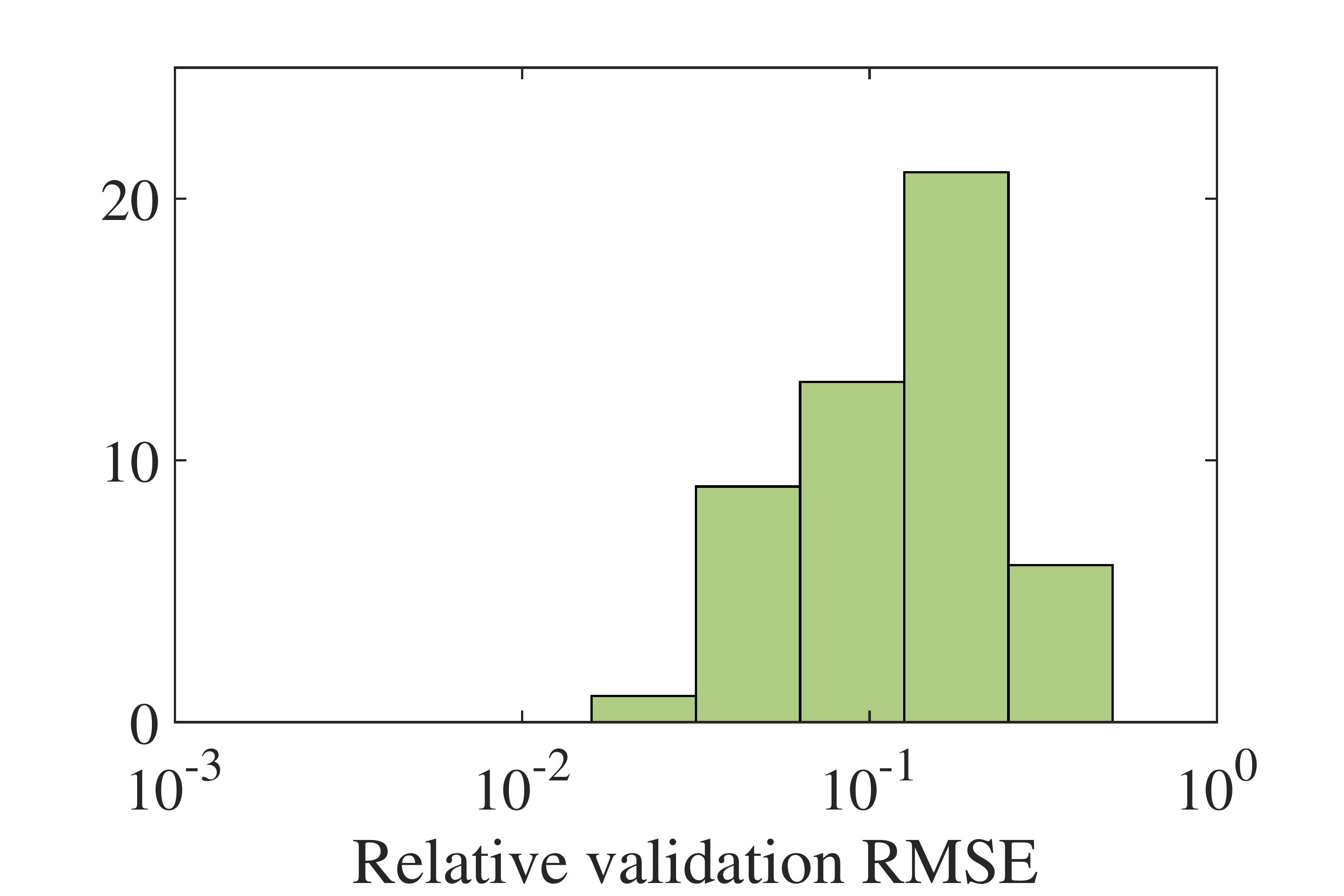}
		\caption{Strategy II: High-fidelity iteratively reweighted $\ell_1$-regularization}
	\end{subfigure}\\
	\begin{subfigure}[!htb]{0.48\textwidth}
		\centering
		\includegraphics[scale=0.25]{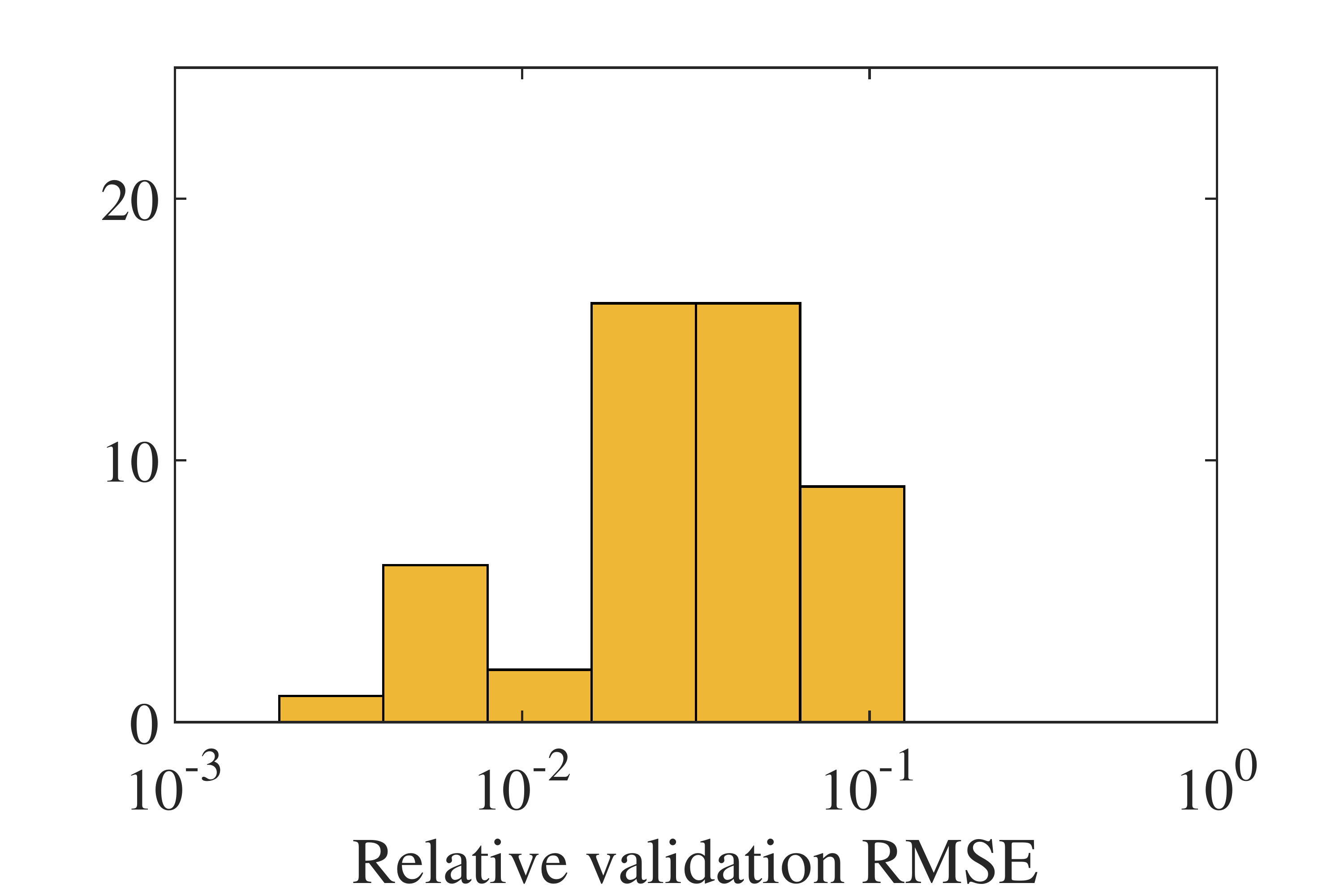}
		\caption{Strategy III: Bi-fidelity standard $\ell_1$-regularization}
	\end{subfigure}%
	\hfill
	\begin{subfigure}[!htb]{0.48\textwidth}
		\centering
		\includegraphics[scale=0.25]{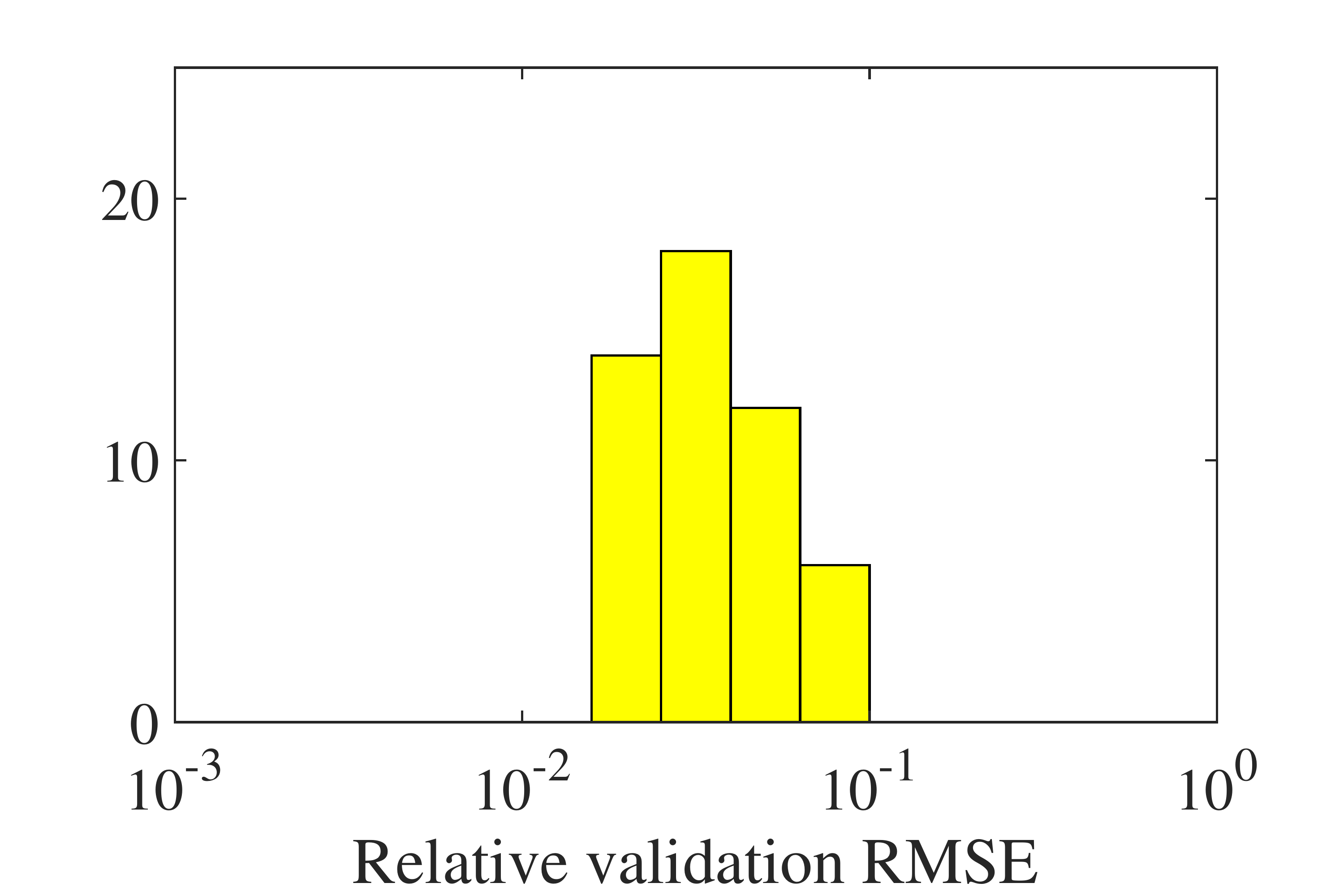}
		\caption{Strategy IV: Bi-fidelity weighted $\ell_1$-regularization}
	\end{subfigure}
	\caption{Histograms of relative validation RMSE, $\varepsilon_{\mathrm{v}}$, for 50 different replications of training/validation datasets for no regularization (with or without dropout) and four different $\ell_1$-regularization strategies in Example I. The high-fidelity training dataset  $\D_{\mathrm{tr},h}$ and the low-fidelity training dataset  $\D_{\mathrm{tr},l}$ consist of three and 250 realizations of the uncertain parameters, respectively. } \label{fig:beam_hist} 
\end{figure}

Figure \ref{fig:beam_hist} shows histograms of $\varepsilon_{\mathrm{v}}$ resulting from different $\ell_1$-regularization strategies for 50 replications of training and validation datasets. The figure shows that using Strategy I, $\varepsilon_{\mathrm{v}}$ is reduced for most of the cases compared to the case in which no regularization is applied. When we use Strategy II the improvement in $\varepsilon_{\mathrm{v}}$ is marginal compared to Strategy I. The bi-fidelity strategies III and IV, on the other hand, significantly improve $\varepsilon_{\mathrm{v}}$. However, the standard bi-fidelity Strategy III shows more variation in $\varepsilon_{\mathrm{v}}$. This is due to heavy reliance on the low-fidelity network parameters $\thetaa_\mathrm{LF}$ as we resist changes from them through $\ell_1$-regularization (see \eqref{eq:P_lambda_BF}).

Table \ref{tab:beam_rmse} shows the mean and standard deviation of $\varepsilon_{\mathrm{v}}$ for the four $\ell_1$-regularization strategies along with a case where no regularization has been implemented. From the table we can conclude that the bi-fidelity Strategies III and IV improve the mean RMSE by one order of magnitude compared to the case where no such regularization has been applied. The standard deviation of $\varepsilon_{\mathrm{v}}$ is also improved significantly by using strategies III-IV. Hence, in this example, a bi-fidelity strategy is advantageous for training of neural networks  if the amount of available high-fidelity data is limited compared to abundant low-fidelity data. Estimating the constants $K$ in \eqref{eq:K} from the trained networks for an instance of the training/validation dataset we obtain 
$K_\mathrm{std}^{\mathrm{HF}}= 535.36$, $K_\mathrm{wgt}^{\mathrm{HF}}=2.15\times10^5$, $K_\mathrm{std}^{\mathrm{BF}}=33.92$, and $K_\mathrm{wgt}^{\mathrm{BF}}=32.72$.
This demonstrates that the bound on the generalization error is tighter for the bi-fidelity training strategies making them better suited for this case, which is evident from Figure \ref{fig:beam_hist} and Table \ref{tab:beam_rmse}. Also, these bounds show that for a small $\D_{\mathrm{tr},h}$, among the high-fidelity strategies, Strategy I has smaller generalization error bound than Strategy II.


\begin{table}[!htb]
	\caption{The mean and standard deviation of relative validation RMSE, $\varepsilon_{\mathrm{v}}$, using different $\ell_1$-regularization strategies  and dropout with a probability $p=0.60$ in Example I. The high-fidelity training dataset  $\D_{\mathrm{tr},h}$ and the low-fidelity training dataset  $\D_{\mathrm{tr},l}$ consist of three and 250 realizations of the uncertain parameters, respectively. }
	\label{tab:beam_rmse}
	\centering
	\begin{tabular}{l|c|c|c|c}
		\hline \Tstrut
		Method &  Dataset &  Regularization  & Mean $\varepsilon_{\mathrm{v}}$ & Std. of $\varepsilon_{\mathrm{v}}$ \Bstrut\\
		\hline \Tstrut
		&     High-fidelity & None & $1.5607\times10^{-1}$  & $9.1084\times10^{-2}$\\
		&     High-fidelity & Dropout ($p=0.60$) & $1.1597\times10^{-1}$  & $6.2743\times10^{-3}$\\
		Strategy I &     High-fidelity & Standard $\ell_1$ ($\lambda=10^{-2}$) & $4.6827\times10^{-2}$  & $3.5092\times10^{-2}$\\
		Strategy II &        High-fidelity & Weighted $\ell_1$ ($\lambda=10^{-2}$) & $1.4481\times10^{-1}$  & $8.3662\times10^{-2}$\\
		Strategy III &       Bi-fidelity & Standard $\ell_1$ ($\lambda=10^{-4}$) & $3.5773\times10^{-2}$  & $2.1596\times10^{-2}$\\
		Strategy IV &       Bi-fidelity & Weighted $\ell_1$ ($\lambda=10^{-4}$) & $3.7563\times10^{-2}$ & $1.7820\times10^{-2}$\Bstrut\\\hline
	\end{tabular}
\end{table}

\begin{figure}[!htb]
	\centering
	\begin{subfigure}[!htb]{0.48\textwidth}
		\centering
		\includegraphics[scale=0.25]{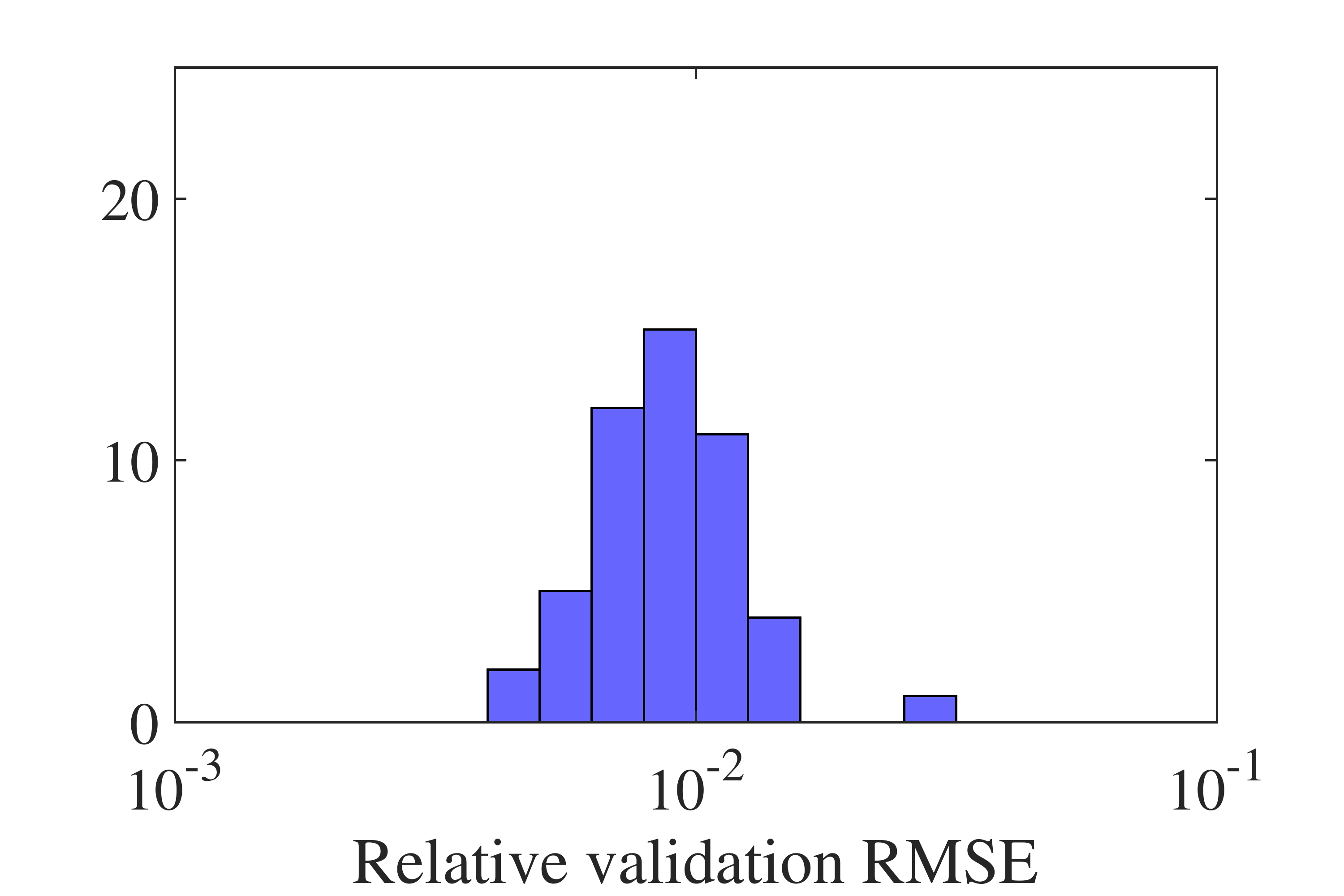}
		\caption{High-fidelity with no regularization}
	\end{subfigure}
	\hfill
	\begin{subfigure}[!htb]{0.48\textwidth}
		\centering
		\includegraphics[scale=0.25]{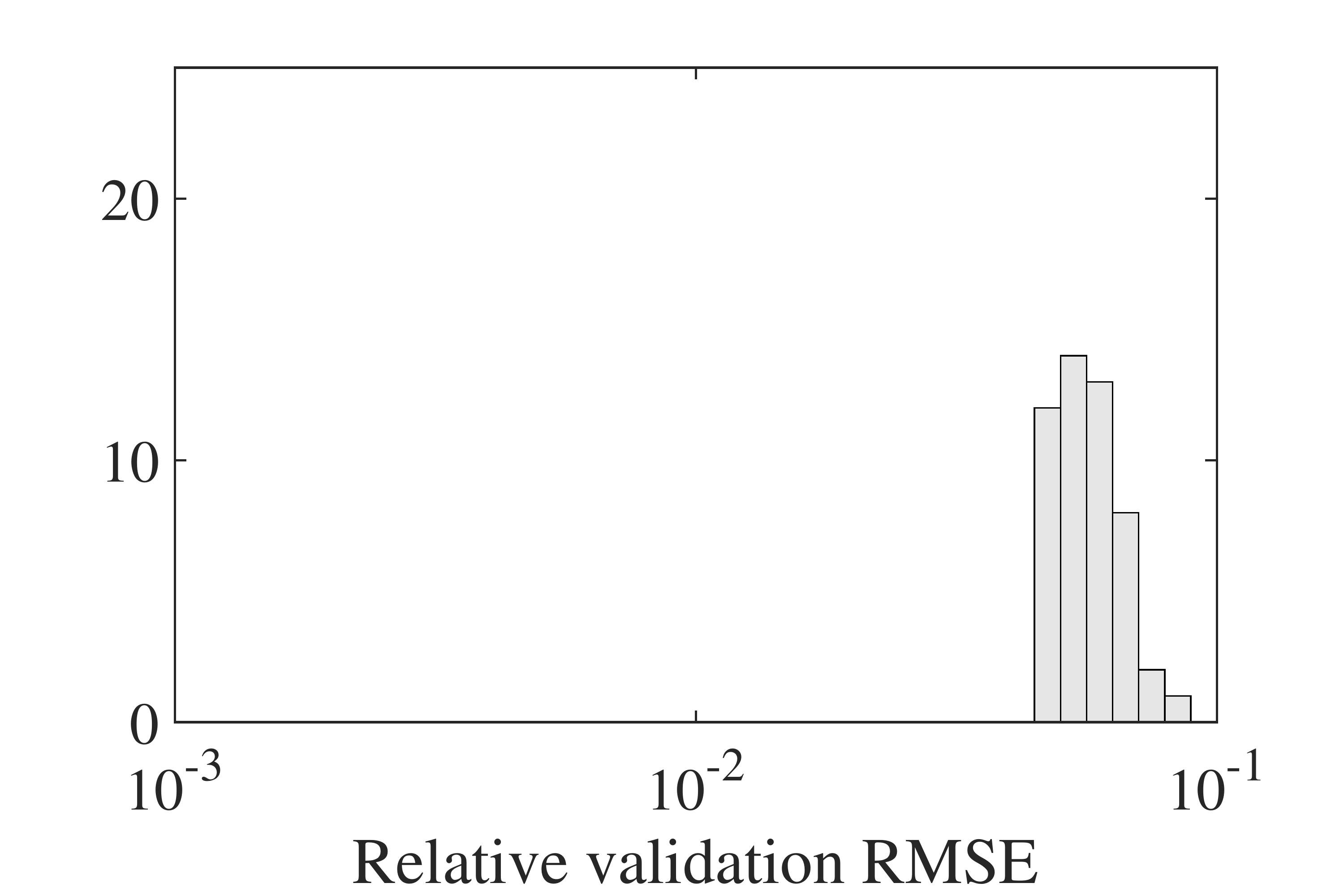}
		\caption{High-fidelity with dropout ($p=0.6$)}
	\end{subfigure}\\
	\begin{subfigure}[!htb]{0.48\textwidth}
		\centering
		\includegraphics[scale=0.25]{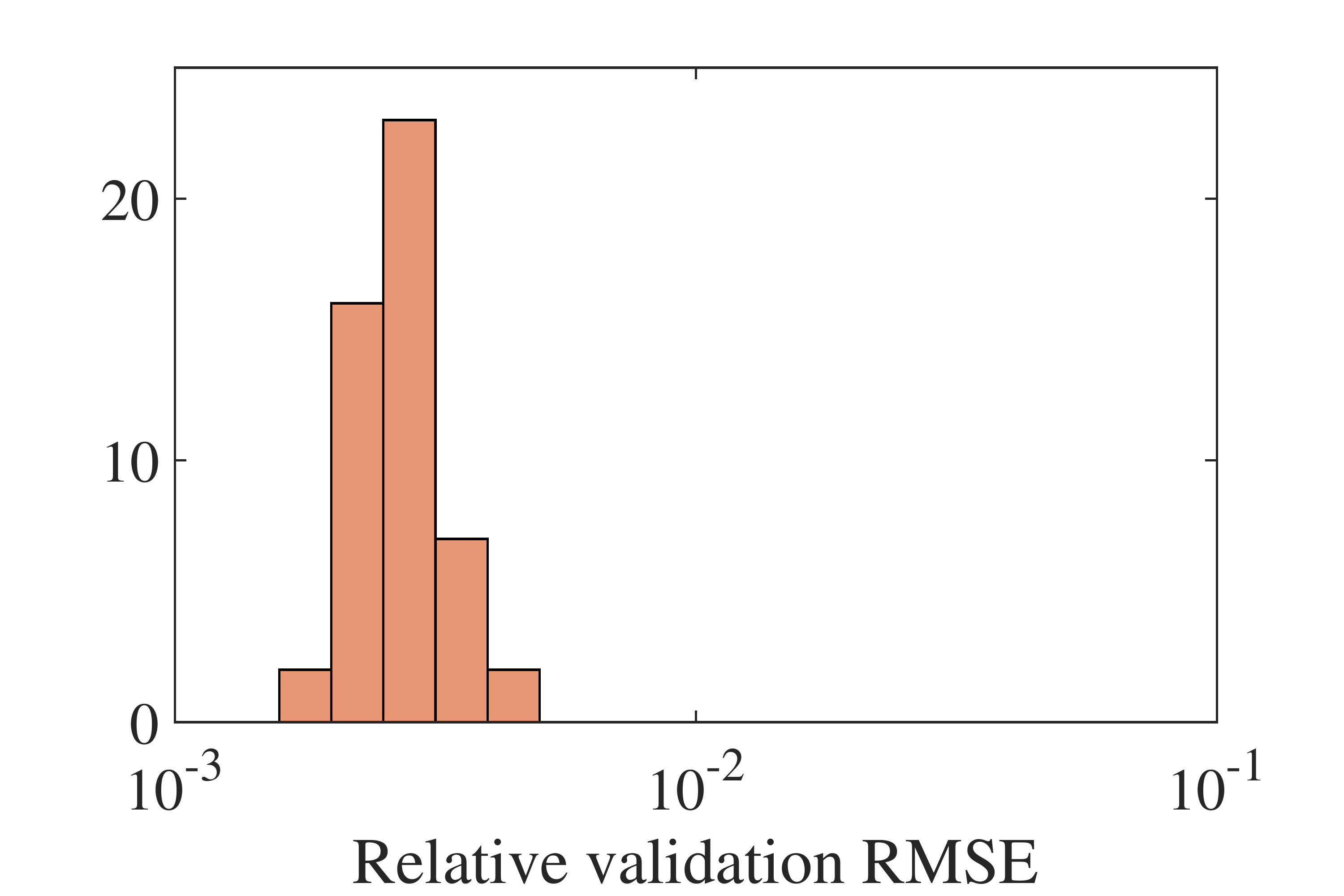}
		\caption{Strategy I: High-fidelity standard $\ell_1$-regularization}
	\end{subfigure}
	\hfill
	\begin{subfigure}[!htb]{0.48\textwidth}
		\centering
		\includegraphics[scale=0.25]{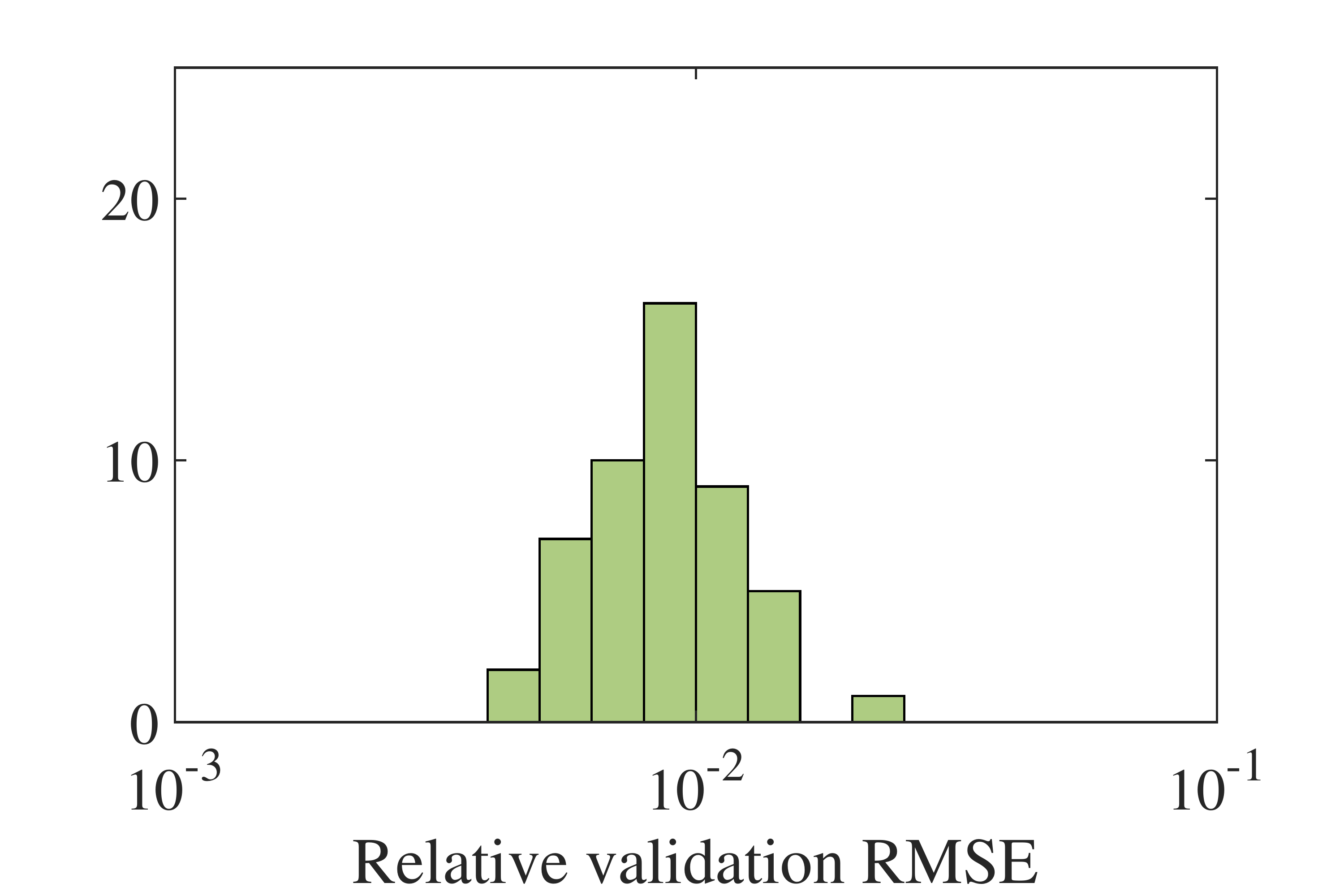}
		\caption{Strategy II: High-fidelity iteratively reweighted $\ell_1$-regularization}
	\end{subfigure}\\
	\begin{subfigure}[!htb]{0.48\textwidth}
		\centering
		\includegraphics[scale=0.25]{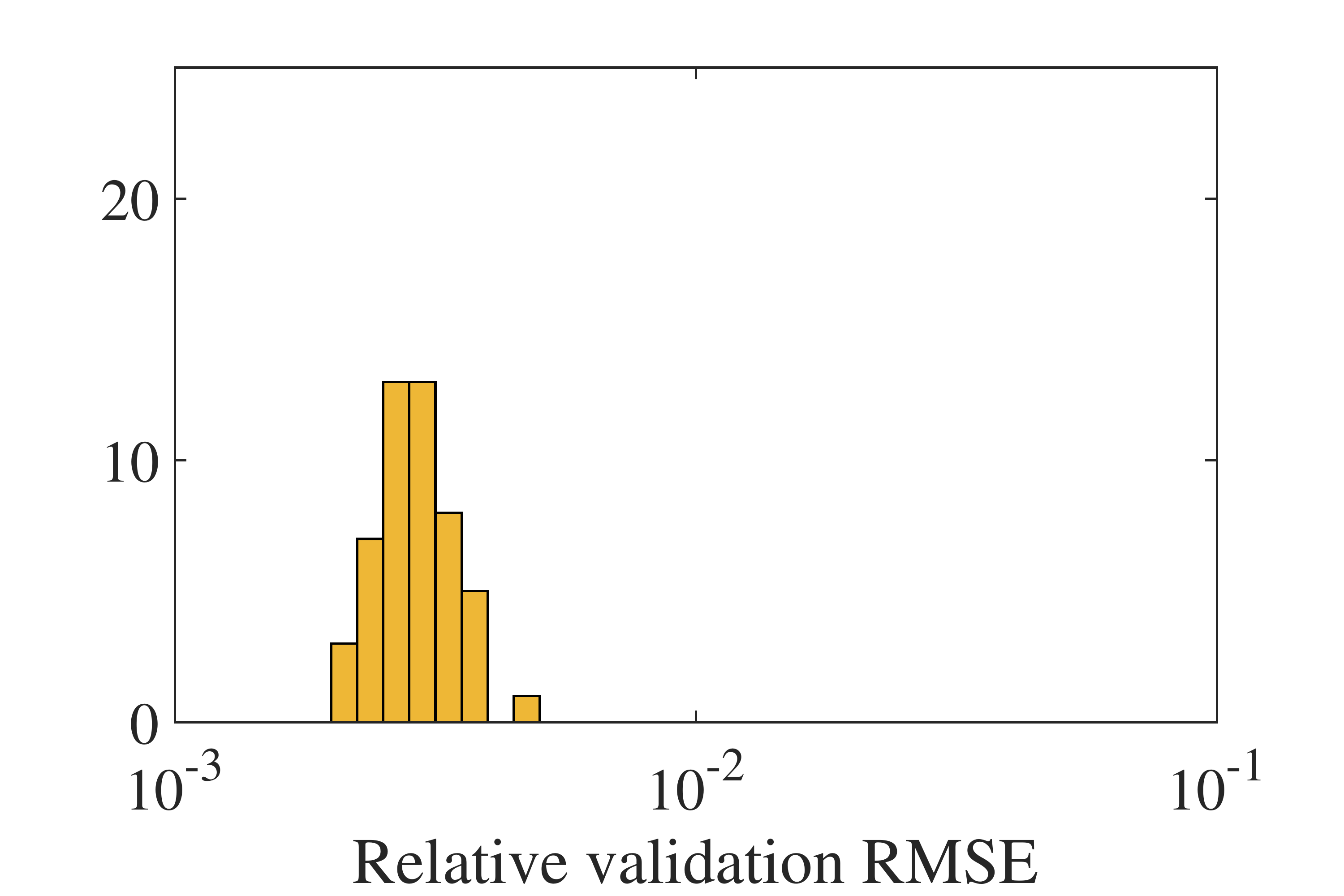}
		\caption{Strategy III: Bi-fidelity standard $\ell_1$-regularization}
	\end{subfigure}%
	\hfill
	\begin{subfigure}[!htb]{0.48\textwidth}
		\centering
		\includegraphics[scale=0.25]{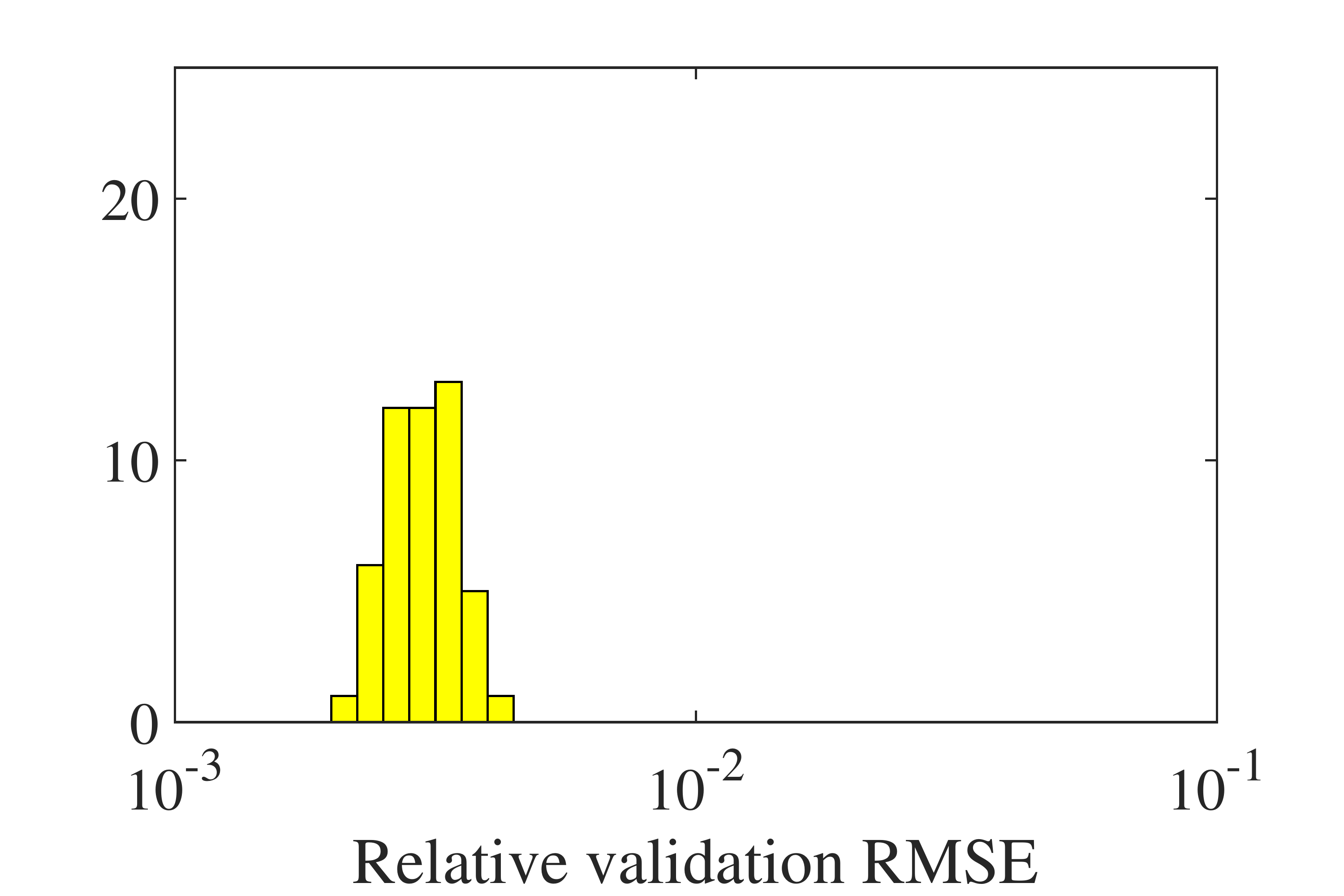}
		\caption{Strategy IV: Bi-fidelity weighted $\ell_1$-regularization}
	\end{subfigure}
	\caption{Histograms of relative validation RMSE, $\varepsilon_{\mathrm{v}}$, for 50 different replications of training/validation datasets for no regularization and four different $\ell_1$-regularization strategies in Example I. The high-fidelity training dataset  $\D_{\mathrm{tr},h}$ and the low-fidelity training dataset  $\D_{\mathrm{tr},l}$ consist of 10 and 250 realizations of the uncertain parameters, respectively. } \label{fig:beam_hist_Nh10} 
\end{figure}

\begin{figure}[!htb]
	\centering
	\begin{subfigure}[!htb]{0.8\textwidth}
		\centering
		\includegraphics[scale=0.25]{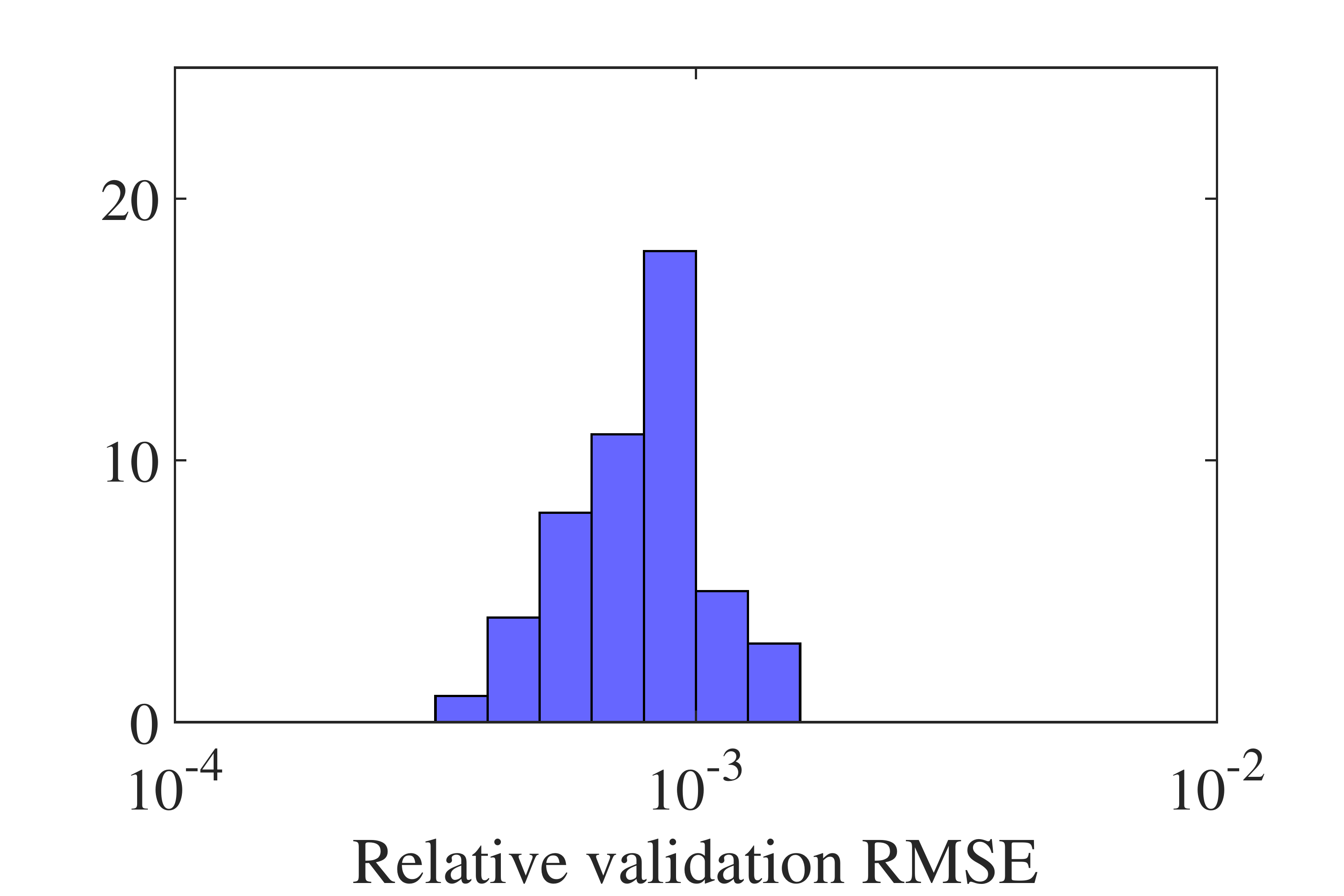}
		\caption{High-fidelity with no regularization}
	\end{subfigure}\\
	\begin{subfigure}[!htb]{0.48\textwidth}
		\centering
		\includegraphics[scale=0.25]{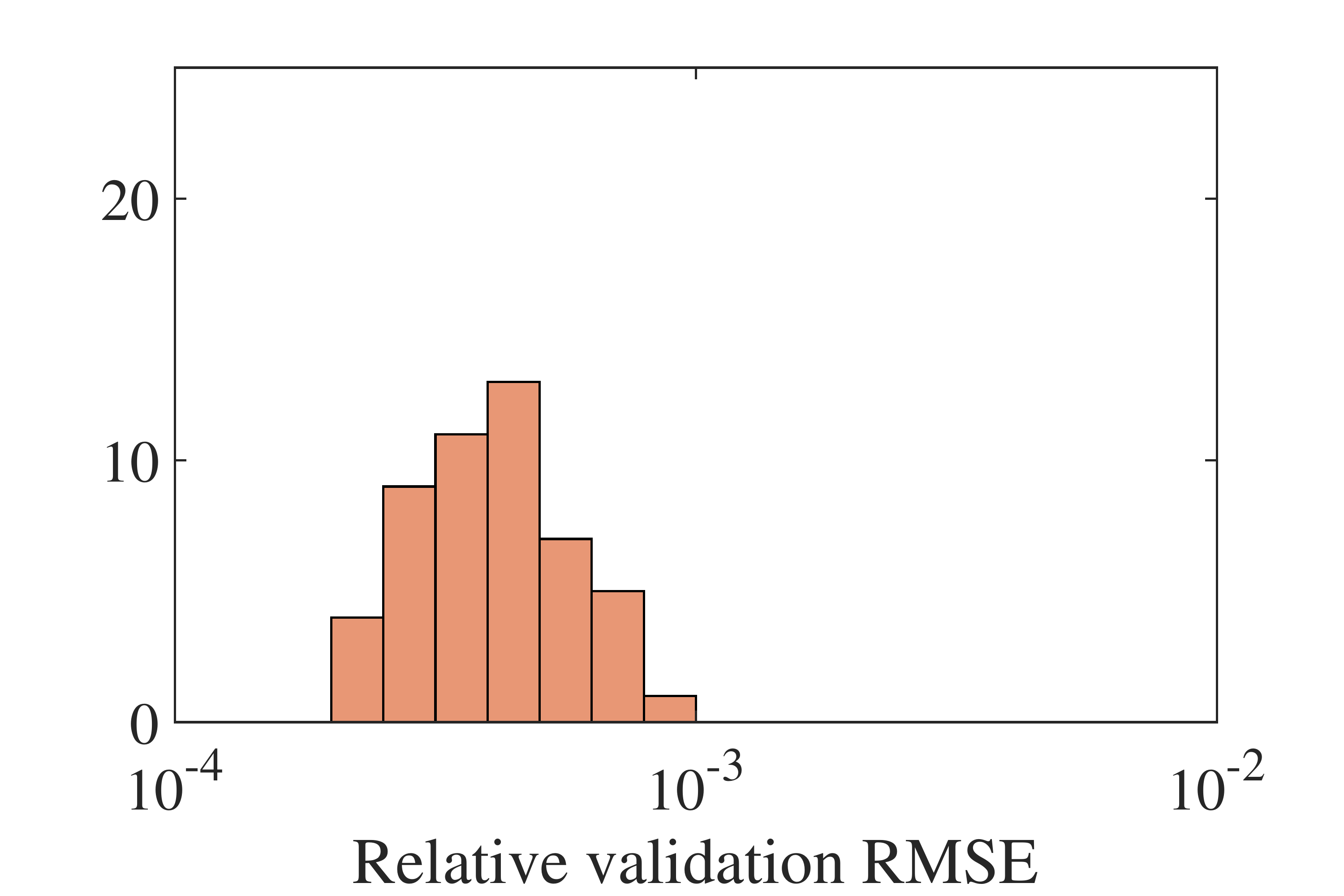}
		\caption{Strategy I: High-fidelity standard $\ell_1$-regularization}
	\end{subfigure}
	\hfill
	\begin{subfigure}[!htb]{0.48\textwidth}
		\centering
		\includegraphics[scale=0.25]{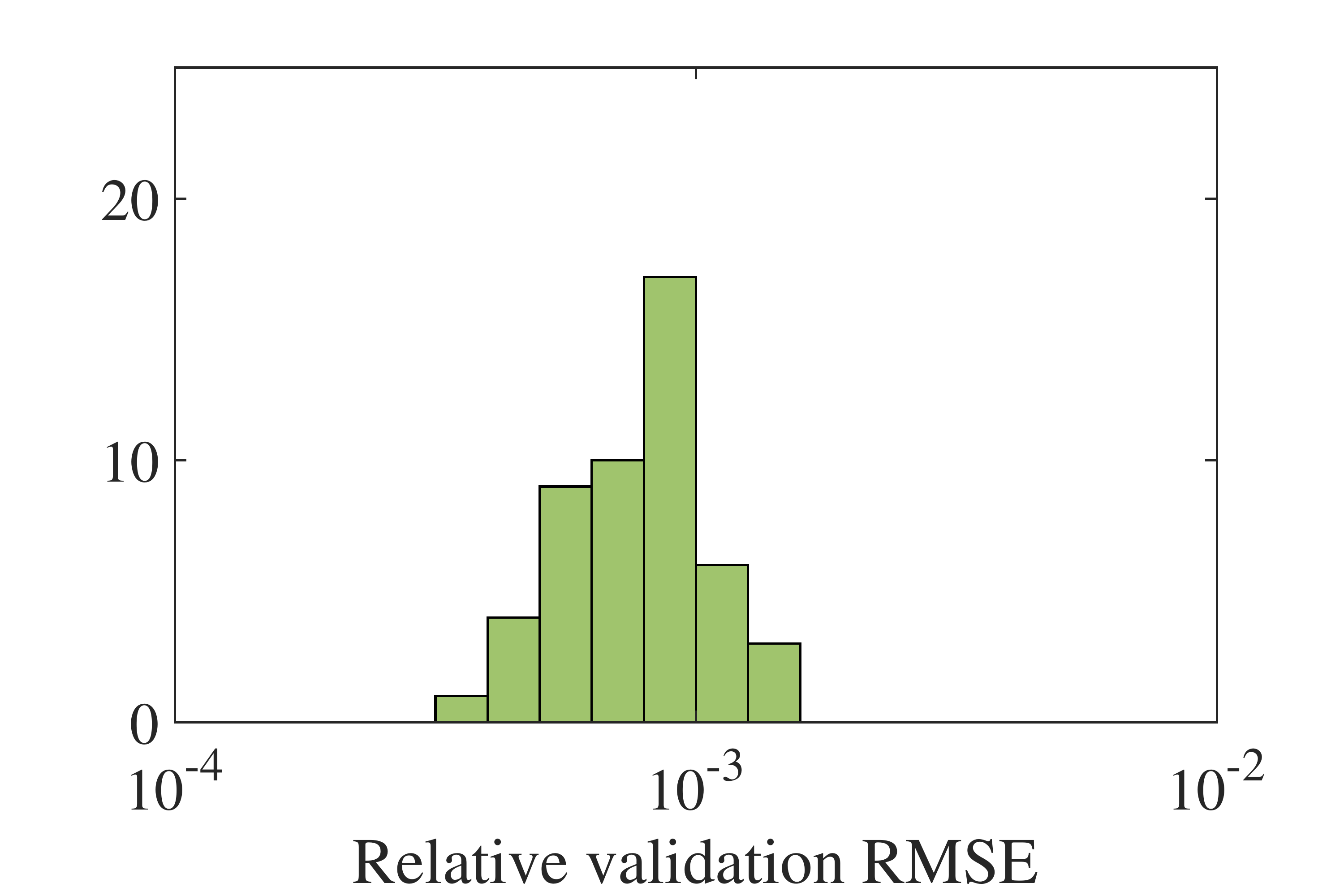}
		\caption{Strategy II: High-fidelity iteratively reweighted $\ell_1$-regularization}
	\end{subfigure}\\
	\begin{subfigure}[!htb]{0.48\textwidth}
		\centering
		\includegraphics[scale=0.25]{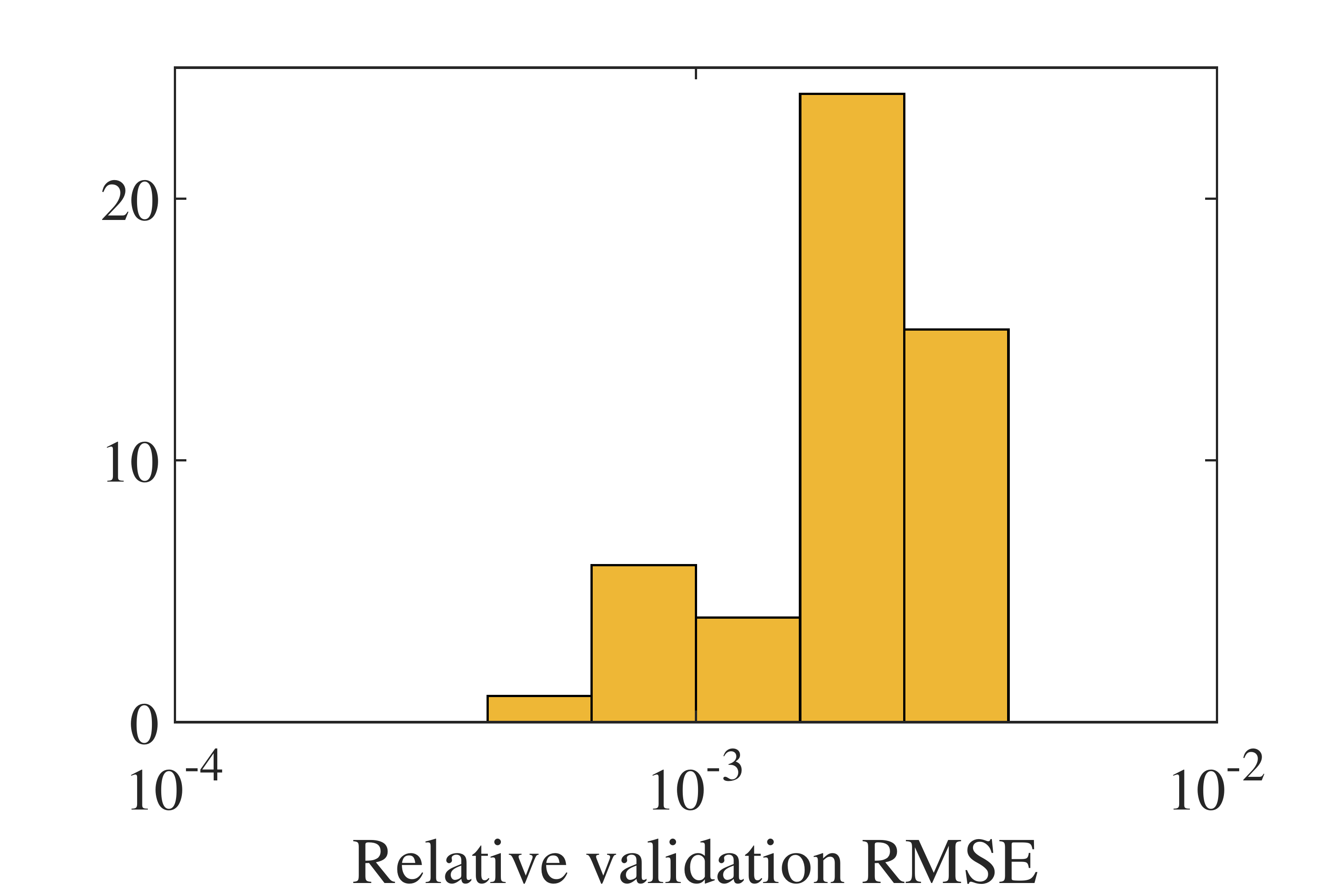}
		\caption{Strategy III: Bi-fidelity standard $\ell_1$-regularization}
	\end{subfigure}%
	\hfill
	\begin{subfigure}[!htb]{0.48\textwidth}
		\centering
		\includegraphics[scale=0.25]{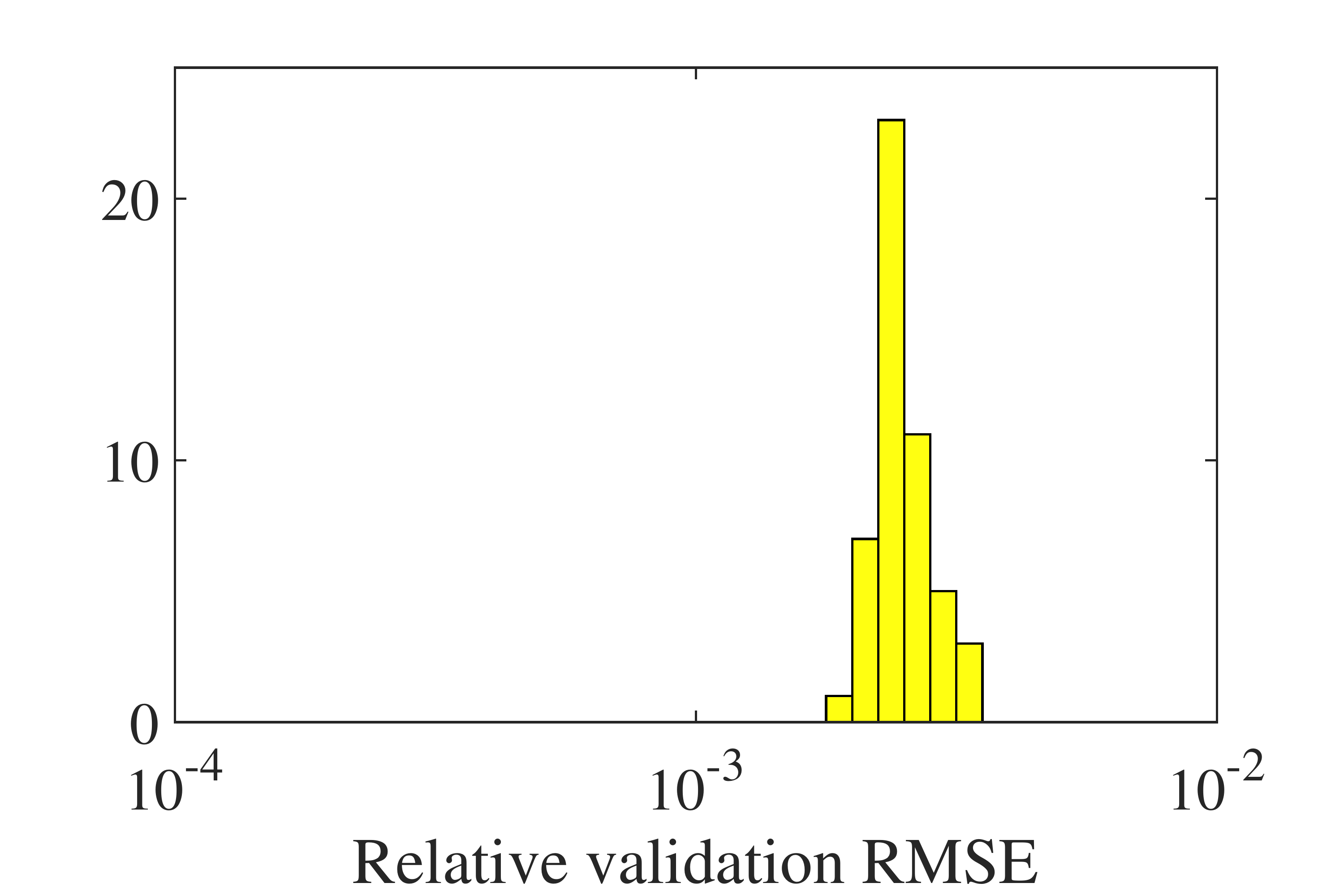}
		\caption{Strategy IV: Bi-fidelity weighted $\ell_1$-regularization}
	\end{subfigure}
	\caption{Histograms of relative validation RMSE, $\varepsilon_{\mathrm{v}}$, for 50 different replications of training/validation datasets for no regularization and four different $\ell_1$-regularization strategies in Example I. The high-fidelity training dataset  $\D_{\mathrm{tr},h}$ and the low-fidelity training dataset  $\D_{\mathrm{tr},l}$ consist of 50 and 250 realizations of the uncertain parameters, respectively. } \label{fig:beam_hist_Nh50} 
\end{figure}

Next, we investigate the effect of increasing the size of $\D_{\mathrm{tr},h}$ to 10 and 50 realizations of the uncertain parameters while we keep the the size of $\D_{\mathrm{tr},l}$ same as before. Figure \ref{fig:beam_hist_Nh10} shows that the high-fidelity standard $\ell_1$-regularization (Strategy I) provides similar $\varepsilon_{\mathrm{v}}$ to the two bi-fidelity $\ell_1$-regularization (strategies III and IV), when we increase the number of data points in $\D_{\mathrm{tr},h}$ to 10. However, these strategies still provide better $\varepsilon_{\mathrm{v}}$ compared to no regularization or Strategy II. If the the number of data points in $\D_{\mathrm{tr},h}$ is increased to 50, the performance of all high-fidelity methods with or without $\ell_1$-regularization improves compared to the two bi-fidelity $\ell_1$-regularization (strategies III and IV) as shown in Figure \ref{fig:beam_hist_Nh50}. This behavior is expected for any bi-fidelity method. In general, a crossover point exists for bi-fidelity methods beyond which these methods cannot show improvement over high-fidelity only methods. This phenomenon is also related to the crossover point in transfer learning of neural networks as commented in Figure 1 of De et al. \cite{de2020transfer}. However, for problems with high-dimensional inputs and/or when $\mathcal{M}$ is highly non-linear, it is likely that the crossover point is encountered at large high-fidelity sample sizes, making the bi-fidelity methods computationally more advantages. 

Dropout improves generalization by dropping some of the connections randomly with probability $p$ and then training the resulting subnetworks \cite{srivastava2014dropout}. During prediction a model averaging over all these subnetworks is performed. 
Figure \ref{fig:beam_hist} shows that the trained neural network with a dropout probability of 0.60 for each of the hidden layers leads to a smaller variance of $\varepsilon_{\mathrm{v}}$ without reducing the mean RMSE by a large factor.
Hence, the figure depicts a bias-variance trade-off in the generalization error for the dropout case \cite{geman1992neural,belkin2019reconciling}. In particular, the  dropout layers in the network introduce bias in the validation error while removing the variation over different validation datasets. This is further evident when the number of data points $N_h$ is increased to 10 as the variance of $\varepsilon_{\mathrm{v}}$ is reduced for $p=0.6$, while the corresponding mean of $\varepsilon_{\mathrm{v}}$ is increased relative to the case with no regularization. In fact, for $N_h=10$ the smallest mean $\varepsilon_{\mathrm{v}}$ is obtained for $p=0$, i.e., no dropout. Further theoretical and numerical investigation of the dropout approach to improve generalization error of neural networks is beyond the scope of the current study.

\begin{landscape}
	
	\begin{figure}[!htb]
		\centering
		\includegraphics[scale=0.5]{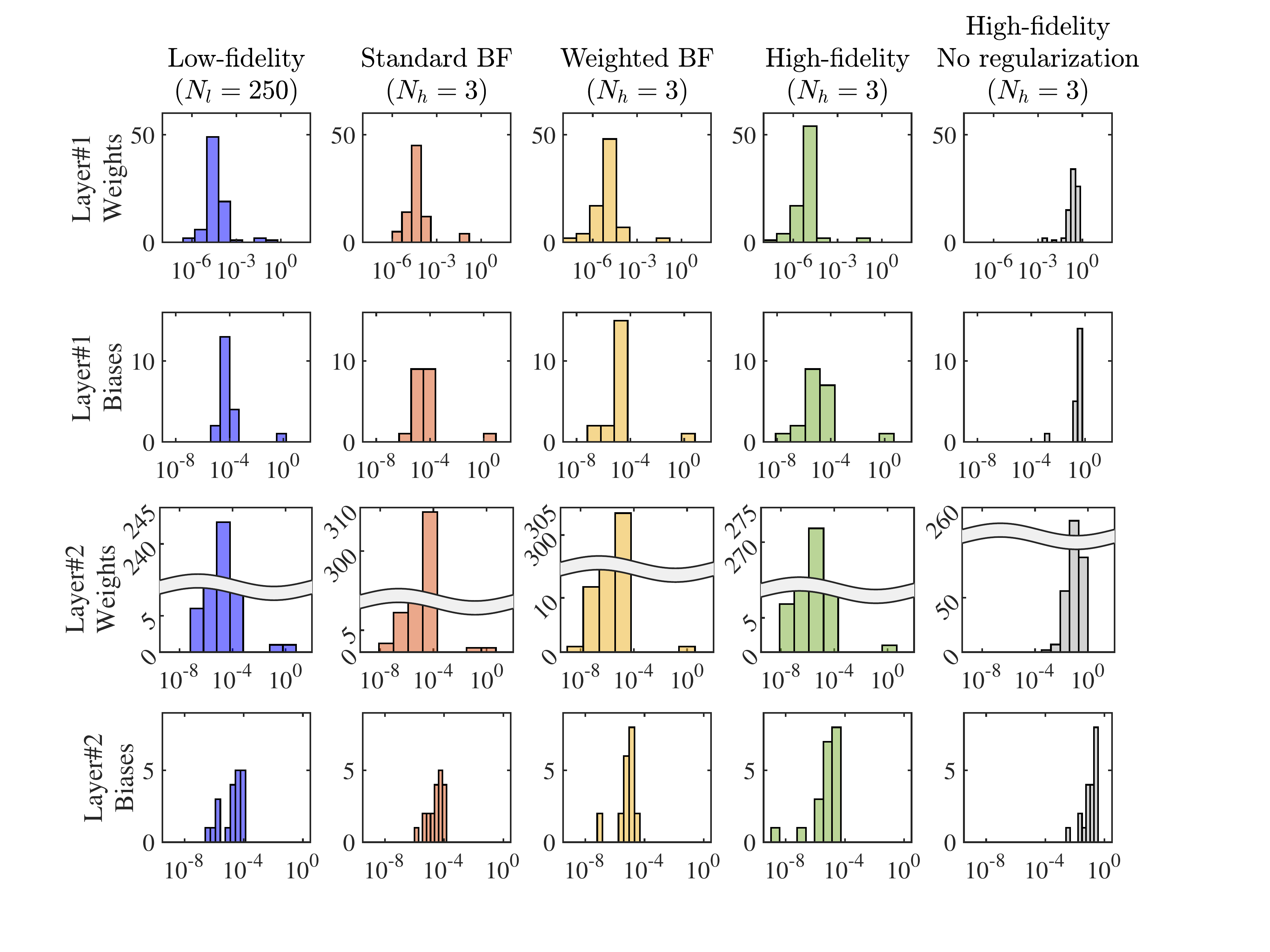}
		\caption{Histograms of the absolute parameter values of the hidden layers of the networks in Example I that are trained using different $\ell_1$-regularization strategies. The low-fidelity network is trained with one replication of the low-fidelity dataset $\D_{\mathrm{tr},l}$ with $N_l=250$. The high-fidelity network is trained for one replication of the high-fidelity dataset $\D_{\mathrm{tr},h}$ with $N_h=3$ and with or without any regularization. However, the bi-fidelity (BF) networks are trained using the parameters of the low-fidelity network and one replication of the high-fidelity dataset $\D_{\mathrm{tr},h}$ with $N_h=3$. 
		} \label{fig:beam_param_hist} 
	\end{figure} 
\end{landscape}

\begin{figure}
	\centering
	\begin{subfigure}[!htb]{0.48\textwidth}
		\centering
		\includegraphics[scale=0.35]{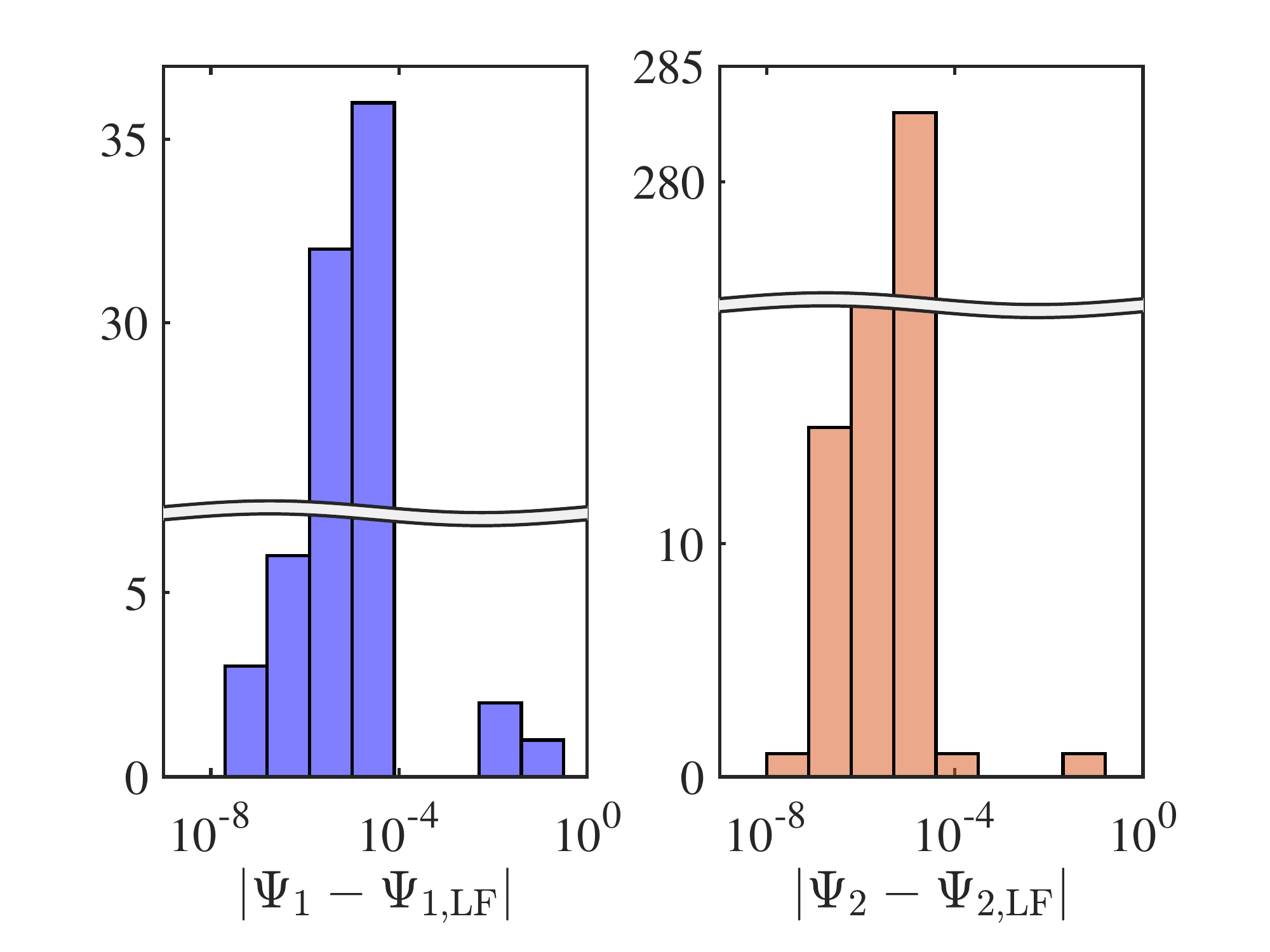}
		\caption{Weights of the two hidden layers}
	\end{subfigure}
	\begin{subfigure}[!htb]{0.48\textwidth}
		\centering
		\includegraphics[scale=0.35]{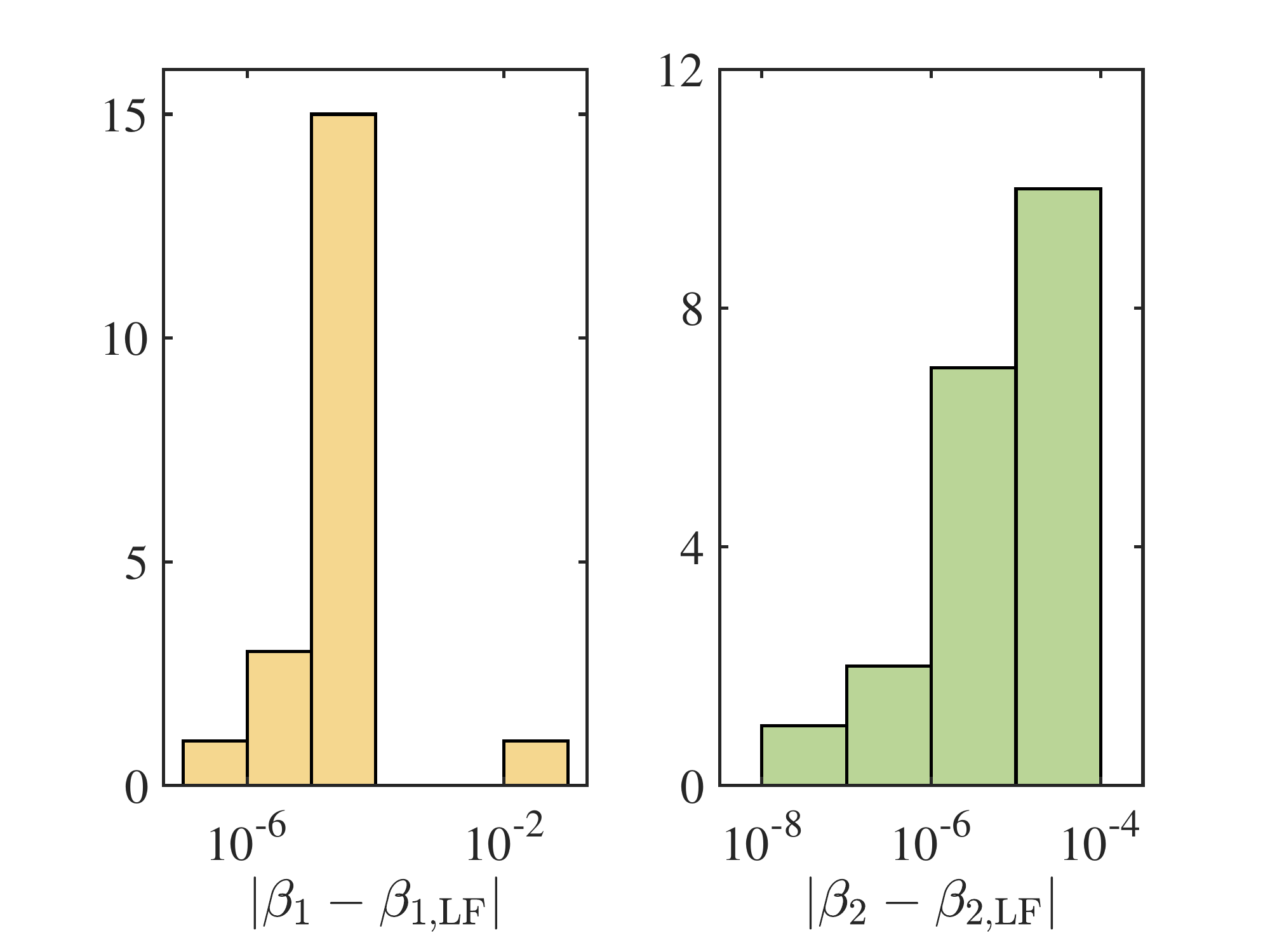}
		\caption{Biases of the two hidden layers}
	\end{subfigure}
	\caption{Histograms of the absolute difference of parameter values of the hidden layers of a network trained using the bi-fidelity Strategy III and a network trained using the low-fidelity dataset $\Dl$ in Example I. }
	\label{fig:beam_param_diff}
\end{figure}
Figure \ref{fig:beam_param_hist} compares histograms of the absolute parameter values of four networks for one replication of the datasets $\D_{\mathrm{tr},l}$ with $N_l=250$ and $\D_{\mathrm{tr},h}$ with $N_h=3$ to show the effects of the $\ell_1$-regularization strategies of this work. 
The bi-fidelity training strategies are implemented using this low-fidelity network and a high-fidelity dataset $\D_{\mathrm{tr},h}$ with $N_h=3$ data points. We compare the parameters of these networks with a high-fidelity network trained using a high-fidelity dataset $\D_{\mathrm{tr},h}$ that consists of $N_h=3$ data points and using either a standard $\ell_1$-regularization (Strategy I) with $\lambda=0.01$ or no regularization. 
The figure shows that the weights and biases of the network trained using the standard bi-fidelity Strategy III are close to the low-fidelity network parameters. However, the network trained using the weighted bi-fidelity Strategy IV gives parameters that have a similar histogram to that of the high-fidelity trained network. The figure also demonstrates that the networks trained using different $\ell_1$-regularization strategies have only a few parameters with large magnitude compared to the no regularization case while providing better accuracy. 
Figure \ref{fig:beam_param_diff} shows that only a small number of parameters of the two hidden layers of a network trained using the bi-fidelity Strategy III differ from the low-fidelity network parameters. Hence, Strategy III resembles the standard transfer learning closely but allows for a change in parameter values from all layers. On the other hand, Strategy IV provides more flexibility for changing parameters with large values. 
Since we allow changes in all parameters but according to their inferred values from the low-fidelity network, both bi-fidelity strategies can be regarded as generalizations of transfer learning using bi-fidelity datasets. Hence, in this example, the training of neural networks using strategies I-IV with a small number of high-fidelity data points results in sparsity in the parameters compared to the no regularization case. However, the networks trained using strategies III-IV have better predictive accuracy compared to networks trained using Strategy II. As such, we conjecture strategies III-IV are better at identifying important weights and biases.




\subsection{Example II: Thermally-driven Flow in a Cavity}

\begin{figure}[!htb]
	\centering
	\includegraphics[scale=1.0]{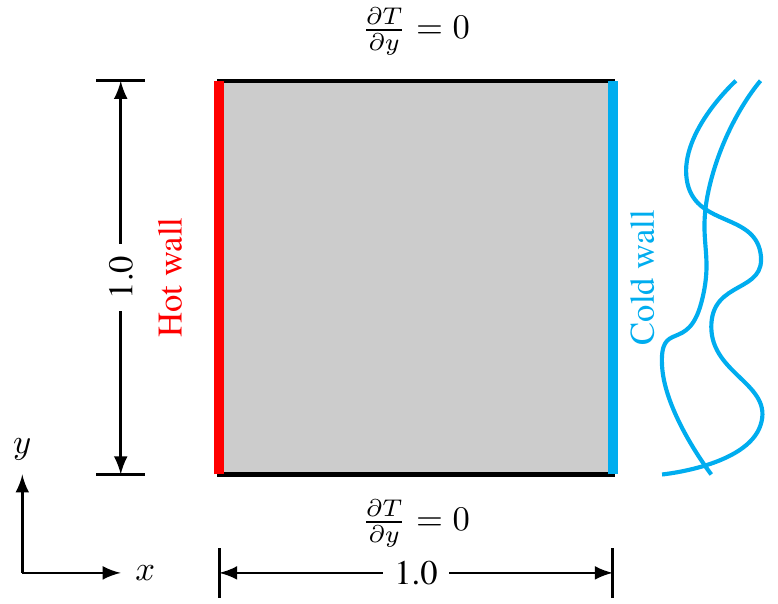}
	
	\caption{Thermally-driven flow in a cavity used in Example II.}
	\label{fig:ex2}
\end{figure}

In our second example, we consider a thermally-driven fluid flow in a cavity adapted from Le Ma{\^i}tre et al. \cite{le2002stochastic}, Peng et al. \cite{peng2014weighted}, and Hampton et al. \cite{hampton2018practical}. Figure \ref{fig:ex2} shows a schematic of the problem, where the left boundary is called the hot wall and has a random temperature $T_h$ with mean $\overline{T}_h$. The right boundary is kept cool using a flow of cold water and is referred to as the cold wall. The cold wall has a random temperature $T_c$ that is spatially varying with mean $\overline{T}_c$ and $T_c<T_h$. The top and bottom walls are thermally insulated. The QoI in this problem is the mean steady-state heat flux along the hot wall. The governing equations using small temperature difference approximation are given by
\begin{equation}\label{eq:cavity_flow}
\begin{split}
& \frac{\partial \um}{\partial t} + \um \cdot \nabla \um = - \nabla p + \frac{\mathrm{Pr}}{\sqrt{\mathrm{Ra}}} \nabla^2 \um + \mathrm{Pr} \phi \eem_y;\\
& \nabla \cdot \um = 0;\\
& \frac{\partial \phi}{\partial t} + \nabla \cdot (\um \phi) = \frac{1}{\sqrt{\mathrm{Ra}}} \nabla^2 \phi,
\end{split}
\end{equation}
where $\um$ is the velocity vector; $t$ is the time; $p$ is the pressure; the temperature gradient $\phi = \frac{T-\overline{T}_c}{T_h-\overline{T}_c}$; $\eem_y$ is the unit vector in $y$-direction. We follow the normalization used in Le Qu{\'e}r{\'e} \cite{le1991accurate} and Le Ma{\^i}tre et al. \cite{le2002stochastic} for \eqref{eq:cavity_flow}. The Prandtl and Rayleigh numbers are defined as $\mathrm{Pr}=\nu/\alpha=1/\sqrt{2}$ and $\mathrm{Ra}=g\tau (T_h-\overline{T}_c)L^3/(\nu \alpha)$, where $\nu$ is the kinematic viscosity; $\alpha$ is the thermal diffusivity; $L=1.0$ is the length of the cavity; $g=10.0$ is the gravitational acceleration; and $\tau=0.5$ is the coefficient of thermal expansion. Herein, we use finite volume discretization to solve \eqref{eq:cavity_flow}. 

For the cold wall, we use Karhunen-Lo\'{e}ve expansion to write the temperature
\begin{equation}
T_c(y) = \overline{T}_c + \sigma_T\sum_{i=1}^d \sqrt{\lambda_i}\varphi_i(y) \xi_i,
\end{equation}
where $\overline{T}_c=100$ is the mean temperature; 
$\{\lambda_i\}_{i=1}^d$ and $\{\varphi_i(y)\}_{i=1}^d$ are the $d=50$ largest eigenvalues and corresponding eigenfunctions, respectively, for an exponential correlation function $C(y_1,y_2) = \exp\left( -\frac{|y_1-y_2|}{l} \right)$ with the correlation length $l=0.15$; $\xi_i$ is an independent and identically distributed uniform random variable on $[-1, 1]$; and $\sigma_T=2$ is assumed. The temperature of the hot wall $T_h$ and the kinematic viscosity $\nu$ are assumed uniformly distributed between $[105,109]$ and $[0.004,0.01]$, respectively, resulting in dimension of the uncertainty as 52 in this problem. Note that we use consistent units for all the parameters here. For the low-fidelity model, we discretize the domain in $16\times16$ grid compared to $256\times 256$ grid for the high-fidelity model. 

\begin{table}[!htb]
	\caption{The mean and standard deviation of relative validation RMSE, $\varepsilon_{\mathrm{v}}$, using different $\ell_1$-regularization strategies in Example II.}
	\label{tab:heat_rmse}
	\centering
	\begin{tabular}{l|c|c|c|c}
		\hline \Tstrut
		Method &  Dataset & $\ell_1$-regularization  & Mean $\varepsilon_{\mathrm{v}}$ & Std.  of $\varepsilon_{\mathrm{v}}$ \Bstrut\\
		\hline \Tstrut
		&     High-fidelity & None & $4.3698\times10^{-1}$ & $5.0680\times10^{-2}$\\
		Strategy I &     High-fidelity & Standard ($\lambda=10^{-5}$) & $2.3461\times10^{-1}$ & $6.0780\times10^{-2}$\\
		Strategy II &        High-fidelity & Weighted ($\lambda=10^{-2}$) & $2.7139\times10^{-1}$ & $3.8954\times10^{-2}$\\
		Strategy III &       Bi-fidelity & Standard ($\lambda=10^{-5}$) & $4.9379\times10^{-2}$ & $7.6861\times10^{-3}$\\
		Strategy IV &       Bi-fidelity & Weighted ($\lambda=10^{-4}$) & $4.3573\times10^{-2}$ & $1.1052\times10^{-2}$\Bstrut\\\hline
	\end{tabular}
\end{table}

\subsubsection{Results} 

We use separate $N_h = 5$ and $N_l = 150$ realizations of the 52 uncertain parameters and their corresponding mean steady-state heat flux on the hot wall to construct the high-fidelity training dataset  $\D_{\mathrm{tr},h}$ and the low-fidelity training dataset $\D_{\mathrm{tr},l}$, respectively. For the validation dataset $\Dval$, we use 50 different realizations of the uncertain parameters. 
For the neural network configuration, we choose an FNN with four hidden layers consisting of 52, 60, 60, 25 neurons, respectively, and ELU activation function.  
We use the Adam algorithm (see Appendix \ref{sec:adam}) with a learning rate of $10^{-4}$ and 40,000 epochs. We select the network configuration that produces the smallest $\varepsilon_{\mathrm{v}}$ at the end of training as the trained network. 
The low-fidelity network $\pinn_l$ is trained using a standard $\ell_1$-regularization with $\lambda=0.001$. 

\begin{figure}[!htb]
	\centering
	\begin{subfigure}[!htb]{0.8\textwidth}
		\centering
		\includegraphics[scale=0.25]{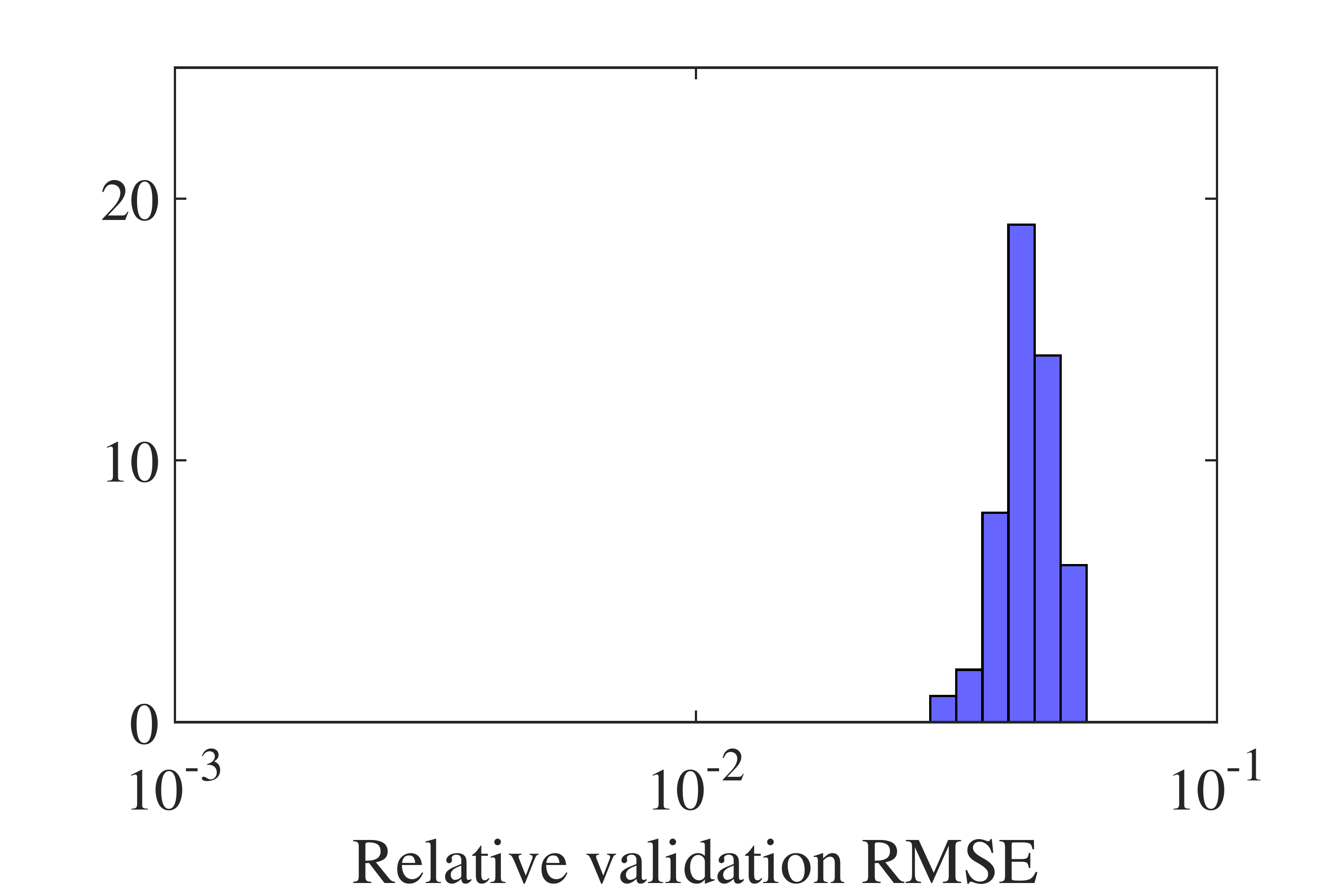}
		\caption{High-fidelity with no regularization}
	\end{subfigure}\\
	\begin{subfigure}[!htb]{0.48\textwidth}
		\centering
		\includegraphics[scale=0.25]{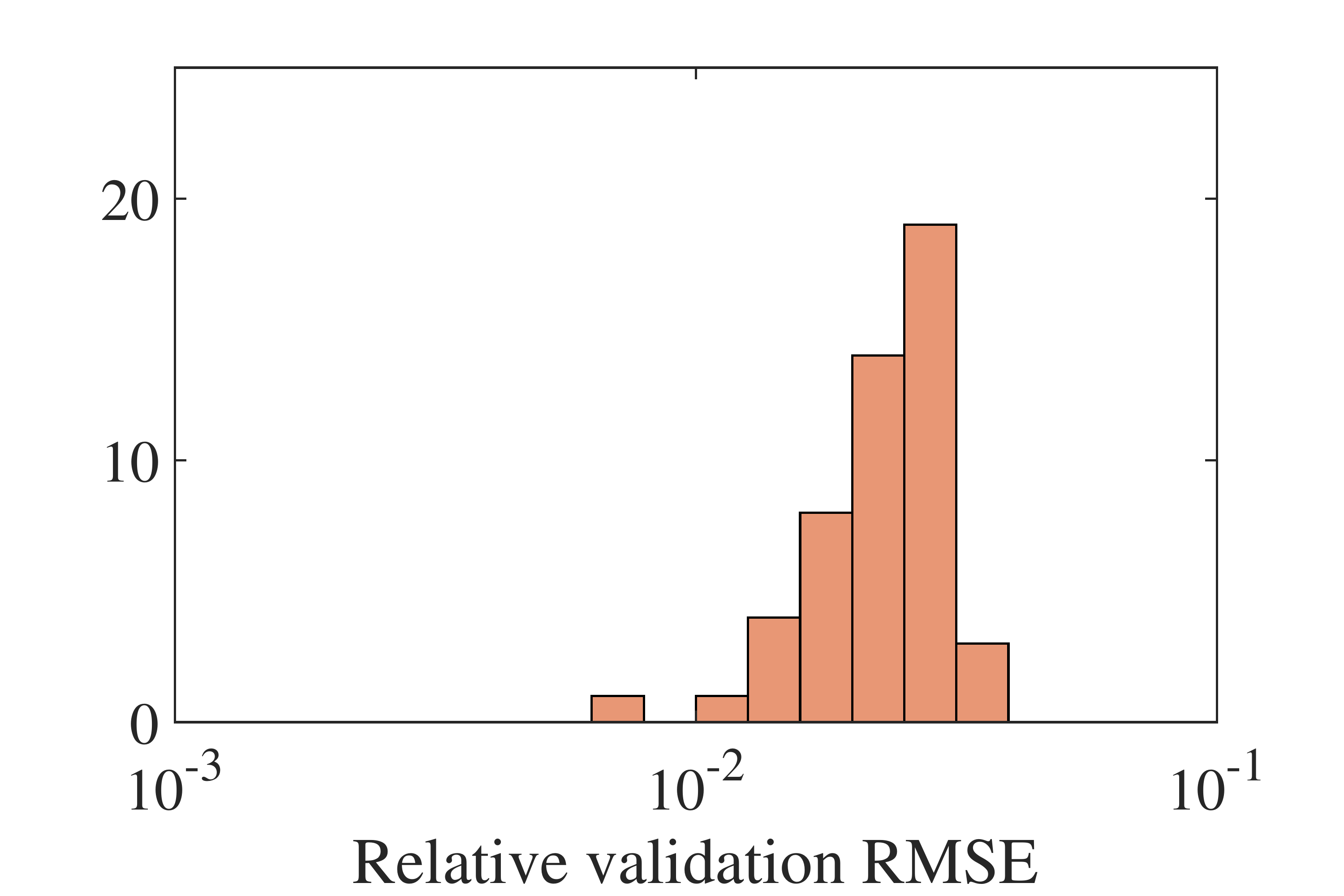}
		\caption{Strategy I: High-fidelity standard $\ell_1$-regularization}
	\end{subfigure}
	\hfill
	\begin{subfigure}[!htb]{0.48\textwidth}
		\centering
		\includegraphics[scale=0.25]{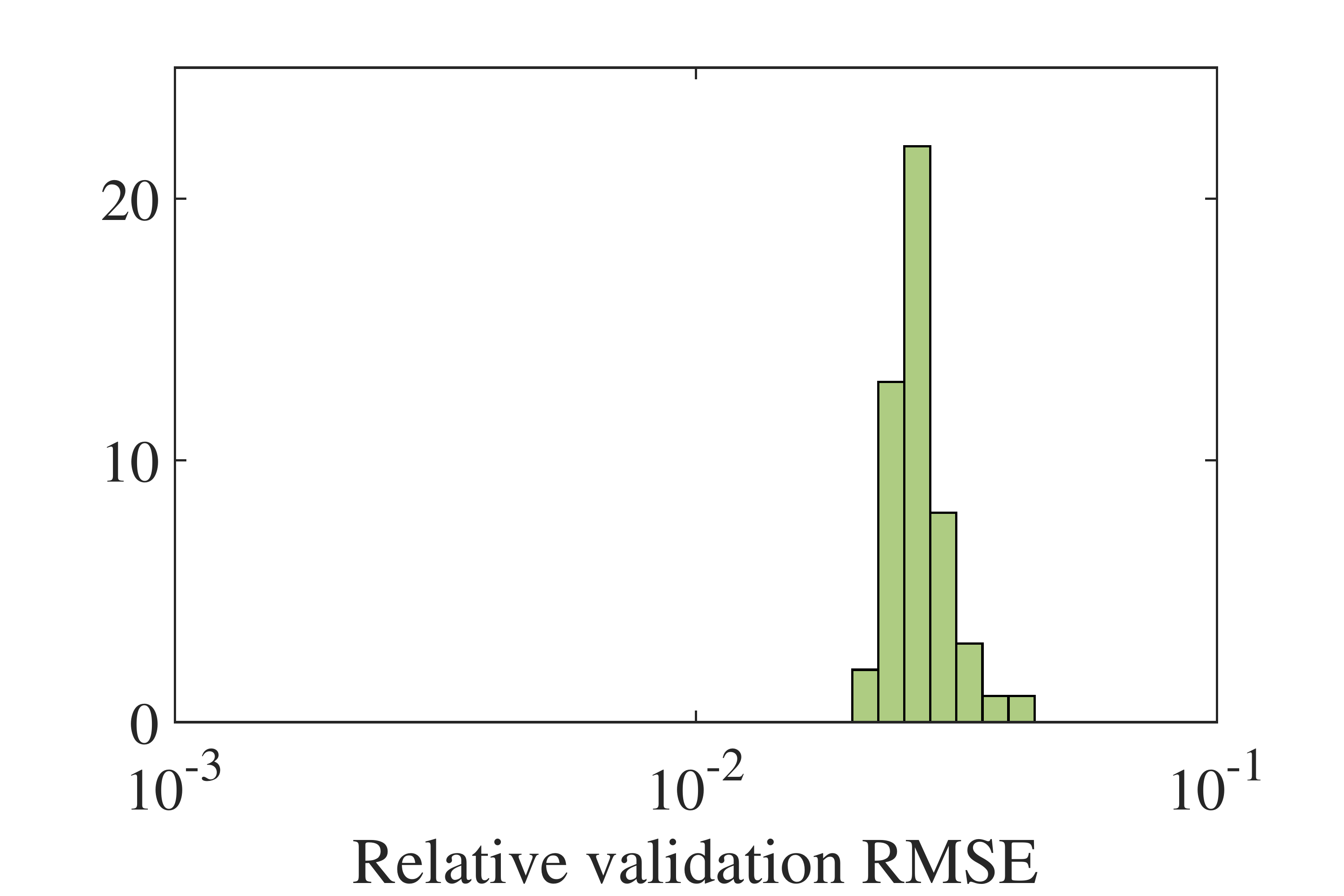}
		\caption{Strategy II: High-fidelity iteratively reweighted $\ell_1$-regularization}
	\end{subfigure}\\
	\begin{subfigure}[!htb]{0.48\textwidth}
		\centering
		\includegraphics[scale=0.25]{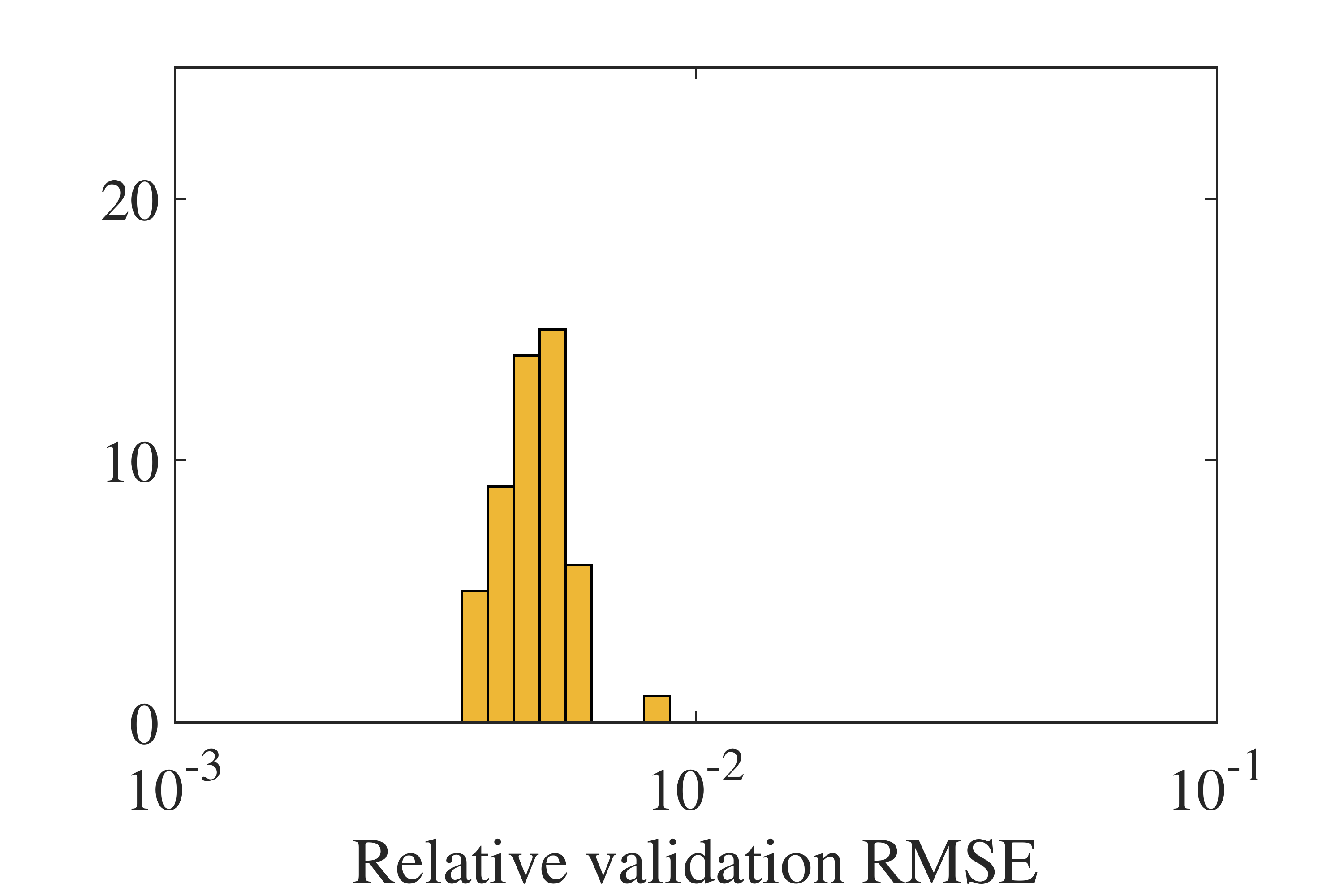}
		\caption{Strategy III: Bi-fidelity standard $\ell_1$-regularization}
	\end{subfigure}%
	\hfill
	\begin{subfigure}[!htb]{0.48\textwidth}
		\centering
		\includegraphics[scale=0.25]{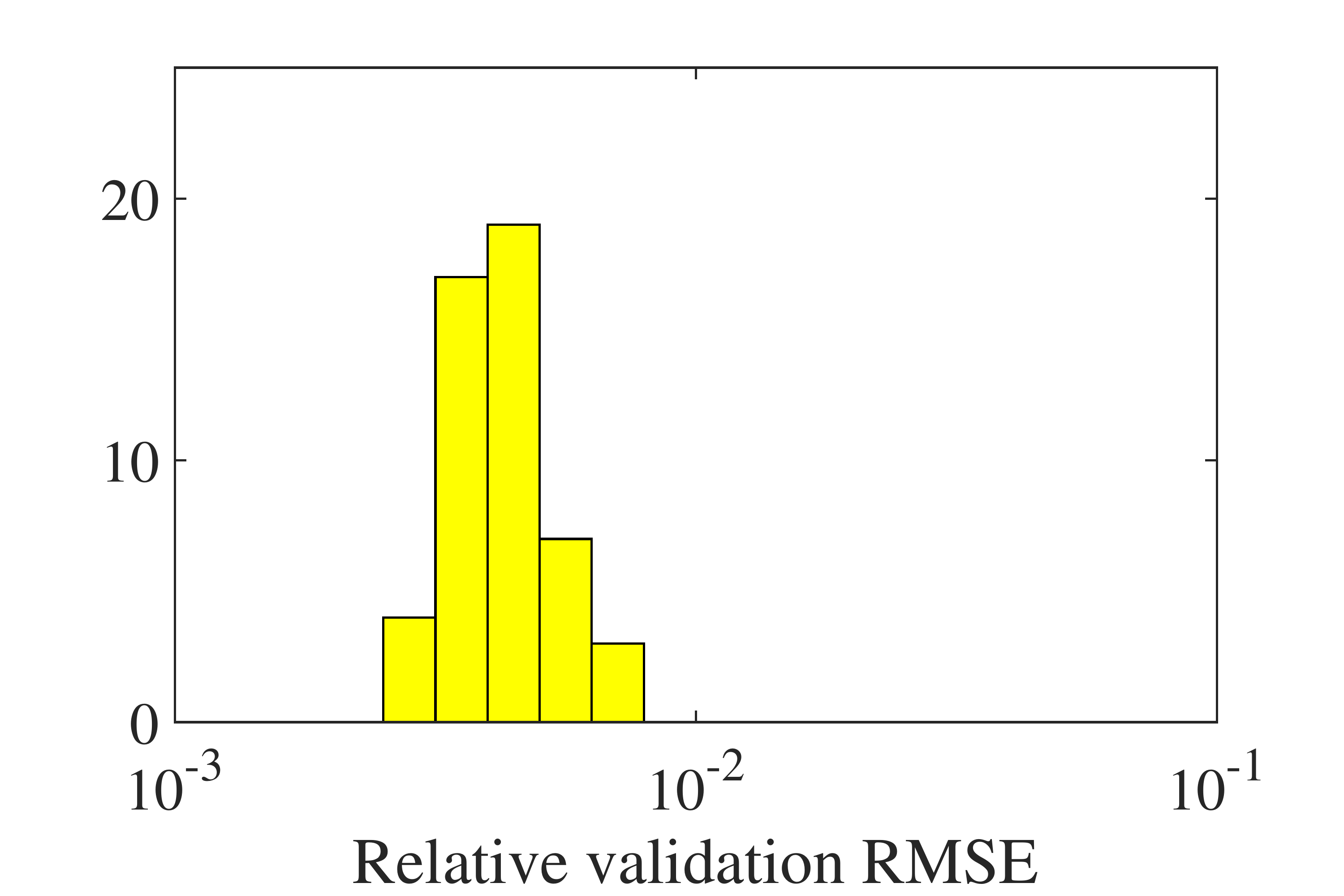}
		\caption{Strategy IV: Bi-fidelity weighted $\ell_1$-regularization}
	\end{subfigure}
	\caption{Histograms of relative validation RMSE, $\varepsilon_{\mathrm{v}}$, for 50 different replications of training/validation datasets for no regularization and four different $\ell_1$-regularization strategies in Example II.} \label{fig:heat_hist} 
\end{figure}

Figure \ref{fig:heat_hist} shows the histograms of $\varepsilon_{\mathrm{v}}$ resulting from different $\ell_1$-regularization strategies for 50 replications of the training and validation datasets. From this figure, we observe that with only the high-fidelity training data, the standard Strategy I and Strategy II reduce $\varepsilon_{\mathrm{v}}$ compared to the case in which no regularization is applied. However, when we use Strategy II, variations over different datasets are small compared to standard $\ell_1$-regularization. The bi-fidelity strategies III and IV, on the other hand, improve $\varepsilon_{\mathrm{v}}$ by one order of magnitude. 

Table \ref{tab:beam_rmse} reports the mean and standard deviation of $\varepsilon_{\mathrm{v}}$ for the four $\ell_1$-regularization strategies along with a case with no regularization. The two bi-fidelity $\ell_1$-regularization strategies III and IV improve the mean $\varepsilon_{\mathrm{v}}$ by one order of magnitude compared to all other cases as can be noticed from the table. The standard deviation of $\varepsilon_{\mathrm{v}}$ is also improved by using strategies II, III, and IV.
Similar to the previous example, this example again shows that if we have a large dataset from a low-fidelity model compared to a small dataset from the high-fidelity model of the physical system then it is useful and advantageous to use the bi-fidelity strategies III and IV for training the neural networks. 

\begin{landscape}
	\begin{figure}[!htb]
		\centering
		\includegraphics[scale=0.5]{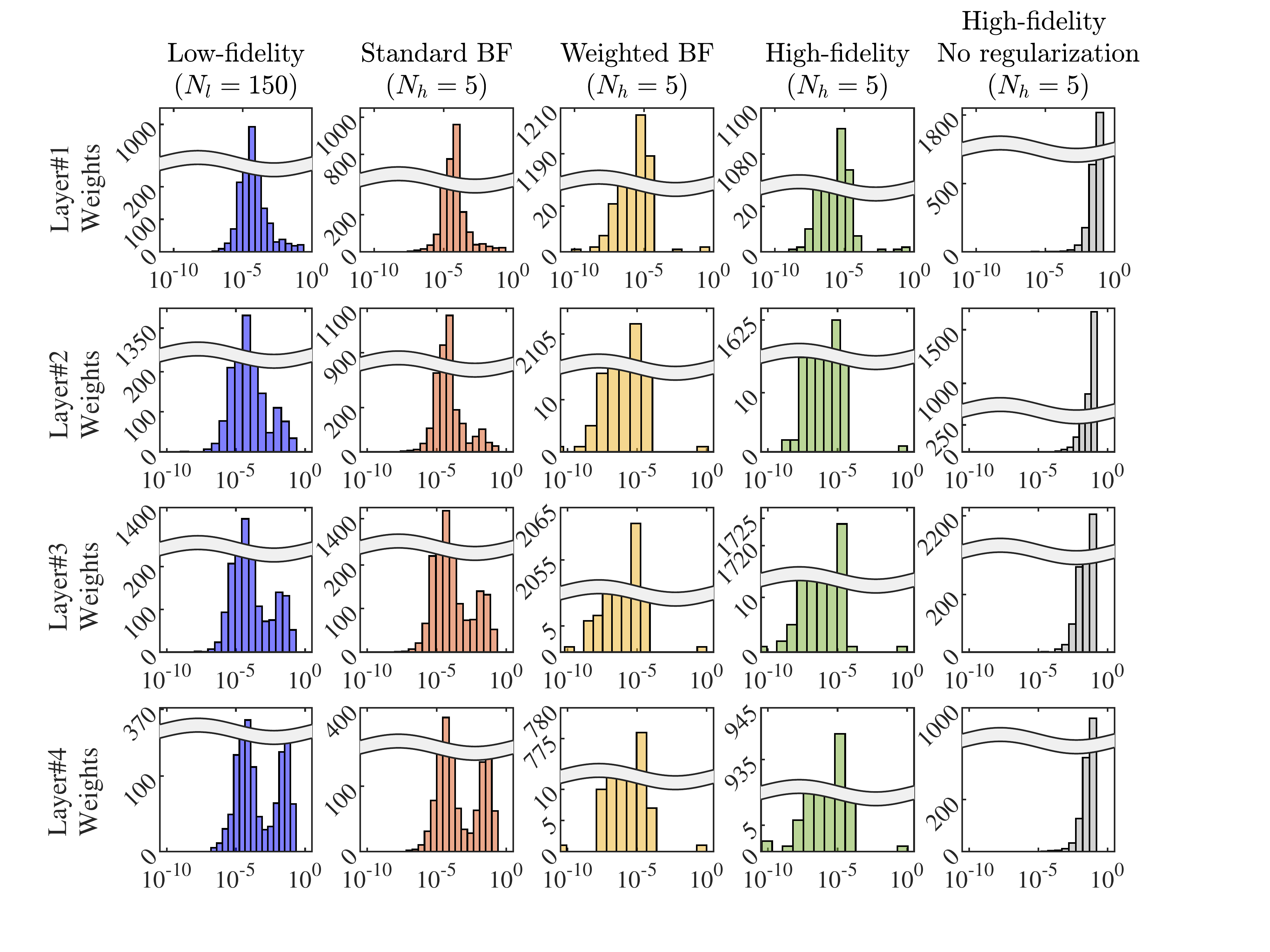}
		\caption{Histograms of weights of the hidden layers of the networks in Example II that are trained using different $\ell_1$-regularization strategies. The low-fidelity network is trained with one replication of the low-fidelity dataset $\D_{\mathrm{tr},l}$ with $N_l=150$. The high-fidelity network is trained using $\D_{\mathrm{tr},h}$ with $N_h=5$, and with or without any regularization. However, the bi-fidelity (BF) networks are trained using the parameters of the low-fidelity network and one replication of the high-fidelity dataset $\D_{\mathrm{tr},h}$ with $N_h=5$. 
		} \label{fig:heat_param_hist} 
	\end{figure} 
\end{landscape}

Figure \ref{fig:heat_param_hist} compares magnitudes of the weights of the hidden layers to show the effects of the bi-fidelity training strategies. 
The histograms show that the standard bi-fidelity Strategy III with $N_h=5$ high-fidelity samples produces similar weights to the low-fidelity network trained using $\D_{\mathrm{tr},l}$ with $N_l=150$. However, the weighted bi-fidelity Strategy IV results in weights more similar to the high-fidelity network trained using Strategy I with $\lambda=10^{-5}$ and $N_h=5$.  The figure also illustrates that the regularized networks have only a few parameters with large magnitude compared the no regularization case, but provide better accuracy. 
Similar to the previous example, these results show that Strategy III closely resembles the standard transfer learning, but Strategy IV provides sparser parameters. Also, strategies III-IV likely identify the important weights and biases better compared to strategies I-II in this example as they result in smaller prediction errors.



%
\subsection{Example III: Isentropic Flow through a Dual Throat Nozzle} 

In our third example, previously studied in Chen et al.~\cite{chen2005uncertainty}, we explore $\ell_1$-regularization of an autoencoder approximating the solution to the steady-state isentropic flow through a dual throat nozzle. Depending on the initial state, the flow results in a fully subsonic, supersonic, or a shock where the flow changes from subsonic to supersonic regime. The position of the shock also depends on the initial condition. In this example, we use a simplified model based on Burger's equation given by
\begin{equation} \label{eq:burger}
\frac{\partial u}{\partial t} + \frac{\partial}{\partial x}\left( \frac{u^2}{2} \right) = \frac{\partial}{\partial x}\left( \frac{\sin^2u}{2} \right); \quad 0\leq x \leq \pi; t>0,
\end{equation}
where the initial condition is $u(x,\delta,0)=\delta\sin x$ and the boundary conditions are $u(0,\delta,t) = u(\pi,\delta,t)=0$. 
The steady-state solution of \eqref{eq:burger} is given by \cite{chen2005uncertainty}
\begin{equation}
\begin{split}
u_{\infty}(x,\delta) = \lim_{t\rightarrow \infty} u(x,\delta,t) = \begin{cases}
u^+=\sin x, \quad\quad~ 0\leq x \leq X_s;\\
u^-=-\sin x, \quad X_s\leq x \leq \pi,\\
\end{cases}
\end{split}
\end{equation}
where the shock is at $x=X_s$, which depends on the parameter $\delta$ as follows
\begin{equation}
X_s = \begin{cases}
\sin^{-1}\left( \sqrt{1-\delta^2} \right) < \pi/2; \qquad -1<\delta\leq 0;\\
\pi - \sin^{-1}\left( \sqrt{1-\delta^2} \right) > \pi/2; \quad 0<\delta< 1.\\
\end{cases}
\end{equation}
For $|\delta|\geq 1$, the shock does not appear and $u$ is smooth \cite{chen2005uncertainty}.
Herein, the parameter $\delta$ is assumed a random variable that takes values in $[-1,1]$ as follows
\begin{equation}
\delta = 
\frac{-1+\sqrt{1+4\xi^2} }{2\xi},
\end{equation}
where $\xi$ is a Gaussian random variable with zero mean and unit variance. For the low-fidelity model, we discretize $x\in[0,\pi]$ to 52 grid points to estimate $u_{\infty}(x,\delta)$, whereas for the high-fidelity model, we discretize $x\in[0,\pi]$ to 1048 grid points to estimate $u_{\infty}(x,\delta)$. Note that due to a smaller number of grid points in the low-fidelity model there will be error in the shock position. 

\subsubsection{Results} 
In this example, we use a high-fidelity training dataset  $\D_{\mathrm{tr},h}$ that consists of $N_h = 50$ realizations of $(\xi,u)$.
The validation dataset $\Dval$ consists of 50 separate realizations of the uncertain parameter $\xi$ and the corresponding $u$. For the low-fidelity training dataset $\D_{\mathrm{tr},l}$, we use 400 realizations of $(\xi,u)$ though using a coarse discretization of $x$ as explained in the last subsection. The $\varepsilon_{\mathrm{v}}$ is measured using the shock position as QoI in this example. 
Here, we use an autoencoder that consists of an encoder with 128, 64, and 16 neurons in three hidden layers, and a decoder with 16, 64, and 128 neurons in three hidden layers. ELU activation (see \eqref{eq:elu}) is used in each of the hidden layers with tanh activation to the output. Hence, the dimension of the latent space/encoded data is 16. We use Adam with a learning rate of $10^{-4}$ to train the autoencoder. At the end of 5,000 iterations, we select the network configuration that produces the smallest $\varepsilon_{\mathrm{v}}$.  Note that the low-fidelity network $\pinn_l$ is trained using a standard $\ell_1$-regularization with $\lambda=10^{-8}$. 

\begin{figure}[!htb]
	\centering
	\begin{subfigure}[!htb]{0.8\textwidth}
		\centering
		\includegraphics[scale=0.25]{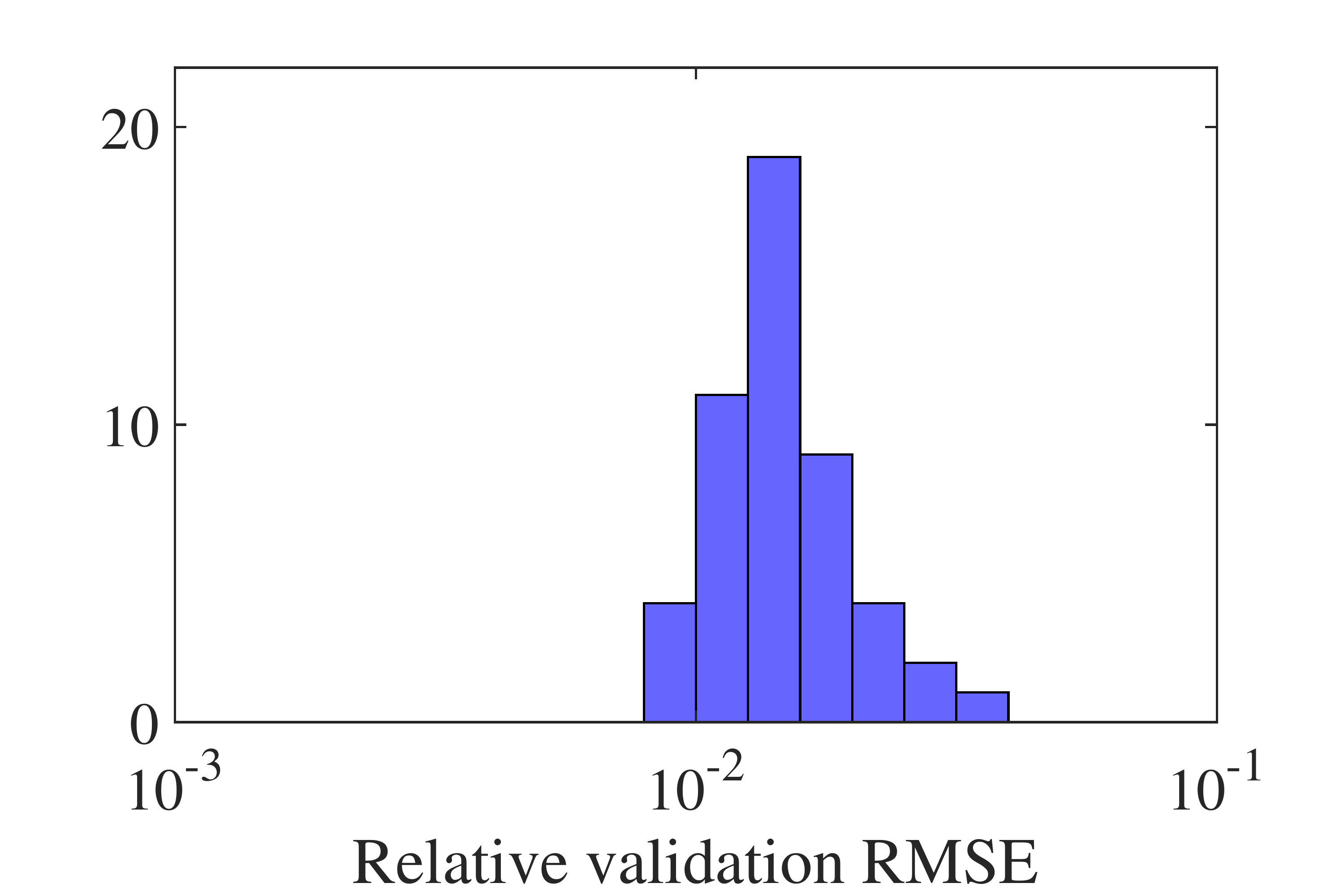}
		\caption{High-fidelity with no regularization}
	\end{subfigure}\\
	\begin{subfigure}[!htb]{0.48\textwidth}
		\centering
		\includegraphics[scale=0.25]{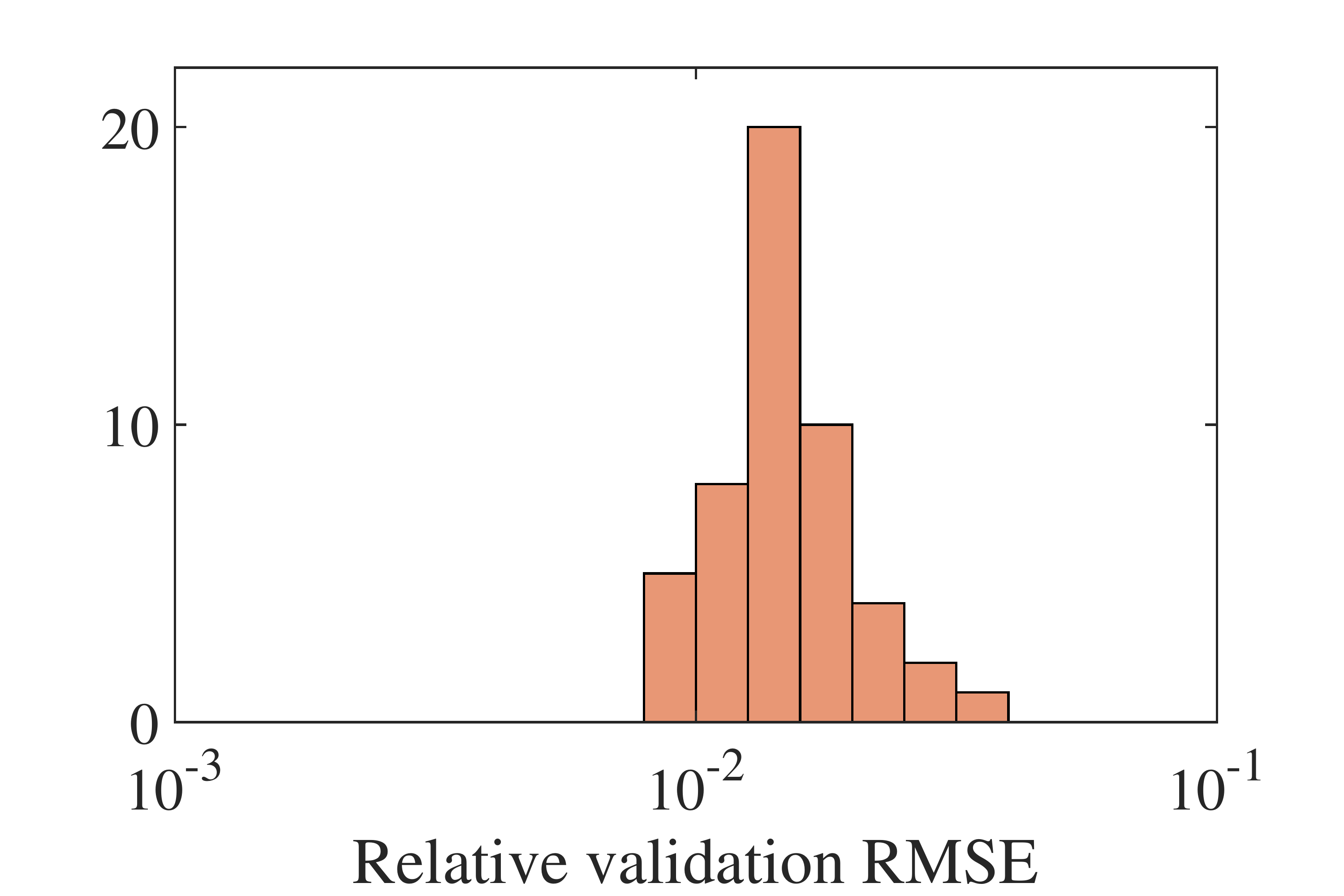}
		\caption{Strategy I: High-fidelity standard $\ell_1$-regularization}
	\end{subfigure}
	\hfill
	\begin{subfigure}[!htb]{0.48\textwidth}
		\centering
		\includegraphics[scale=0.25]{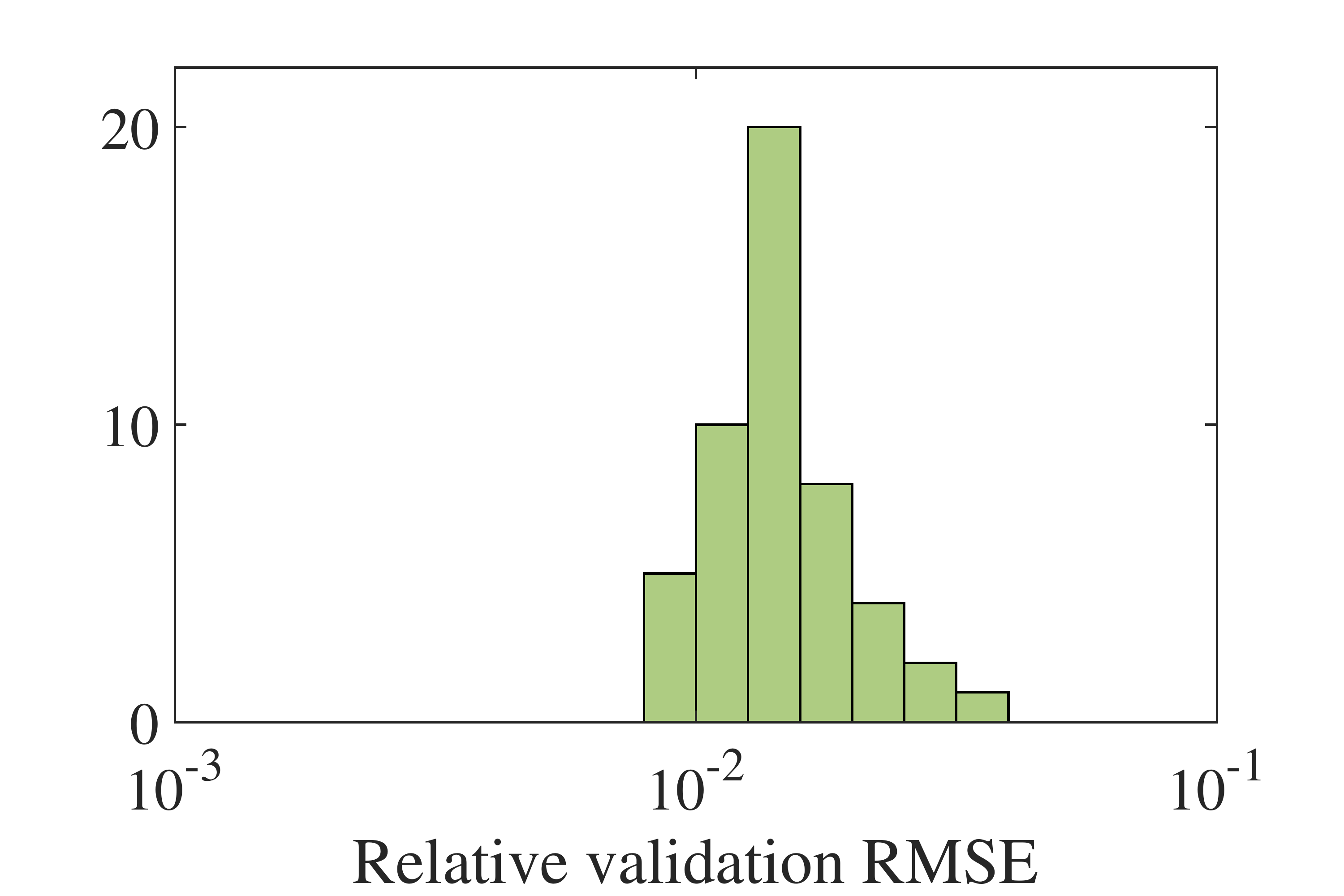}
		\caption{Strategy II: High-fidelity iteratively reweighted $\ell_1$-regularization}
	\end{subfigure}\\
	\begin{subfigure}[!htb]{0.48\textwidth}
		\centering
		\includegraphics[scale=0.25]{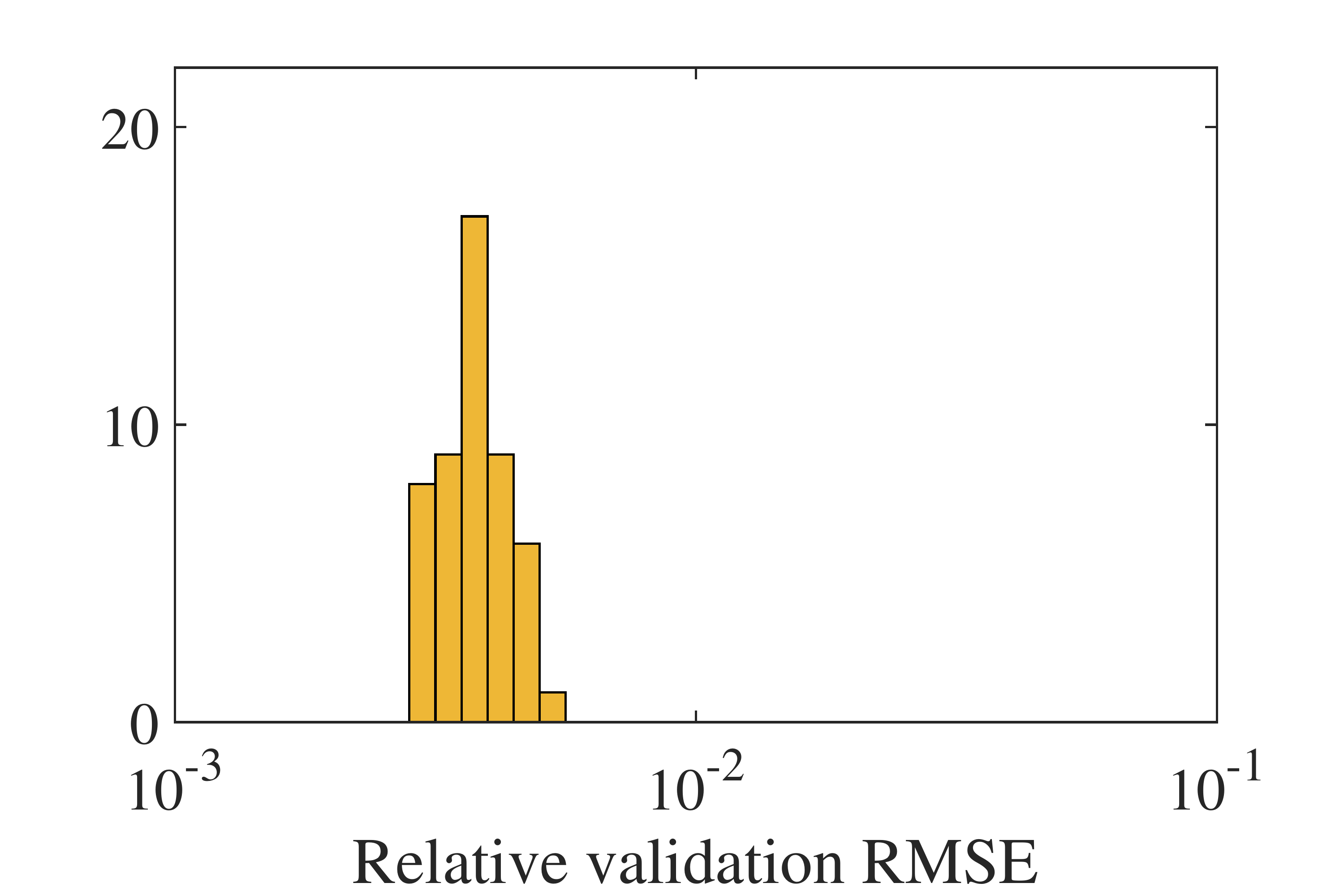}
		\caption{Strategy III: Bi-fidelity standard $\ell_1$-regularization}
	\end{subfigure}%
	\hfill
	\begin{subfigure}[!htb]{0.48\textwidth}
		\centering
		\includegraphics[scale=0.225]{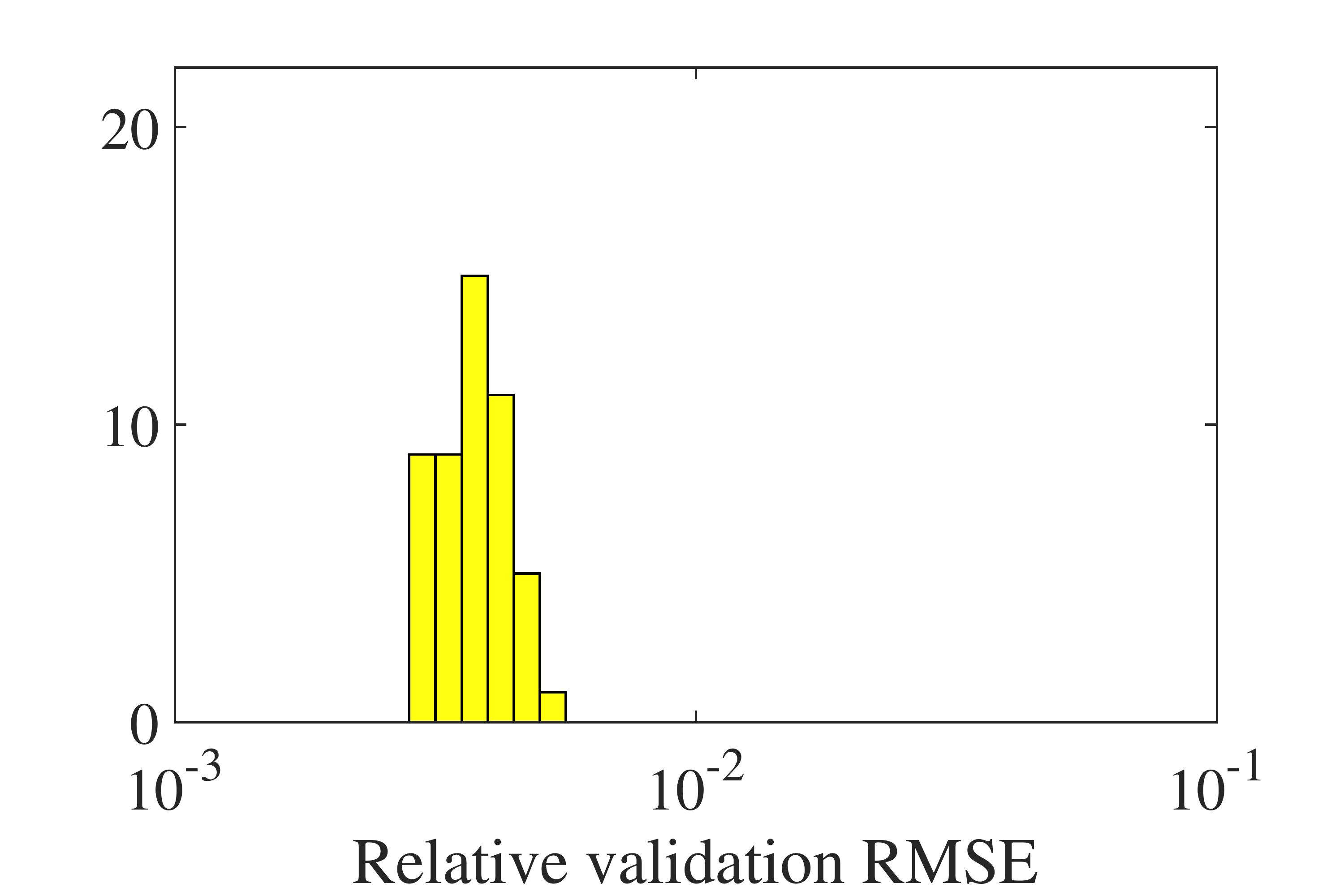}
		\caption{Strategy IV: Bi-fidelity weighted $\ell_1$-regularization}
	\end{subfigure}
	\caption{Histograms of relative validation RMSE, $\varepsilon_{\mathrm{v}}$, for 50 different replications of training/validation datasets for no regularization and four different $\ell_1$-regularization strategies in Example III.} \label{fig:ae_hist} 
\end{figure}

\begin{figure}
	\centering
	\includegraphics[scale=0.3]{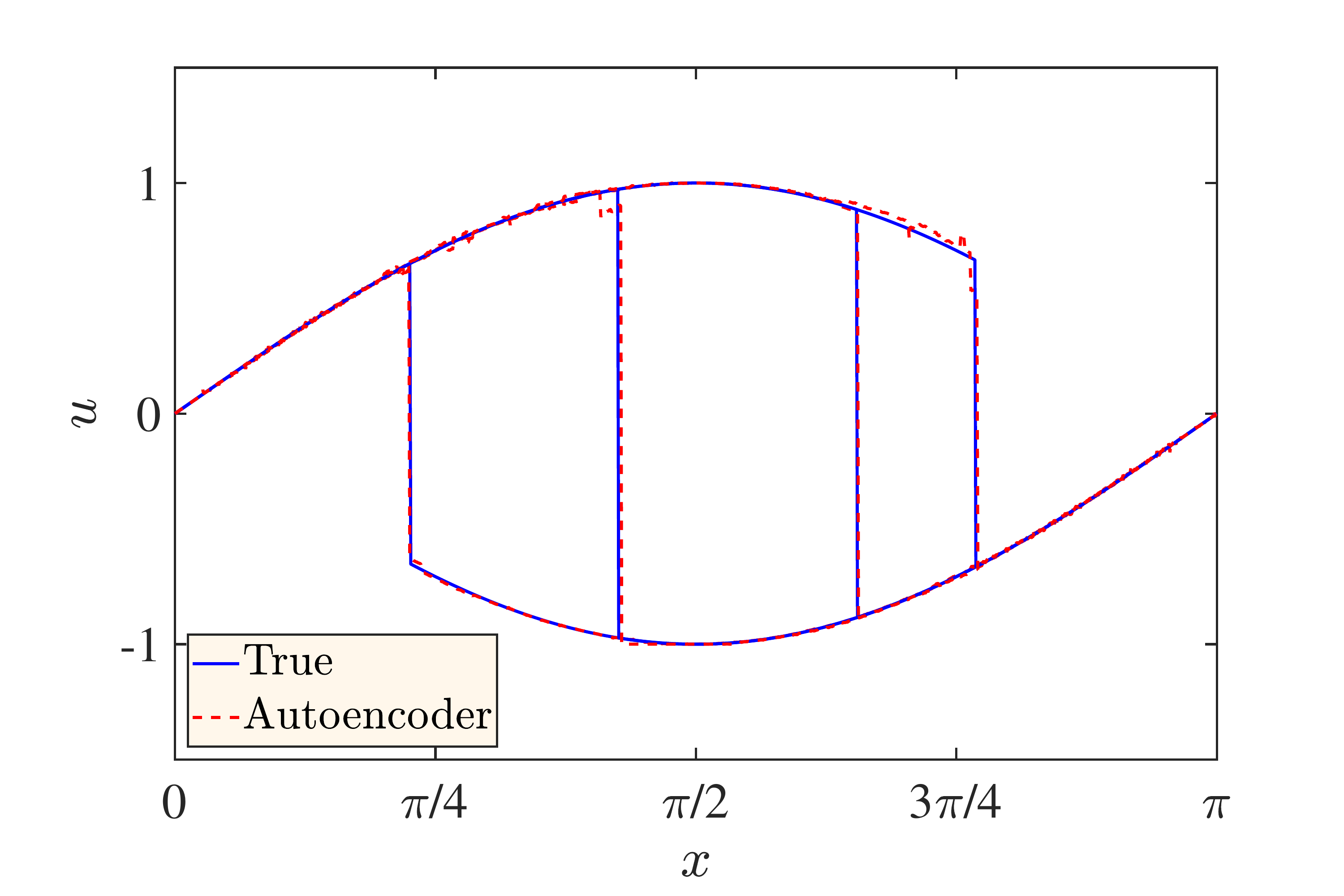}
	\caption{Comparison of the recosnstructed solution from the autoencoder trained using the bi-fidelity Strategy IV with the true solution in Example III for four different realizations of $\xi$. } 
	\label{fig:ae_compare}
\end{figure}

Table \ref{tab:ae_rmse} shows the mean and standard deviation of $\varepsilon_{\mathrm{v}}$ for the four $\ell_1$-regularization strategies described in Section \ref{sec:l1_strategies} with corresponding regularization strength $\lambda$ and a case where we do not use any regularization. Note that the optimal $\lambda$ is very small compared to previous examples since the number of connections and hence the number of parameters are very large in this example. Figure \ref{fig:ae_hist} plots histograms of $\varepsilon_{\mathrm{v}}$ for 50 different replications of training and validation datasets. These results show that the high-fidelity Strategy I does not provide significant improvements in $\varepsilon_{\mathrm{v}}$. On the other hand, the two bi-fidelity strategies III and IV provide one order of magnitude improvement in the mean and standard deviation of $\varepsilon_{\mathrm{v}}$ compared to the case where no regularization or only high-fidelity dataset is used for regularization during training. 
We also compare the reconstructed data from the autoencoder trained using the bi-fidelity Strategy IV with true values for four different realizations of $\xi$ in Figure \ref{fig:ae_compare}. 
Note that there may exist other dimension reduction methods that perform better than an autoencoder in this example, but a detailed comparison is beyond the scope of the current paper. Readers are referred to Hinton and Salakhutdinov \cite{hinton2006reducing}  and Makhzani et al. \cite{makhzani2015adversarial} for more details on comparison of autoencoder with other methods.

\begin{table}[!htb]
	\caption{The mean and standard deviation of relative validation RMSE, $\varepsilon_{\mathrm{v}}$, using different $\ell_1$-regularization strategies for Example III.}
	\label{tab:ae_rmse}
	\centering
	\begin{tabular}{l|c|c|c|c}
		\hline \Tstrut
		Method &  Dataset & $\ell_1$-regularization  & Mean $\varepsilon_{\mathrm{v}}$ & Std. of $\varepsilon_{\mathrm{v}}$ \Bstrut\\
		\hline \Tstrut
		&     High-fidelity & None & $1.5670\times10^{-2}$ & $5.5478\times10^{-3}$\\
		Strategy I &     High-fidelity & Standard ($\lambda=10^{-9}$) & $1.5668\times10^{-2}$ & $5.5447\times10^{-3}$\\
		Strategy II &        High-fidelity & Weighted ($\lambda=10^{-11}$) & $1.5603\times10^{-2}$ & $5.5505\times10^{-3}$\\
		Strategy III &       Bi-fidelity & Standard ($\lambda=10^{-9}$) & $3.7637\times10^{-3}$ & $5.3682\times10^{-4}$\\
		Strategy IV &       Bi-fidelity & Weighted ($\lambda=10^{-11}$) & $3.7616\times10^{-3}$ & $5.3952\times10^{-4}$\Bstrut\\\hline
	\end{tabular}
\end{table}


\section{Conclusions} 
Training of neural networks with sparse data has recently attracted the attention of researchers as they are increasingly used in scientific applications. 
In scientific computing applications, we often do not have a large amount of accurate high-fidelity data required for training of these networks. A trained network with a small dataset may remain deficient of producing accurate predictions. To alleviate this issue, in this paper, we investigate novel $\ell_1$-regularization methods for training to avoid overfitting in the absence of a large high-fidelity dataset. Apart from implementing a standard $\ell_1$-regularization, we explore a strategy where the $\ell_1$-regularization is applied to weighted parameters of the network using their values from the previous iteration. Further, we investigate strategies that use a combination of a small but accurate high-fidelity dataset and a large but less accurate low-fidelity dataset. In these bi-fidelity strategies, we first train an identical network using the low-fidelity dataset and then the parameters of this network are used for $\ell_1$-regularization of the high-fidelity network parameters. In particular, we use $\ell_1$-norm of the difference between the parameters of these two networks or a weighted $\ell_1$-norm where the weights are selected based on the magnitude of the low-fidelity network parameters. Three numerical examples are used to compare these different $\ell_1$-regularization strategies. In our first example, we use a composite beam with uncertain material properties and external loading. The mean validation error in this example shows one order of magnitude improvement compared to the case in which only a small high-fidelity dataset is used without any regularization. Similar observation has been made for the second example, where we train a feed-forward neural network for modeling the mean of the heat flux for thermally driven flow in a cavity. The third example, however, tackles the training of an autoencoder which can be used for non-linear dimensionality reduction. As the number of connections in this autoencoder are large the bi-fidelity $\ell_1$-regularization strategies proved effective. In future studies, we plan to investigate sparsity promoting strategies other than $\ell_1$-regularization and higher order optimization schemes in the presence of a bi-fidelity dataset.

\section*{Acknowledgment}
The authors acknowledge the support of the United States Department of Energy (DOE) under the ASCR
program under agreement DE-AC36-08GO28308. The work of AD was also supported by the AFOSR grant FA9550-20-1-0138. The opinions and conclusions presented in this paper are those of
the authors and do not necessarily reflect the views of DOE or AFOSR.

\appendix 

\section{Adam Algorithm} \label{sec:adam}

Adam, \cite{kingma2014adam}, leverages past gradient information to retard the descent along large gradients. This information is stored in the momentum vector $\mm$ and squared gradient vector $\vm$ at $k$th iteration as
\begin{equation}
\label{eqn:adam_updates1}
\begin{split}
\mm_{k} &= b_m \mm_{k-1} + (1-b_m) \frac{\partial  R_{i_k}}{\partial \thetaa_{k}}; \quad\widehat{\mm}_{k} = \frac{\mm_{k}}{1-b_m^k};\\
\vm_{k} &= b_v \vm_{k-1}+(1-b_v)\left[\frac{\partial R_{i_k}}{\partial \thetaa_{k}}\right]^{\circ 2}; \quad \widehat{\vm}_{k}=\frac{\vm_{k}}{1-b_v^k},\\
\end{split}
\end{equation}
where $[\cdot]^{\circ p}$ denotes a Hadamard power of $p$, i.e., the square in $\left[\frac{\partial R_{i_k}}{\partial \thetaa_{k}}\right]^{\circ 2}$ is performed element-wise; and $b_m$ and $b_v$ are parameters with default values 0.9 and 0.999, respectively. In (\ref{eqn:adam_updates1}), $\widehat{\mm}_{k}$ and $\widehat{\vm}_{k}$ are the unbiased momentum and  squared gradient vectors, respectively. The gradient descent step is applied next as follows
\begin{equation}
\label{eqn:adam_updates2}
\thetaa_{k+1} = \thetaa_{k} - \eta \widehat{\mm}_{k} \oslash \left( \widehat{\vm}_{k}^{\circ\frac{1}{2}} + \epsilon_a \right)  
\end{equation}
where $\eta$ is the learning rate; $\oslash$ denotes a Hadamard division (i.e., element-wise division); and $\epsilon_a$ is a small number to avoid division by zero. We use this algorithm to train the neural networks in this study. An illustration of the steps of this algorithm is shown in Algorithm \ref{alg:adam}. 

\begin{algorithm}
	\begin{algorithmic}
		\State \textbf{Network}: $\pinn(\cdot;\thetaa_0)$
		\State \textbf{Given}: $\eta$, $b_m$, $b_v$, and $\epsilon_a$.
		\Procedure{Adam}{}
		\State Initialize $\thetaa_{1}=\thetaa_0$.
		\State Initialize $\mm_0 = \mathbf{0}$.
		\State Initialize $\vm_0 = \mathbf{0}$.
		\For {$k=1,2,\dots,N_{\max}$}
		\State Compute $\frac{\partial J}{\partial \thetaa_{k}}$.
		\State Set $\mm_k \leftarrow b_m\mm_{k-1} + (1-b_m)\frac{\partial J}{\partial \thetaa_{k}}$. 
		\State Set $\vm_k \leftarrow b_v\vm_{k-1} + (1-b_v)\left[\frac{\partial J}{\partial \thetaa_{k}}\right]^{\circ 2}\qquad$(element-wise). 
		\State Set $\widehat\mm_k \leftarrow \mm_k/(1-b_m^k)$. 
		\State Set $\widehat\vm_k \leftarrow \vm_k/(1-b_v^k)$. 
		\State Set $\thetaa_{k+1} \leftarrow \thetaa_{k} - \eta\widehat{\mm}_{k} \oslash \left( \widehat{\vm}_{k}^{\circ\frac{1}{2}} + \epsilon_a \right)$
		\qquad(element-wise). 
		\EndFor
		\EndProcedure
	\end{algorithmic}
	\caption{\textit{Adam} \cite{kingma2014adam}}
	\label{alg:adam}
\end{algorithm}

\newtheorem{lemma}{Lemma}

\section{Theoretical Analysis of Proposition \ref{prop}}\label{sec:lemmas}
%
%

In this section, first we present two lemmas from statistical learning theory in the context of a neural network from Shalev-Shwartz and Ben-David \cite[Lemma 26.2]{shalev2014understanding}, Koltchinskii \cite[Theorem 2.3]{Koltchinskii2011}, and Rakhlin and Sridharan  \cite[Lemma 15.1]{rakhlin2014statistical}. 
Then, using these two lemmas we provide the proof of Proposition \ref{prop} for the four $\ell_1$-regularization strategies implemented in this work. 

\begin{lemma}  \label{lem1} (Shalev-Shwartz and Ben-David \cite[Lemma 26.2]{shalev2014understanding} and Koltchinskii \cite[Theorem 2.3]{Koltchinskii2011})
	The generalization error $\mathcal{L}_{\mathcal{S}}({\mathcal{F}},\mathcal{D})$ of a neural network  defined in \eqref{eq:gen_err} for a dataset $\mathcal{D}=\{\mathcal{D}_i\}_{i=1}^N$ generated from the probability measure $\mathcal{S}$, the hypothesis space $\mathcal{H}$, and $\mathcal{F}:=J\circ \mathcal{\mathcal{H}}$ for the loss function $J(\widehat{\mathcal{M}};\mathcal{D})$ is bounded by 
	\begin{equation}
	\mathbb{E}_{\mathcal{D}\sim \mathcal{S}}[\mathcal{L}_{\mathcal{S}}({\mathcal{F}},\mathcal{D})] \leq 2 \mathbb{E}_{\mathcal{D}\sim \mathcal{S}}\left[ \mathcal{R}_{{\mathcal{D}}}(\mathcal{H}) \right],
	\end{equation}
	where $\mathcal{R}_{{\mathcal{D}}}(\mathcal{H})$ is the empirical Rademacher complexity of the hypothesis space $\mathcal{H}$ and dataset $\mathcal{D}$. 
\end{lemma}
Note that the above lemma is also known as symmetrization. Next, we present a version of Lemma 15.1 from Rakhlin and Sridharan \cite{rakhlin2014statistical}. We provide the proof of this lemma here as we extend it to a case that incorporates biases in the network and also for completeness.

\begin{lemma} \label{lem2} (Rakhlin and Sridharan \cite[Lemma 15.1]{rakhlin2014statistical})
	For a neural network with hypothesis space  $\mathcal{H}:=\mathcal{H}_{\mathrm{inp}} \cup \left(\bigcup_{j=0}^{N_H}\mathcal{H}_j\right)$ as defined in \eqref{eq:nn_hyp}, the Rademacher complexity is given by 
	\begin{equation}
	\mathcal{R}_{\mathcal{D}}(\mathcal{H}) \leq \left(\prod_{j=0}^{N_H} 2L_j \right) \mathcal{R}_{\mathcal{D}}(\mathcal{H}_\mathrm{inp}),
	\end{equation}
	where $\lVert \thetaa_j \rVert_1 \leq L_j$ for the $j$th layer. 
\end{lemma}

\textit{Proof}: For the $j$th hidden layer, the empirical Rademacher complexity is defined as
\begin{equation}
\begin{split}
\mathcal{R}_{\mathcal{D}}(\mathcal{H}_j) = \frac{1}{N} \mathbb{E}_{\boldsymbol{\mathcal{E}}}\left[ \sup_{\substack{h_l\in\mathcal{H}_{j-1}\\ \forall l}} \sum_{i=1}^{N} \epsilon_i \left( \sum_{l}     \psi_{j,l}\sigma(h_l(\mathcal{D}_i))+\beta_{j} \right)  \right], 
\end{split}
\end{equation}
where the hypothesis space $\mathcal{H}_j$ for the $j$th hidden layer is defined in \eqref{eq:nn_hyp}. Next, using the H\"{o}lder's inequality and recognizing that $\epsilon$ and $-\epsilon$ both are Rademacher random variables we can write
\begin{equation}
\begin{split}
\mathcal{R}_{\mathcal{D}}(\mathcal{H}_j) & \leq \frac{1}{N} \mathbb{E}_{\boldsymbol{\mathcal{E}}}\left[ \sup_{\substack{h_l\in\mathcal{H}_{j-1}\\ \forall l}} \lVert \thetaa_j \rVert_1 \sup_{l}\Bigg\lvert \sum_{i=1}^{N} \epsilon_i \bar{\sigma}(h_l(\mathcal{D}_i)) \Bigg\rvert \right]\\
& \leq \frac{L_j}{N} \mathbb{E}_{\boldsymbol{\mathcal{E}}}\left[ \sup_{\substack{h\in\mathcal{H}_{j-1}}} \Bigg\lvert \sum_{i=1}^{N} \epsilon_i \bar{\sigma}(h(\mathcal{D}_i)) \Bigg\rvert \right]\\
& \leq \frac{2L_j}{N} \mathbb{E}_{\boldsymbol{\mathcal{E}}}\left[ \sup_{\substack{h\in\mathcal{H}_{j-1}}} \sum_{i=1}^{N} \epsilon_i \bar{\sigma}(h(\mathcal{D}_i)) \right]\\
& = 2L_j \mathcal{R}_{\mathcal{D}}(\bar{\sigma}\circ\mathcal{H}_{j-1}) \leq 2L_j \mathcal{R}_{\mathcal{D}}(\mathcal{H}_{j-1}),
\end{split}
\end{equation}
where $\bar{\sigma}(h(\mathcal{D}_i))$ consists of the activation function for the weights and the constant function 1 for the bias. In the last inequality we use the fact that $\bar{\sigma}(h(\mathcal{D}_i))$ is Lipschitz continuous with constant 1 for the activation functions used in this paper. 
Hence, combining the results from all the layers, we arrive at  Lemma \ref{lem2}.

\subsection{Proof of Proposition \ref{prop}}
From Lemmas \ref{lem1} and \ref{lem2}, we can write 
\begin{equation} 
\mathbb{E}_{\mathcal{D}\sim \mathcal{S}}[\mathcal{L}_{\mathcal{S}}({\mathcal{F}},\mathcal{D})] \leq 2 \left(\prod_{j=0}^{N_H} 2L_{j}\right) \mathbb{E}_{\mathcal{D}\sim \mathcal{S}}\left[ \mathcal{R}_{\mathcal{D}}(\mathcal{H}_\mathrm{inp}) \right].
\end{equation}
This directly gives $K_\mathrm{std}^{\mathrm{HF}} = \prod_{j=0}^{N_H} 2L_{j}$ in \eqref{eq:gen_err_bound} for the standard high-fidelity Strategy I. For the weighted strategies, we use the fact that the weights are positive and $\lVert \Wm_j\thetaa_j \rVert_1 = \sum_{i=1}^{n_{\thetaa_j}} w_{ii}^j\lvert \theta_{j,i} \rvert \leq L_{w,j}$ for network parameters of the $j$th layer  to write 
\begin{equation} \label{eq:Lwj}
\begin{split}
\lVert \thetaa_j \rVert_1 = \sum_{i=1}^{n_{\thetaa_j}} \lvert \theta_{j,i} \rvert \leq L_j \leq \sum_{i=1}^{n_{\thetaa_j}} \frac{w_{ii}^j}{w^j_{\min}} \lvert \theta_{j,i} \rvert \leq \frac{L_{w,j}}{w^j_{\min}},
\end{split}
\end{equation}
where $w_{\min}^j$ is the minimum weight for the parameters of the $j$th layer; $\frac{L_{w,j}}{w^j_{\min}}=L_{w,j}\left( \Big\lvert \thetaa_{k-1,j}^{\mathrm{max}}\Big\rvert + \epsilon_w \right)$ for Strategy II; and $\frac{L_{w,j}}{w^j_{\min}}=L_{w,j}\left(\Big\lvert \thetaa_{\mathrm{LF},j}^{\mathrm{max}}\Big\rvert + \epsilon_w \right)$ for the weighted bi-fidelity Strategy IV. For the standard bi-fidelity Strategy III, we use
\begin{equation}\label{eq:Ldj}
\lVert \thetaa_j \rVert \leq L_j \leq  \lVert \thetaa_j - \thetaa_{\mathrm{LF},j} \rVert_1 + \lVert \thetaa_{\mathrm{LF},j} \rVert = L_{d,j} + L_{\mathrm{LF},j}.
\end{equation}
Using \eqref{eq:Lwj} and \eqref{eq:Ldj}, we can replace $L_j$ in \eqref{eq:gen_err_bound} for the final three strategies to write  \eqref{eq:K}. 

Next, we use the result from Theorem \ref{thm1} for a convex objective $R(\thetaa)$ and assume the parameter values at $k^*$th iteration produces the optimal objective to write 
\begin{equation} \label{eq:B8}
\mathbb{E} \left[ J(\thetaa_{k^*}) + \lambda \lVert \thetaa_{k^*} \rVert_1 - \left(J(\thetaa^*) + \lambda \lVert \thetaa^* \rVert_1\right) \right] \leq \frac{\rho^2 + \gamma^2 \sum_{i=1}^k \eta_i^2}{2\sum_{i=1}^k \eta_i}. 
\end{equation}
Dividing both sides of \eqref{eq:B8} by the positive parameter $\lambda$ and denoting $C=\frac{\rho^2 + \gamma^2 \sum_{i=1}^k \eta_i^2}{2\lambda\sum_{i=1}^k \eta_i}$ and the best training error $\mathcal{E}_\mathrm{tr} = \mathbb{E}[J(\thetaa_{k^*}) - J(\thetaa^*)]$ leads to 
\begin{equation}
\mathbb{E} \left[\lVert \thetaa_j \rVert_1 \right] \leq \mathbb{E} \left[\lVert \thetaa \rVert_1 \right] \leq C + \lVert \thetaa^* \rVert_1 -\frac{\mathcal{E}_\mathrm{tr}}{\lambda},
\end{equation}
where we select the network parameters $\thetaa$ from $k^*$th iteration; $\thetaa_j$ is the selected parameters of the $j$th layer; and we use $\mathbb{E} \left[\lVert \thetaa^* \rVert_1 \right] = \lVert \thetaa^* \rVert_1$. 
Similarly, we can show the other right-hand inequalities in \eqref{eq:Lj_inequals}, where we use $\mathbb{E}[XY] = \mathrm{Cov}(X,Y)+\mathbb{E}[X]\mathbb{E}[Y]$ for strategies II and IV.

\bibliographystyle{unsrt}  
\bibliography{references} 

\end{document}